\documentclass[11pt,a4paper]{article}
\pdfoutput=1 
\usepackage{authblk}
\usepackage[hyperref]{main}
\usepackage{times}
\usepackage{latexsym}
\usepackage{amsmath}
\usepackage{caption}
\usepackage{subcaption}
\usepackage{colortbl}
\usepackage{float}
\usepackage{multirow}
\usepackage[many]{tcolorbox}
\usepackage{xcolor}
\usepackage{graphicx}
\usepackage{tabularx}
\usepackage{dcolumn}
\usepackage{array}

\usepackage{url}

\definecolor{navy}{rgb}{0.0, 0.0, 0.5}
\definecolor{crimson}{rgb}{0.6, 0.0, 0.0}

\aclfinalcopy 
\def\aclpaperid{2238} 

\title{Analyzing Polarization in Social Media:\\Method and Application to Tweets on 21 Mass Shootings}
\author{\vspace{-3mm}{\bf Dorottya Demszky}\textsuperscript{1}\quad {\bf Nikhil Garg}\textsuperscript{1}\quad {\bf Rob Voigt}\textsuperscript{1}\quad {\bf James Zou}\textsuperscript{1}\\ {\bf Matthew Gentzkow}\textsuperscript{1}\quad {\bf Jesse Shapiro}\textsuperscript{2}\quad {\bf Dan Jurafsky}\textsuperscript{1}\\\textsuperscript{1}Stanford University\quad \textsuperscript{2}Brown University \\ \texttt{\{ddemszky, nkgarg, robvoigt, jamesz, gentzkow, jurafsky\}@stanford.edu}\\\texttt{jesse\_shapiro\_1@brown.edu}}

\date{}

\begin{document}
\maketitle
\begin{abstract}
 We provide an NLP framework to uncover four linguistic dimensions of political polarization in social media: topic choice, framing, affect and illocutionary force. We quantify these aspects with existing lexical methods, and propose clustering of tweet embeddings as a means to identify salient topics for analysis across events; human evaluations show that our approach generates more cohesive topics than traditional LDA-based models. We apply our methods to study 4.4M tweets on 21 mass shootings. We provide evidence that the discussion of these events is highly polarized politically and that this polarization is primarily driven by partisan differences in framing rather than topic choice. We identify framing devices, such as grounding and the contrasting use of the terms ``terrorist'' and ``crazy'', that contribute to polarization. Results pertaining to topic choice, affect and illocutionary force suggest that Republicans focus more on the shooter and event-specific facts (news) while Democrats focus more on the victims and call for policy changes. Our work contributes to a deeper understanding of the way group divisions manifest in language and to computational methods for studying them.\footnote{All data and code is available at: \url{https://github.com/ddemszky/framing-twitter}}
\end{abstract}

\section{Introduction}

Elites, political parties, and the media in the US are increasingly polarized \cite{layman2010party,prior2013media,gentzkow2018measuring}, and the propagation of partisan frames
can influence public opinion \cite{chong2007framing} and party identification \cite{fiorina2008political}.

Americans increasingly get their news from internet-based sources \cite{mitchell2016modern}, and political information-sharing is highly ideologically segregated on platforms like Twitter \cite{conover2011political,halberstam2016homophily} and Facebook \cite{bakshy2015exposure}. Prior NLP work has
shown, e.g., that
polarized messages are more likely to be shared \cite{zafar2016message} and that certain topics are more polarizing \cite{balasubramanyan2012modeling}; however, we lack a more broad understanding of the many ways that polarization can be instantiated linguistically.

This work builds a more comprehensive framework for studying linguistic aspects of polarization in social media, by looking at topic choice, framing, affect, and illocutionary force.

\subsection{Mass Shootings}

We explore these aspects of polarization by studying a sample of more than 4.4M tweets about 21 mass shooting events, analyzing polarization within and across events.

Framing and polarization in the context of mass shootings is well-studied, though much of the literature studies the role of media~\cite{chyi_media_2004,schildkraut_mass_2016} and politicians~\cite{johnson2017modeling}. Several works find that frames have changed over time and between such events~\cite{muschert_media_2006,schildkraut_media_2014}, and that frames influence opinions on gun policies~\cite{haidermarkel_gun_2001}. Prior NLP work in this area has considered how to extract factual information on gun violence from news~\citep{pavlick2016gun} as well as quantify stance and public opinion on Twitter~\citep{benton2016after} and across the web~\citep{ayers2016can};
here we advance NLP approaches to the public discourse surrounding gun violence by introducing methods to analyze other linguistic manifestations of polarization.





\subsection{The Role of the Shooter's Race}

We are particularly interested in the role of the shooter's race in shaping polarized responses to these events. Implicit or explicit racial biases can be central in people's understanding of social problems \cite{drakulich2015explicit}; in the mass shooting context, race is a factor in an event's newsworthiness \cite{schildkraut_mass_2018} and is often mentioned prominently in media coverage, particularly when the shooter is non-white \cite{mingus2010white,park2012race}.
\citet{Duxbury2018} find that media representations of white shooters disproportionately divert blame by framing them as mentally ill while representations of non-white shooters are more frequently criminalized, highlighting histories of violent behavior.
    
The important question remains as to how polarized ideologies surrounding race take shape on forums such as Twitter. Therefore, in all of the analyses throughout this paper we consider the race of the shooter as a potential factor. We note that in the 21 shooting events we study, shootings in schools and places of worship are overwhelmingly carried out by white perpetrators, so we cannot fully disentangle the effect of race from other factors.

\section{Data: Tweets on Mass Shootings}
\label{sec:data}

\paragraph{Data collection.} We compiled a list of mass shootings between 2015 and 2018 from the Gun Violence Archive.\footnote{\url{https://www.gunviolencearchive.org}} For each, we identified a list of keywords representative of their location (see Appendix~\ref{sec:appendix_data}). Given Twitter search API's limitations on past tweets, we retrieved data from a Stanford lab's archived intake of the Twitter firehose.\footnote{With the exception of the two most recent shootings in Pittsburgh and Thousand Oaks, for which we collected tweets real time via the Twitter API.}



For each event, we built a list of relevant tweets for the two weeks following the event. A tweet is relevant if it contained at least one of the event's location-based representative keywords and at least one lemma from the following list: ``shoot'', 
``gun'', ``kill'', ``attack'', ``massacre'', ``victim''. We filtered out retweets and tweets from users who have since been deactivated. We kept those 21 events with more than 10,000 tweets remaining. For more details see  Appendix~\ref{sec:appendix_data}.

\begin{figure}[]
 \centering
   \centering
   \includegraphics[width=\linewidth]{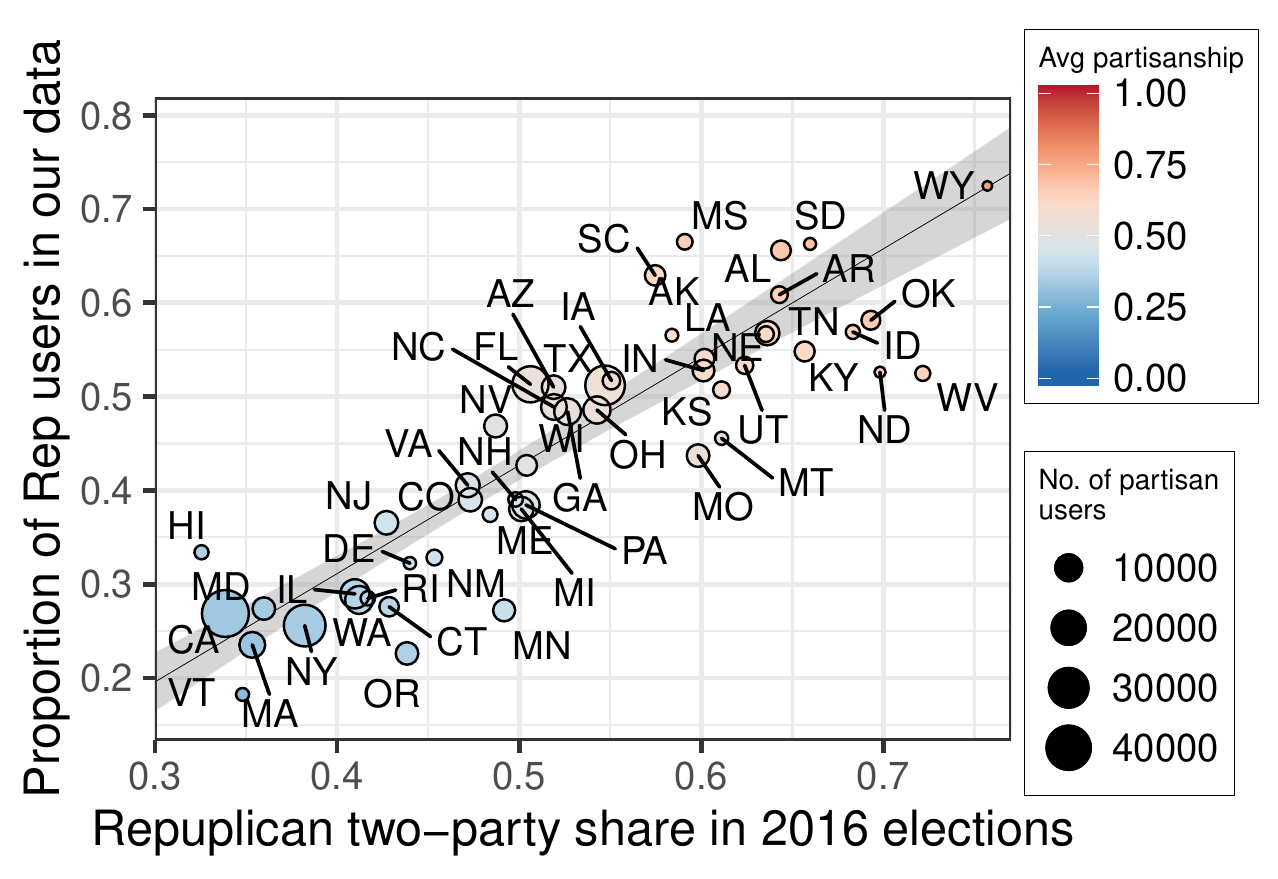}
   \caption{To validate our partisanship assignment, we compare the proportion of users we classify as Republican separately for each US state against the Republican share of the two-party vote. A weighted linear regression with the number of partisan users per state as weights obtains an adjusted $R^2$ of $.82$, indicating that the distribution of Republican users in our data and that of Republican voters across states is similar.} 
   \label{fig:sanity_check}
\end{figure}

\paragraph{Partisan assignment.} We estimate the party affiliation of users in the dataset from the political accounts they follow using a method similar to that of Volkova et al. \shortcite{volkova2014inferring}, which takes advantage of homophily in the following behavior of users on twitter~\cite{halberstam2016homophily}. We compile a list of Twitter handles of US Congress members in 2017, the 2016 presidential and vice presidential candidates, and other party-affiliated pages.\footnote{See Appendix~\ref{ssec:appendix_politician_handles} for the complete list.} We label a user as a Democrat if they followed more Democratic than Republican politicians in November 2017, and as a Republican if the reverse is true. For each event,  51--72\% of users can be assigned partisanship in this way; to validate our method we compare state averages of these inferred partisan labels to state two-party vote shares, finding a high correlation (Figure~\ref{fig:sanity_check}).\footnote{We performed the sanity check for all partisan users with a valid US state as part of their geo-location ($\sim$350k users).}

\section{Quantifying Overall Polarization}
\label{sec:exp1_polarization}

We begin by quantifying polarization (equivalently, partisanship) between the language of users labeled Democrats and Republicans after mass shooting events. We establish that there is substantial polarization, and that the polarization increases over time within most events. 


\subsection{Methods}
\label{ssec:exp1_methods}
\paragraph{Pre-processing.} We first build a vocabulary for each event as follows. Each vocabulary contains unigrams and bigrams that occur in a given event's tweets at least 50 times, counted after stemming via NLTK's SnowballStemmer and stopword removal.\footnote{Stopword list is provided in Appendix~\ref{ssec:appendix_stopwords}} We refer to these unigrams and bigrams collectively as \emph{tokens}.

\paragraph{Measure of partisanship.} We apply the leave-out estimator of phrase partisanship from Gentzkow et al. \shortcite{gentzkow2018measuring}. Partisanship is defined as the expected posterior probability that an observer with a neutral prior would assign to a tweeter's true party after observing a single token drawn at random from the tweets produced by the tweeter. If there is no difference in token usage between the two parties, then this probability is $.5$, i.e. we cannot guess the user's party any better after observing a token.

The leave-out estimator consistently estimates partisanship under the assumption that a user's tokens are drawn from a multinomial logit model. The estimator is robust to corpus size. The leave-out estimate of partisanship $\pi^{LO}$ between Democrats $i \in D$ and Republicans $i \in R$ is
\begin{equation*}
    \pi^{LO} = \frac{1}{2} \bigg( \frac{1}{|D|} \sum_{i \in D} \mathbf{\hat{q}}_i \cdot \boldsymbol{\hat{\rho}}_{-i} + \frac{1}{|R|} \sum_{i \in R} \mathbf{\hat{q}}_i \cdot (1-\boldsymbol{\hat{\rho}}_{-i}) \bigg)
\end{equation*}
where $\mathbf{\hat{q}}_i = \mathbf{c}_i / m_i$ is the vector of empirical token frequencies for tweeter $i$, with $\mathbf{c}_i$ being the vector of token counts for tweeter $i$ and $m_i$ the sum of token counts for tweeter $i$; and $\boldsymbol{\hat{\rho}}_{-i} = ( \mathbf{\hat{q}}^{D\setminus i } \oslash (\mathbf{\hat{q}}^{D\setminus i} + \mathbf{\hat{q}}^{R\setminus i}) )$ is a vector of empirical posterior probabilities, excluding speaker $i$ and any token that is not used by at least two speakers. Here we let $\oslash$ denote element-wise division and $\mathbf{\hat{q}}^{G}=\sum_{i \in G} \mathbf{c}_i/ \sum_{i \in G} m_i$ denote the empirical token frequency of tweeters in group $G$. The estimator thus captures two intuitive components of partisanship: between-group difference (posterior probability for each feature), and within-group similarity (dot-product between the feature vector of each speaker and that of their group).

\paragraph{User-level measures.} As the above leave-out estimator represents the average of user-level polarization values, we take the user-level dot product ($\mathbf{\hat{q}}_i \cdot \boldsymbol{\hat{\rho}}_{-i}$) as an estimate of the polarization of user $i$'s language. We consider the correlation of this value and the number of politicians a user follows in total and from their preferred party.

\subsection{Results and Discussion}
\label{ssec:exp1_results}

\paragraph{Overall polarization.}
\begin{figure}[]
 \centering
   \centering
   \includegraphics[width=\linewidth]{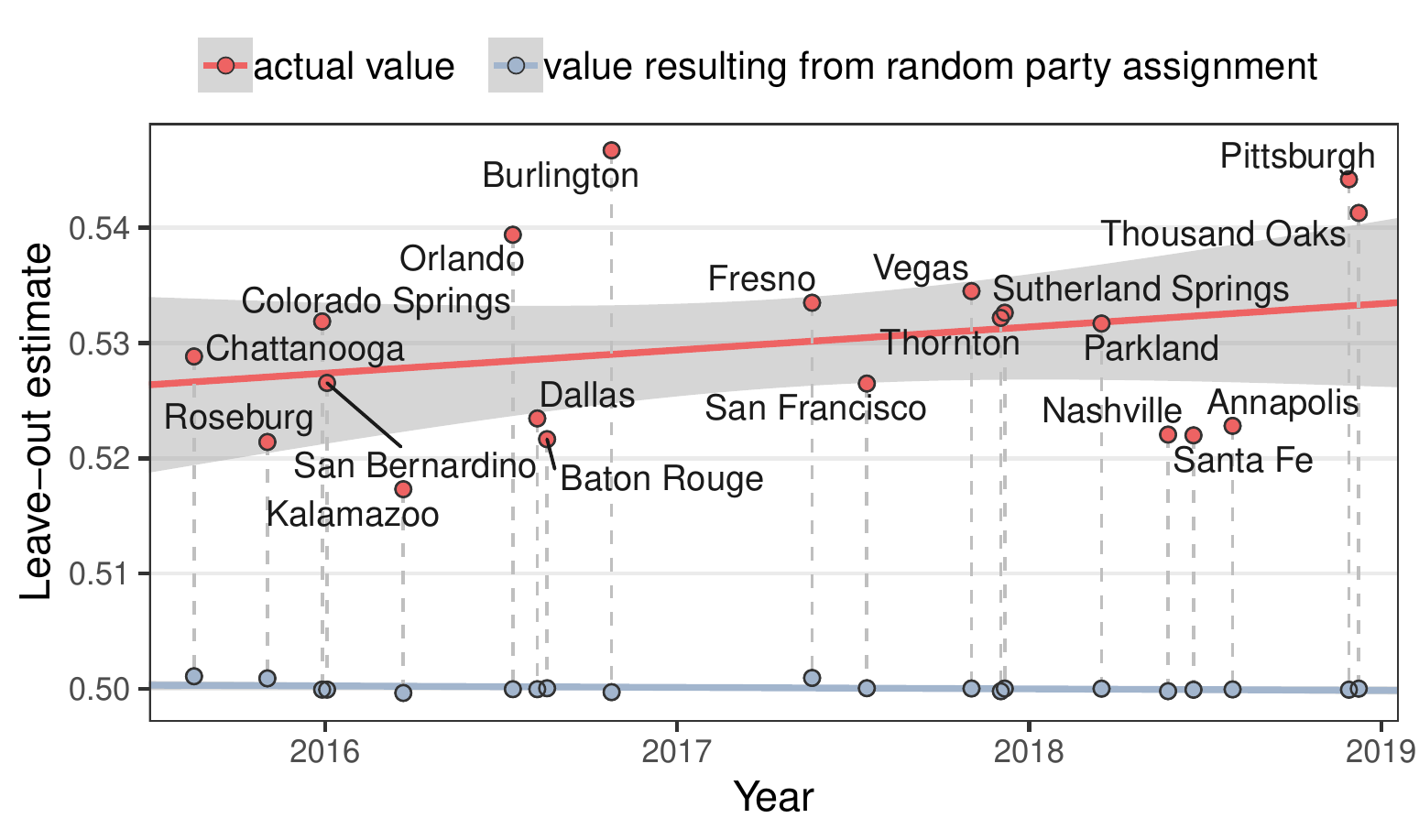}
   \caption{Tweets on mass shootings are highly polarized, as measured by the leave-out estimate of phrase partisanship \cite{gentzkow2018measuring}. The shaded region represents the 95\% confidence interval of the linear regression fit to the actual values. To quantify noise, we also calculate the leave-out estimate after randomly assigning users to parties, matching the ratio of parties in the true data. The ``values resulting from random assignment" are all close to $.5$, suggesting that the actual values are not a result of noise.}
   \label{fig:polarization_over_years}
\end{figure}

As Figure~\ref{fig:polarization_over_years} shows, the discussion of each event\footnote{For all the experiments using the leave-out estimator, we exclude Fort Lauderdale for which we only have tweets for the first day after the event;  polarization is most dominant a few days after each event, making it incomparable.
For reference, the leave-out estimate for Fort Lauderdale is $.51$.} is highly polarized: values range from $.517$ to $.547$. For comparison, this measure, for most events, is similar to or higher than the polarization in the US congress ($\sim.53$  in recent years)~\cite{gentzkow2018measuring}. While we observe a slight increase in polarization over the past three years, this increase is not statistically significant ($p \approx .26$). 

\paragraph{Post-event polarization.} 

\begin{figure}[htb]
 \centering
   \centering
   \includegraphics[width=\linewidth]{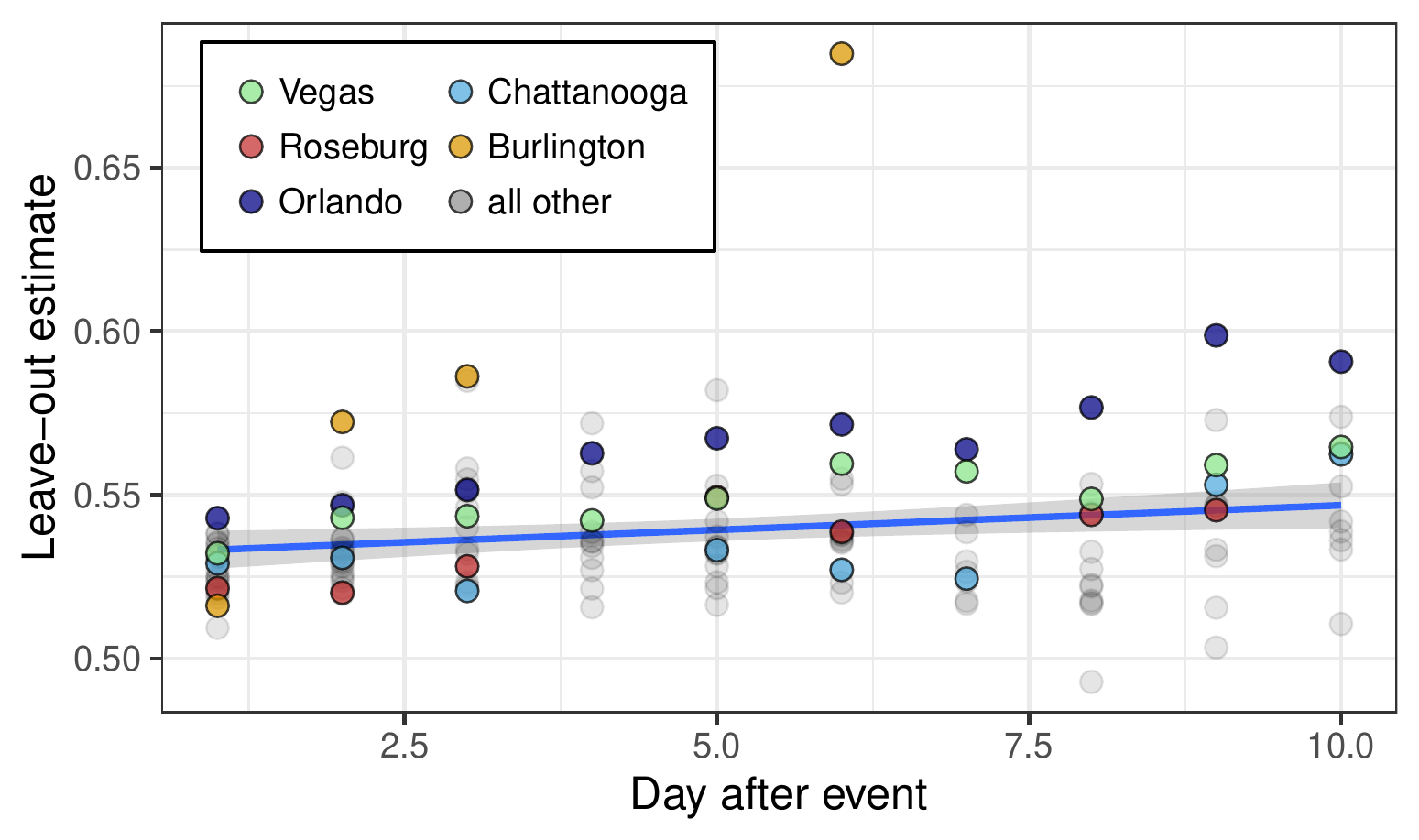}
   \caption{Leave-out estimate post-event.}
   \label{fig:leaveout_over_time}
\end{figure}

To see how polarization changes at the event level, we computed the leave-out estimate for each of the first 10 days following the events (see Figure~\ref{fig:leaveout_over_time}). An event-day level regression of partisanship on days since the event suggests a slight increase in post-event polarization across events (slope = $.002$, $p < 0.05$).  Fitting separate regressions, we find that the five events with the steepest increase in polarization are Burlington (slope = $.03$, $p < 0.05$), Orlando (slope = $.006$, $p<0.001$), Las Vegas (slope = $.003$, $p<0.001$), Chattanooga (slope = $.003$, $p<0.05$) and Roseburg (slope = $.003$, $p<0.05$) --- note that except for Las Vegas, these shootings are all committed by a person of color. 

Are the changes in the leave-out score due to \emph{different} users tweeting at different times or due to the \emph{same} users becoming more or less political? We found that while on average only $\sim10\%$ of users tweeted on multiple days ($SD=5\%$) across the events, these users contribute $\sim28\%$ of the tweets ($SD=15\%$). After removing these users from the leave-out estimation, we found that the temporal patterns remain with the same statistical significance, providing one piece of evidence that changes in polarization are not due to changes within users who tweet on multiple days.

\paragraph{User-level polarization.} We estimated a linear regression of the leave-out score on the total number of followed politicians and the number from the user's preferred party, with controls for event indicators. The estimates imply that, fixing the total number of followed politicians, one more followed politician from one's preferred party is associated with an \emph{increase} of $.009$ SD in the leave-out. Fixing the number of followed politicians from the user's preferred party, one more followed politician is associated with a \emph{decrease} of $.02$ SD in the leave-out.
\section{Topics and Framing}
\label{sec:exp2_topics}

Topic choice can be a tool for agenda-setting by establishing what an author or institution deems worthy of discussion \cite{mccombs2002agenda}, and works in NLP have used topic modeling as an approach to measure this effect \cite{tsur2015frame,field2018framing}. 
The strategy of highlighting particular aspects within topics as a means of framing \cite{entman2007framing} has also been quantified in the NLP literature \cite{boydstun2013identifying,card2015media,naderi2017classifying}.

Previous work largely focuses on the relation between topic and framing in the news media; we study social media, proposing methods to identify general, non-event-specific topics and to quantify between- and within-topic polarization.

 %



\subsection{Methods}
\label{ssec:exp2_methods}

\paragraph{Topic assignment.} Our goal is to induce topics that are salient in our narrow domain and comparable across events. This presents a challenge for traditional topic modeling approaches, since the discourse surrounding these events is inherently tied to concrete aspects of the events that tend to co-vary with topic usage, like location, setting, and demographics of the shooter and victims.

We build on the ability of vector space models to represent higher-level semantics to develop our own embedding-based topic assignment approach, comparing it with two traditional LDA-based methods: MALLET\footnote{\url{http://mallet.cs.umass.edu/topics.php}} and the Biterm Topic Model (BTM) \cite{yan2013biterm}; BTM was developed specifically for tweets. For all of these methods, we first randomly sample 10k tweets from each event forming our subset $S$ of all tweets $T$; then, we create a vocabulary $V$ of word stems that occur at least ten times in at least three events within $S$ ($\sim$2000 word stems) and remove all stems from $T$ are not part of $V$. Sampling is crucial for encouraging event-independent topics given the large disparity among event-level tweet counts (the largest event, Orlando, has 225$\times$ more tweets than the smallest event, Burlington).

For the embedding-based approach, we:
    
   \noindent 1.\quad Train GloVe embeddings \cite{pennington2014glove} on $V$ based on 11-50k random samples of tweets from each event.\footnote{This sample is different from $S$ as it includes more tweets to increase data size, which is important for training the embeddings, where the slightly disproportional representation of events is less problematic.}
   
   \noindent 2.\quad Create sentence embeddings $\mathbf{e}_t, \forall t \in T$ using \newcite{arora2017simple}'s method, by computing the weighted average $\mathbf{v}_t$ of the embeddings of stems within $t$ and removing $\mathbf{v}_t$'s projection onto the first principal component of the matrix the rows of which are $\mathbf{v}_t, \forall t\in S$. Stem weights are set to be inversely proportional to their frequencies in $S$.
   
   \noindent 3.\quad Jointly cluster the embeddings $\mathbf{e}_t,\forall t \in S$ via k-means using cosine distance and assign all tweet embeddings $\mathbf{e}_t,\forall t \in T$ to the centroids to which they are closest.
 
We also trained MALLET and BTM on $S$ and used the resulting models to infer topics for all tweets in $T$, assigning each tweet to its highest probability topic. Henceforth, we use $d$ to mean cosine distance for k-means and probabilities for MALLET and BTM.
 
A manual inspection found that about $25\%$ of the tweets are either difficult to assign to any topic or they represent multiple topics equally. To filter out such tweets, for each tweet we looked at the ratio of $d$ to its closest and second closest topic and removed tweets that have ratios higher than the $75^\text{th}$ percentile (calculated at the model-level).\footnote{This procedure filters out $11$-$26\%$ tweets (M=$22\%$, SD=$4\%$) across events, for our model, for eight topics.}

To compare the models, we ran two MTurk experiments: a word intrusion task \cite{chang2009reading} and our own, analogically defined tweet intrusion task, with the number of topics $k$ ranging between $6$-$10$. Turkers were presented with either a set of $6$ words (for word intrusion) or a set of $4$ tweets (for tweet intrusion), all except one of which was close (in terms of $d$) to a randomly chosen topic and one that was far from that topic but close to another topic. Then, Turkers were asked to pick the odd one out among the set of words / tweets. More details in Appendix~\ref{sec:appendix_topic_eval}.

\begin{figure}[]
 \centering
   \includegraphics[width=\linewidth]{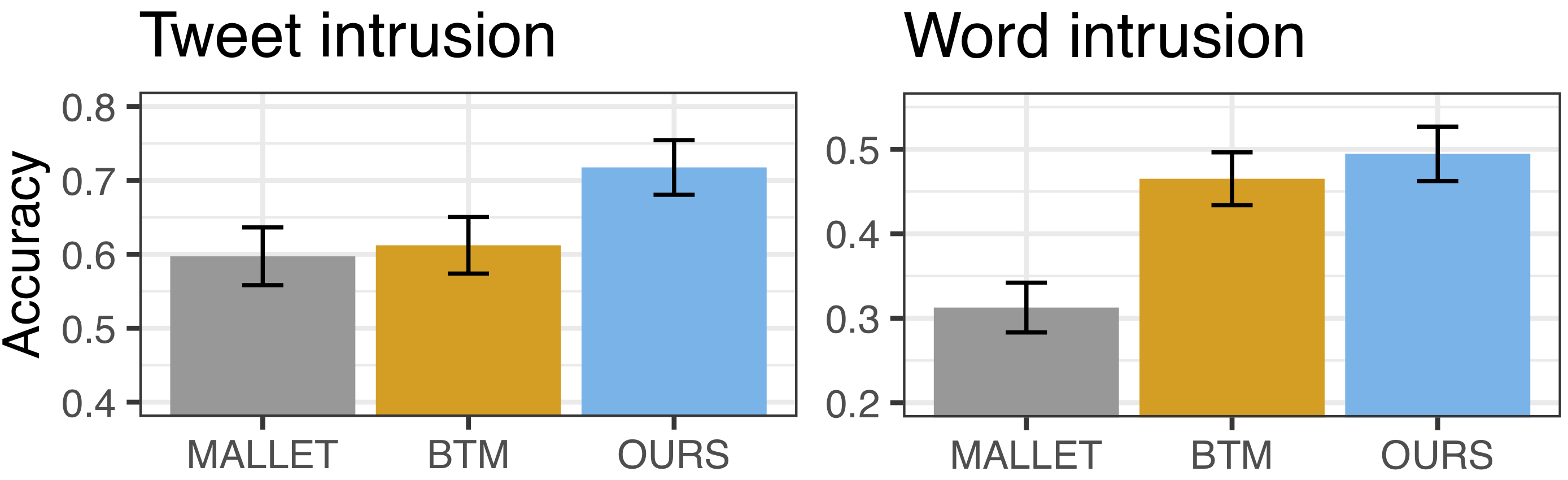}
   \caption{Topic model evaluations, collapsed across $k=6-10$. Error bars denote standard errors.}
   \label{fig:topic_eval}
\end{figure}

We find that our model outperforms the LDA-based methods with respect to both tasks, particularly tweet intrusion --- see Figure~\ref{fig:topic_eval}. This suggests that our model both provides more cohesive topics at the word level and more cohesive groupings by topic assignment. The choice of $k$ does not yield a significant difference among model-level accuracies.
However, since $k = 8$ slightly outperforms other $k$-s in tweet intrusion, we use it for further analysis. See Table~\ref{tab:topic_names} for nearest neighbor stems to each topic and Appendix~\ref{ssec:appendix_sample_topic_tweets} for example tweets.
 \begin{table}[]
\centering
\resizebox{\linewidth}{!}{%
\begin{tabular}{|l|l|}
\hline
\rowcolor[HTML]{C0C0C0} 
\textbf{Topic}                                                                  & \textbf{10 Nearest Stems}                                                                                                                \\ \hline
\begin{tabular}[c]{@{}l@{}}news\\ (19\%)\end{tabular}                           & \begin{tabular}[c]{@{}l@{}}break, custodi, \#breakingnew, \#updat, confirm,\\ fatal, multipl, updat, unconfirm, sever\end{tabular}      \\ \hline
\begin{tabular}[c]{@{}l@{}}investigation\\ (9\%)\end{tabular}                   & \begin{tabular}[c]{@{}l@{}}suspect, arrest, alleg, apprehend, custodi,\\ charg, accus, prosecutor, \#break, ap\end{tabular}             \\ \hline
\begin{tabular}[c]{@{}l@{}}shooter's identity\\ \& ideology (11\%)\end{tabular} & \begin{tabular}[c]{@{}l@{}}extremist, radic, racist, ideolog, label,\\ rhetor, wing, blm, islamist, christian\end{tabular}              \\ \hline
\begin{tabular}[c]{@{}l@{}}victims \& location\\ (4\%)\end{tabular}             & \begin{tabular}[c]{@{}l@{}}bar, thousand, california, calif, among,\\ los, southern, veteran, angel, via\end{tabular}                   \\ \hline
\begin{tabular}[c]{@{}l@{}}laws \& policy\\ (14\%)\end{tabular}                 & \begin{tabular}[c]{@{}l@{}}sensibl, regul, requir, access, abid, \#gunreformnow,\\ legisl, argument, allow, \#guncontolnow\end{tabular} \\ \hline
\begin{tabular}[c]{@{}l@{}}solidarity\\ (13\%)\end{tabular}                     & \begin{tabular}[c]{@{}l@{}}affect, senseless, ach, heart, heartbroken,\\ sadden, faculti, pray, \#prayer, deepest\end{tabular}          \\ \hline
\begin{tabular}[c]{@{}l@{}}remembrance\\ (6\%)\end{tabular}                     & \begin{tabular}[c]{@{}l@{}}honor, memori, tuesday, candlelight, flown,\\ vigil, gather, observ, honour, capitol\end{tabular}            \\ \hline
\begin{tabular}[c]{@{}l@{}}other \\ (23\%)\end{tabular}                         & \begin{tabular}[c]{@{}l@{}}dude, yeah, eat, huh, gonna, ain,\\ shit, ass, damn, guess\end{tabular}                                      \\ \hline
\end{tabular}%
}\vspace*{-4pt}
  \caption{Our eight topics (with their average proportions across events) and nearest-neighbor stem embeddings to the cluster centroids. Topic names were manually assigned based on inspecting the tweets.}
\label{tab:topic_names}
\end{table}

\paragraph{Measuring within-topic and between-topic partisanship.}

Recall that the leave-out estimator from Section~\ref{ssec:exp1_methods} provides a measure of partisanship. The information in a tweet, and thus partisanship, can be decomposed into \textit{which} topic is discussed, and \textit{how} it's discussed.

To measure \textit{within-topic} partisanship for a particular event, i.e. how a user discusses a given topic, we re-apply the leave-out estimator. For each topic, we calculate the partisanship using only tweets categorized to that topic. Then, overall within-topic partisanship for the event is the weighted mean of these values, with weights given by the proportion of tweets categorized to each topic within each event.

\textit{Between-topic} partisanship is defined as the expected posterior that an observer with a neutral prior would assign to a user's true party after learning only the topic --- but not the words --- of a user's tweet. We estimate this value by replacing each tweet with its assigned topic and applying the leave-out estimator to this data.

\subsection{Results}
\label{ssec:exp2_results}

\begin{figure}[]
 \centering
   \includegraphics[width=\linewidth]{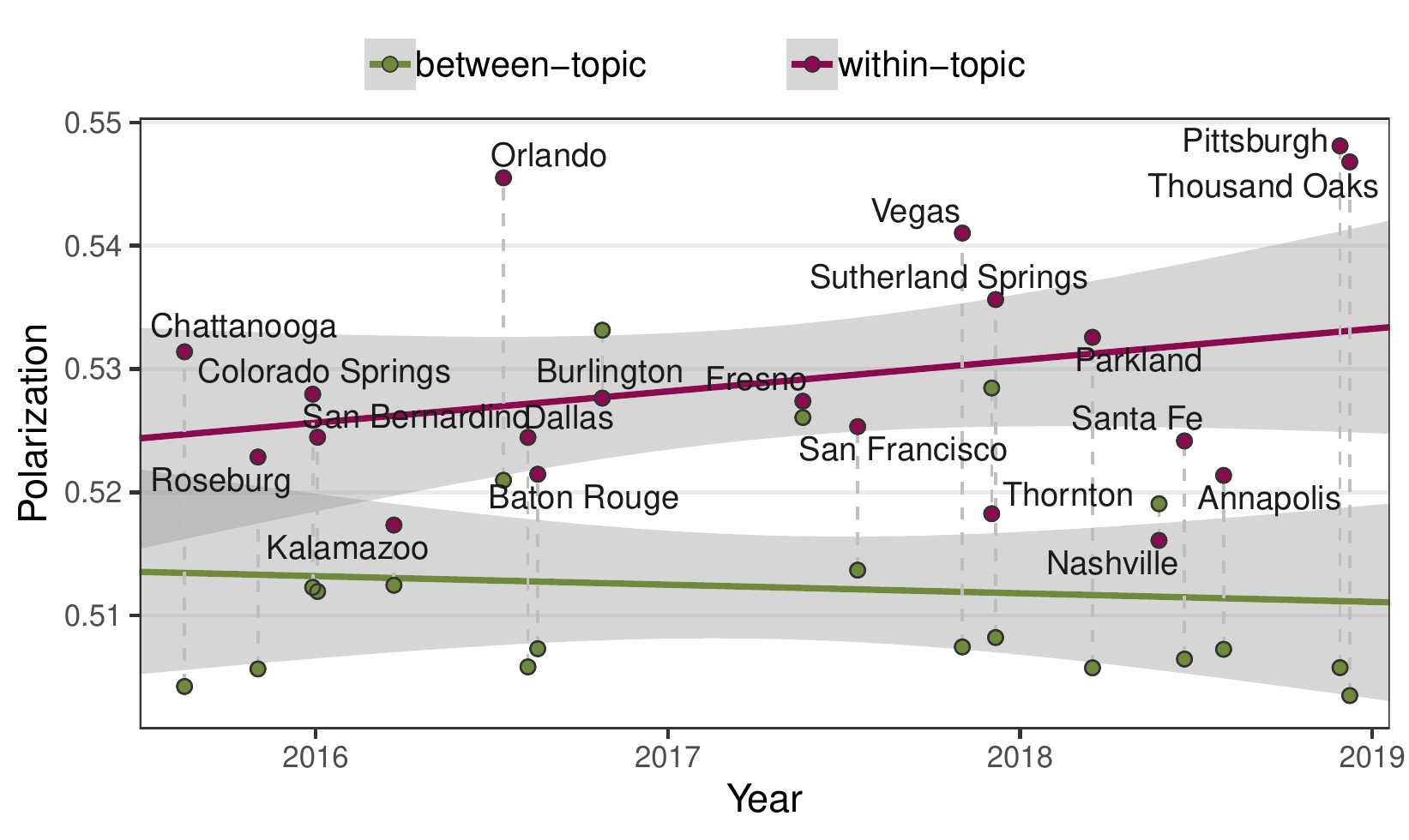}\vspace*{-5pt}
   \caption{Measurements of between-topic and within-topic polarization of the 21 events in our dataset show that within-topic polarization is increasing over time while between-topic polarization remains stable.}
   \label{fig:topic_polarization}
\end{figure}

Figure~\ref{fig:topic_polarization} shows that for most events within-topic is higher than between-topic partisanship, suggesting that while topic choice does play a role in phrase partisanship (its values are meaningfully higher than $.5$), within-topic phrase usage is significantly more polarized. Linear estimates of the relationship between within and between topic partisanship and time show that while within-topic polarization has increased over time, between-topic polarization has remained stable. This finding supports the idea that topic choice and topic-level framing are distinct phenomena.

\begin{figure}[]
   \includegraphics[width=1.01\linewidth]{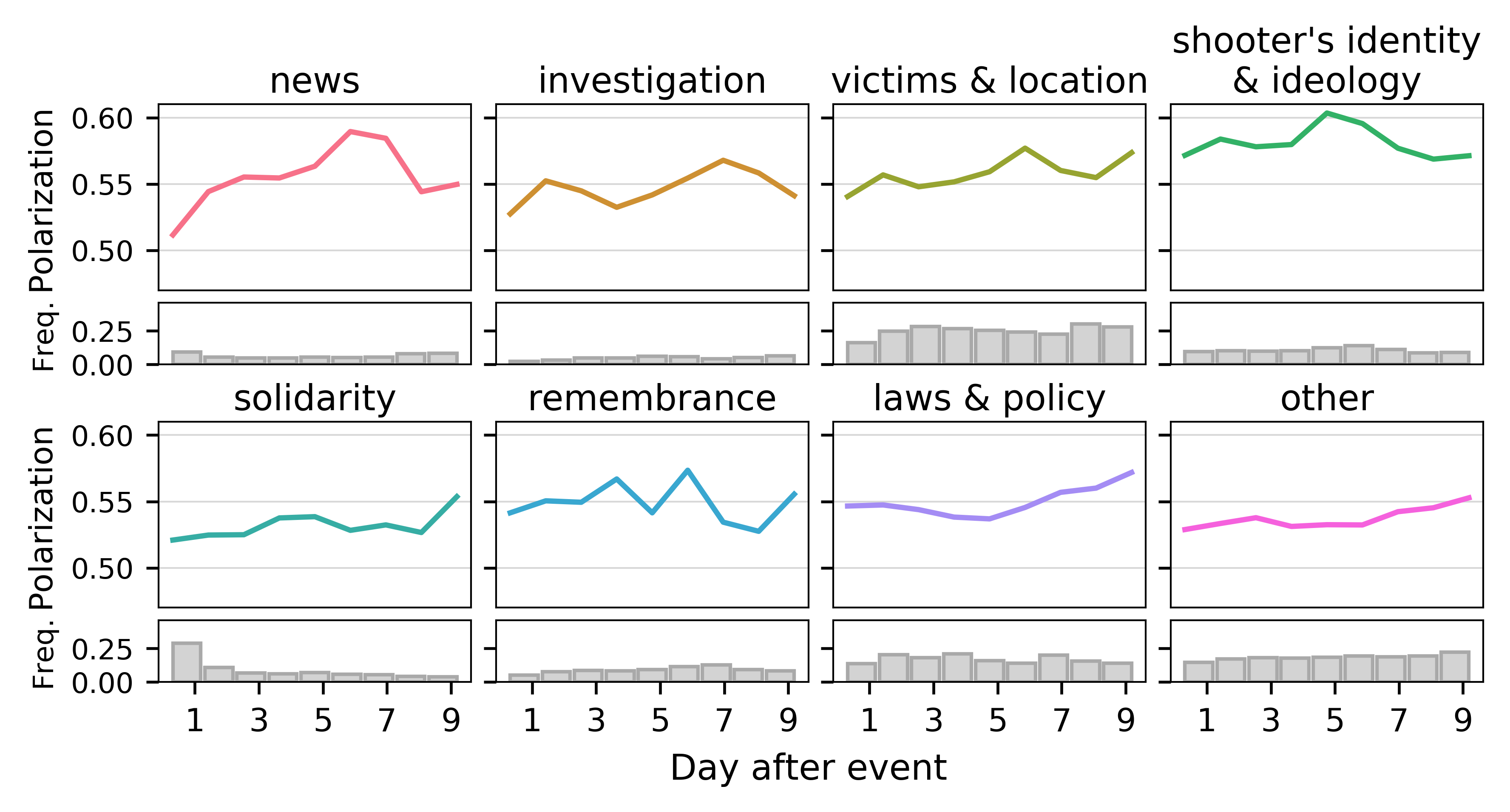}
   \caption{Las Vegas within-topic polarization in the days after the event. The bar charts show the proportion of each topic in the data at a given time.
}
   \label{fig:topic_overtime}
\end{figure}

Partisanship also differs by topic, and within days after a given event. Figure~\ref{fig:topic_overtime} shows polarization within topics for 9 days after Las Vegas. We find that \emph{solidarity} has the lowest and \emph{shooter's identity \& ideology} the highest polarization throughout; polarization in most topics increases over time and \emph{news} has the steepest increase. Similar patterns are present after Orlando (Figure~\ref{fig:orlando_topic_overtime} in Appendix~\ref{sec:appendix_plots}). Measuring polarization of topics for other events over time is noisy, given the sparsity of the data, but overall within-topic polarization is consistent: the most polarized topics on average across events are \emph{shooter's identity \& ideology} ($.55$) and \emph{laws \& policy} ($.54$), where people are apparently polarized about both why an event happened and what to do about it. Fact- and sympathy-based topics display less polarization: \emph{news} ($.51$), \emph{victims \& location} ($.52$), \emph{solidarity} ($.52$) and \emph{remembrance} ($.52$). 

\begin{figure}[]
   \includegraphics[width=\linewidth]{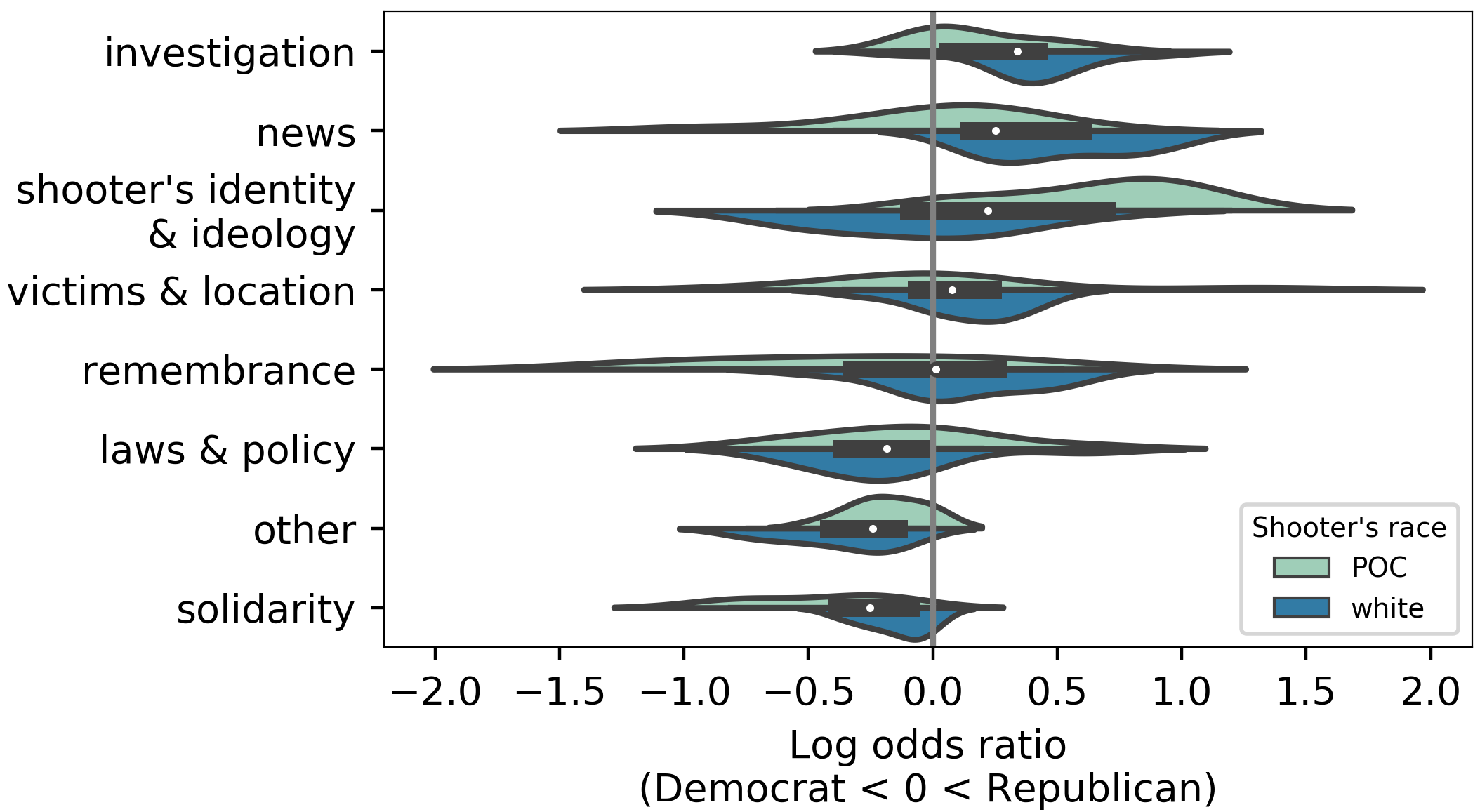}
   \caption{The plot shows the kernel density of the partisan log odds ratios of each topic (one observation per event). The white points show the median and the black rectangles the interquartile range across events.}
   \label{fig:topic_log_odds}
\end{figure}

As shown in Figure~\ref{fig:topic_log_odds}, \emph{investigation}, \emph{news}, and \emph{shooter's identity \& ideology} are more likely to be discussed by Republicans and \emph{laws \& policy} and \emph{solidarity}   more likely to be discussed by Democrats across events.\footnote{$p$-values are calculated using a one sample t-test, comparing with zero: \emph{shooter's identity \& ideology} ($p < 0.05$), \emph{investigation} ($p < 0.001$), \emph{laws \& policy} ($p < 0.1$), \emph{news} ($p < 0.05$), \emph{solidarity} ($p < 0.001$).} Topics preferred by Republicans seem to relate more to the shooter than to the victims, while topics preferred by Democrats seem to relate more closely to the victims. 
The shooter's race appears to play a role in topic preference: if the shooter is white, Democrats become more likely to focus on \emph{shooter's identity \& ideology} and \emph{laws \& policy} and Republicans on \emph{news} and \emph{investigation} than if the shooter is a person of color. 


\section{Specific Framing Devices}
In the previous section, we show that topic-level framing and topic choice are different dimensions of polarization. We now look at the specific terms and types of grounding that are used as partisan framing devices, contributing to polarization.

\subsection{Methods}
\label{ssec:exp3_methods}
 
 \paragraph{Partisan tokens.} We estimate the partisanship of tokens via their event-level log odds ratio of Democrats relative to Republicans (based on the vocabularies we create in Section~\ref{ssec:exp1_methods}). We compare these estimates across events.\footnote{To compare the partisanship of tokens at the event- and topic-level, we also $z$-score the log odds ratios \cite{monroe2008fightin} within events and topics; the most partisan tokens are reported in Appendix~\ref{sec:appendix_partisan_words}. The reason why we do not $z$-score for the between-event comparison is because data verbosity disproportionately affects the range of the values' magnitudes. Note that the signs of values --- which we focus on for the cross-event comparison --- are not affected by $z$-scoring.}
 
\paragraph{Grounding.} We study whether there is polarization in which prior tragic events are referenced in the context of a particular mass shooting. We compile a list of keywords representing major events of mass violence in the US in the past two decades and kept those that were mentioned at least 100 times by Democrat or Republican users. For all tweets for each event in our dataset, we counted the mentions of past \emph{context events}. For example, in the following tweet posted after Las Vegas: ``Dozens of preventable deaths should not be the cost of living in America. Stand up to the \#NRA. \#LasVegasShooting \#SandyHook \#Charleston'', Sandy Hook and Charleston are the context events. Finally, we calculated the partisan log odds ratio of each context event.

\subsection{Results}
\label{ssec:exp3_results}
\begin{figure}[]
 \centering
   \centering
   \includegraphics[width=\linewidth]{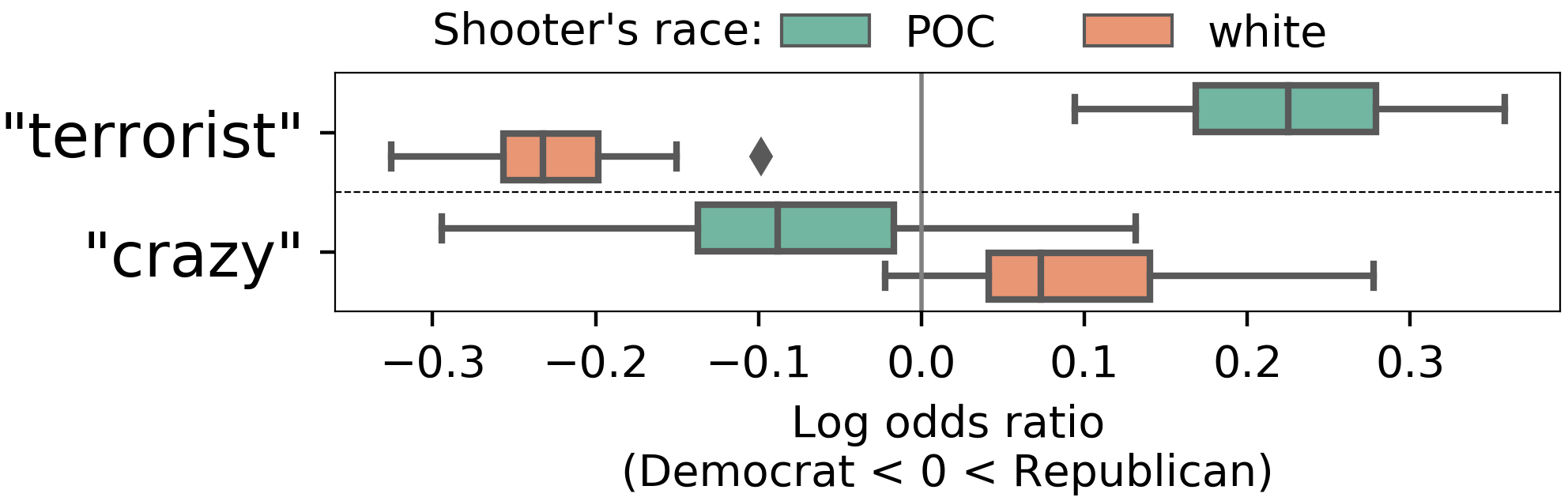}\vspace*{-3pt}
   \caption{The log odds ratios of ``terrorist'' and ``crazy'' across events, grouped by the shooter's race. The boxes show the interquartile range and the diamond an outlier.}
   \label{fig:terrorist_log_odds}
\end{figure}
We focus on the partisanship of the term ``terrorist'' and ``crazy'', which exhibit differential patterns across events based on the shooter's race.\footnote{Note that these words in fact have the \textit{largest} difference (negative and positive, respectively) if we calculate the differences between the mean $z$-scores --- grouped by the shooter's race --- for all tokens in our joint vocabulary.} ``Terrorist'' is \emph{always} more likely to be used by Democrats than Republicans in events where the shooter is white, and the opposite is true when the shooter is a person of color (Figure~\ref{fig:terrorist_log_odds}); ``crazy'' is more likely used by Republicans if the shooter is white than if they are a person of color and the opposite is true (although the pattern is weaker) when a shooter is white.


These findings support related work \cite{perea1998black, delgado2017critical} discussing binary conceptualization of race in the US, and its influence in determining whether a shooter's mental health or aspects of their identity are discussed. However, the fact that the influence of race flips completely for Democrats and Republicans is a striking result that calls for further exploration.

\begin{figure*}[]
 \centering
   \centering
   \includegraphics[width=.85\linewidth]{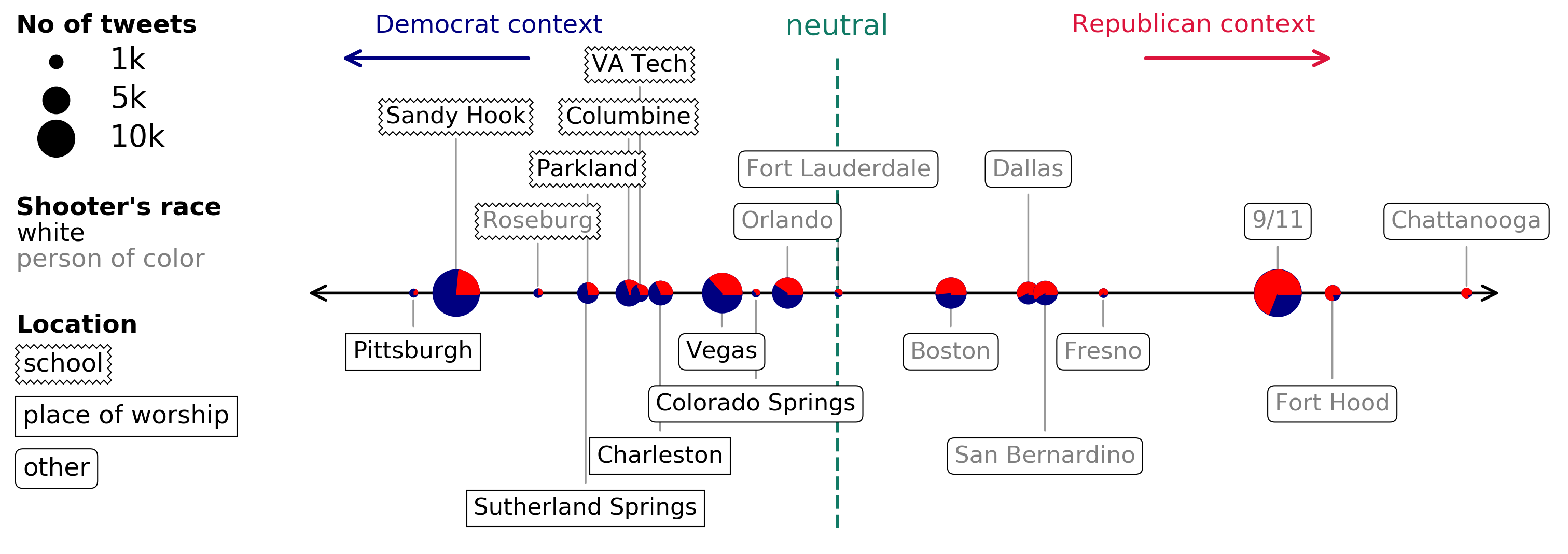}
   \caption{The partisanship of events of mass violence when used as a context for a given mass shooting. The position of the events on the line represents their partisan log odds ratio (Democrat $<$ 0 (neutral) $<$ Republican). The pie charts indicate the proportion of \textcolor{navy}{Democrat} and \textcolor{crimson}{Republican} users' tweets that involve this ``context'' event.}
   \label{fig:event_log_odds}
\end{figure*}

The partisanship of contextual grounding also corroborates our finding that the shooter's race influences how people conceptualize a certain event. Our results in Figure~\ref{fig:event_log_odds} suggest a few key take-aways: the two most frequently employed context events are both highly partisan (Sandy Hook for Democrats and 9/11 for Republicans); shootings at schools and places of worship are more likely to be brought up by Democrats; Democrats are more likely to reference events with white shooters, while Republicans are more likely to reference those with shooters who are people of color.

\section{Affect}
\label{sec:exp3_other_aspects}
Affect is intimately tied to ideological reasoning \cite{redlawsk2002hot,taber2009motivated}, and so emotional expression represents another semantic layer relevant to polarization \cite{iyengar2012affect,suhay2015explaining}. Others have shown that emotion words can help detect political ideology on Twitter \cite{preoctiuc2017beyond} and that emotive political tweets are more likely to be shared \cite{stieglitz2012political}.
Here, we employ a lexicon-based approach to measure valence (positive and negative) and five basic emotion categories (disgust, fear, trust, anger, and sadness).

\subsection{Methods}
\label{ssec:exp4_methods}

Since word-affect associations are highly domain dependent, we tailored an existing affect lexicon, the NRC Emotion Lexicon \cite{mohammad2013nrc}, to our domain via label propagation \cite{hamilton2016inducing}. 

Specifically, we stem all the words in the lexicon and select 8-10 representative stems per emotion category that have an association with that emotion in the context of mass shootings.
For each emotion category, we compute pairwise cosine distances between the GloVe embedding of each in-vocabulary stem and the representative stems for that emotion, and include the 30 stems with the lowest mean cosine distances. The resulting lexicons can be found in Appendix~\ref{sec:appendix_emotion}.

We use these lexicons to measure the partisanship of each affect category. For each event and each party we aggregate stem frequencies per emotion category. We then calculate the partisan log odds ratio of each category for each event.

\subsection{Results}
\label{ssec:exp4_results}
\begin{figure}[h]
 \centering
   \centering
   \includegraphics[width=\linewidth]{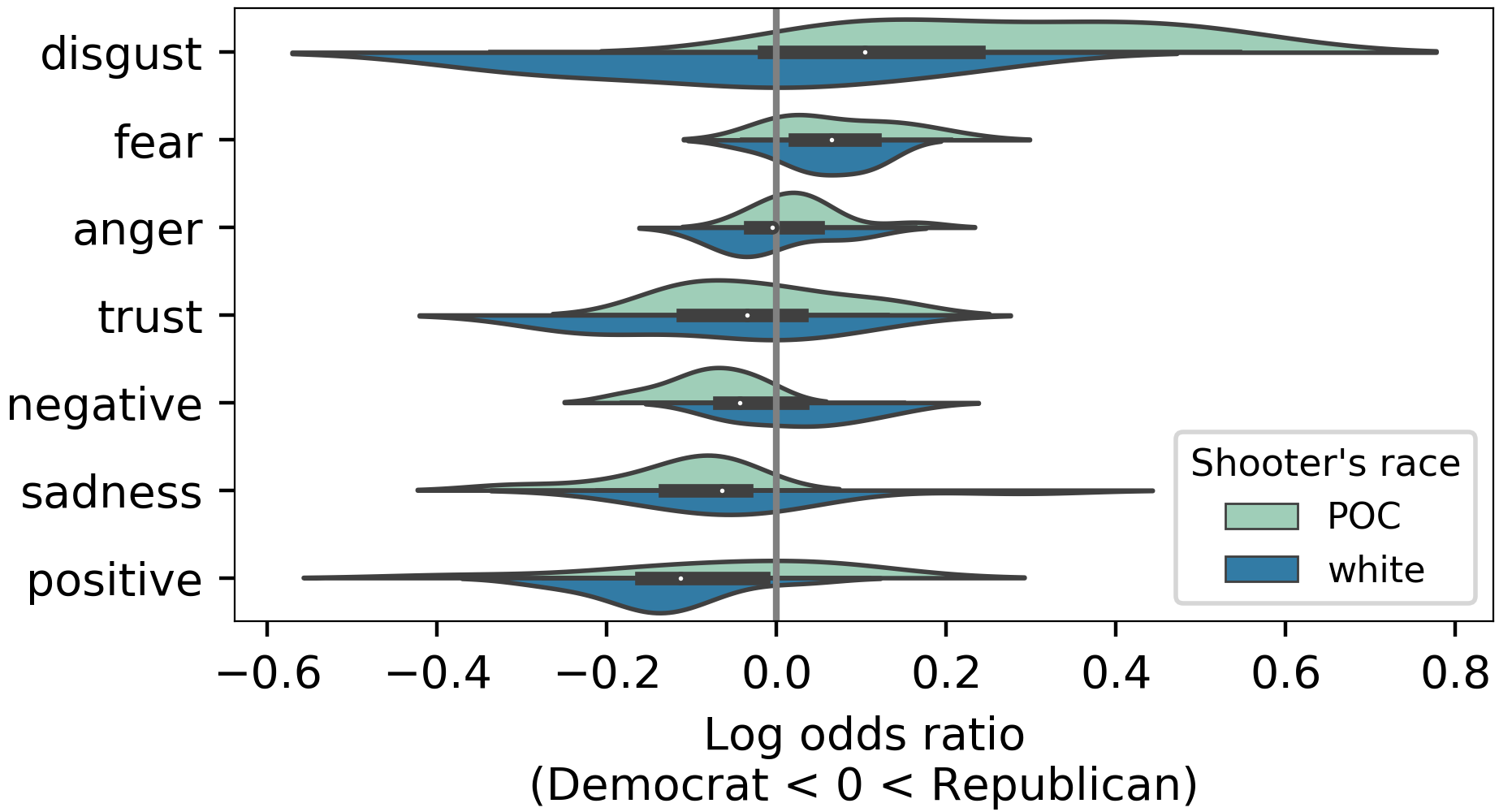}
   \caption{The log odds of each emotion category in our lexicon (one observation represents one event). }
   \label{fig:affect_log_odds}
\end{figure}

The log odds ratio of each affect category is shown in Figure~\ref{fig:affect_log_odds}. These findings suggest that positive sentiment, sadness and trust are more likely to be expressed by Democrats across events, while fear and disgust are more likely to be expressed by Republicans, particularly when the shooter is a person of color. Anger, trust and negative sentiment is similarly likely to be expressed by both parties.\footnote{$p$-values are calculated using a one sample t-test, comparing to zero: anger ($p\approx0.43$), disgust ($p\approx0.06$), fear ($p<0.001$), negative ($p\approx.2$), positive ($p<0.001$), sadness ($p<0.02$), trust ($p\approx0.07$).}

Our results about fear and disgust accord with existing literature on emotion and political ideology: conservatives score higher than liberals on subjective measures of fear  \citep[e.g.][]{jost2017politics, federico2009political, hibbing2014differences} and disgust sensitivity is also associated with 
political conservativism \citep[e.g.][]{inbar2009conservatives, inbar2012disgust}. 


\section{Modality and Illocutionary  Force} 
\label{ssec:modality}
\begin{table}[]
\centering
\resizebox{\linewidth}{!}{%
\begin{tabular}{|l|}
\hline
\begin{tabular}[c]{@{}l@{}}This roller coaster debate \textbf{MUST} STOP! Sensible gun ownership is one\\thing but assault weapons massacre innocent lives. The savagery of gore\\at \#Parkland was beyond belief \& \textbf{must} be the last.\end{tabular} \\ \hline
\begin{tabular}[c]{@{}l@{}}In times of tragedy \textbf{shouldn't} we all come together?! Prayers for those\\ harmed in the \#PlannedParenthood shooting.\end{tabular}  \\ \hline
\begin{tabular}[c]{@{}l@{}}Communities \textbf{need to} step up and address white on white crime like the\\ Las Vegas massacre. White men are out of control.\end{tabular}     \\ \hline
\begin{tabular}[c]{@{}l@{}}he BLM protest shooting, planned parenthood, now cali... domestic\\terrorism will crumble this country, SANE PPL \textbf{HAVE TO} FIGHT BACK\end{tabular}  \\ \hline
\begin{tabular}[c]{@{}l@{}}Shooting cops is horrible, cannot be condoned. But \textbf{must be} understood\\ these incidents are outgrowth of decades of police abuses. \#BatonRouge\end{tabular} \\ \hline
\begin{tabular}[c]{@{}l@{}}1. Islamic terrorists are at war with us 2. Gun free zones = kill zones \\3. Americans \textbf{should be} allowed to defend themselves \#Chattanooga\end{tabular}  \\ \hline
\begin{tabular}[c]{@{}l@{}}Las Vegas shooting Walmart shooting and now 25 people killed in\\ Texas over 90 people killed Mexico \textbf{should} build that wall to keep the US out\end{tabular} \\ \hline
\begin{tabular}[c]{@{}l@{}}CNN reporting 20 dead, 42 injured in Orlando night club shooting.\\ Just awful. The US \textbf{must} act to control guns or this carnage will continue. \end{tabular} \\\hline   \end{tabular}%
}
  \caption{Random sample of tweets with modals. Only one of the eight (Ex. 6) tweets expresses ideas traditionally associated with conservative ideology.}
\label{tab:modal_ex}
\end{table} 
Modality is a lexical category concerned with necessity and possibility \cite{kratzer2002notional, von2006modality}. In the aftermath of a tragic event, people seek solutions, a process that often involves reflecting on what \emph{should have} happened or \emph{should} happen now or in the future (e.g. to prevent such events). We hypothesize that the use of modals in our data gives insight into the kind of (illocutionary) acts \cite{austin1962things} the users are performing via their tweets, such as calling for action, assigning blame, expressing emotions, and stating facts.

\subsection{Methods}

We work with all forms of the four most frequent necessity modals in our data --- \emph{should, must, have to} and \emph{need to}. For each, we quantify its partisanship via its partisan log odds ratio. We also annotate a random sample of 200 tweets containing modals to see whether they are indeed used in contexts that imply calls for change / action (e.g. `We must have gun control!') and / or to express the user's mental state about the event, such as despair or disbelief (e.g. `Why do people have to die?').

\subsection{Results}
\label{ssec:exp5_results}
\begin{figure}[htb]
 \centering
   \centering
   \includegraphics[width=\linewidth]{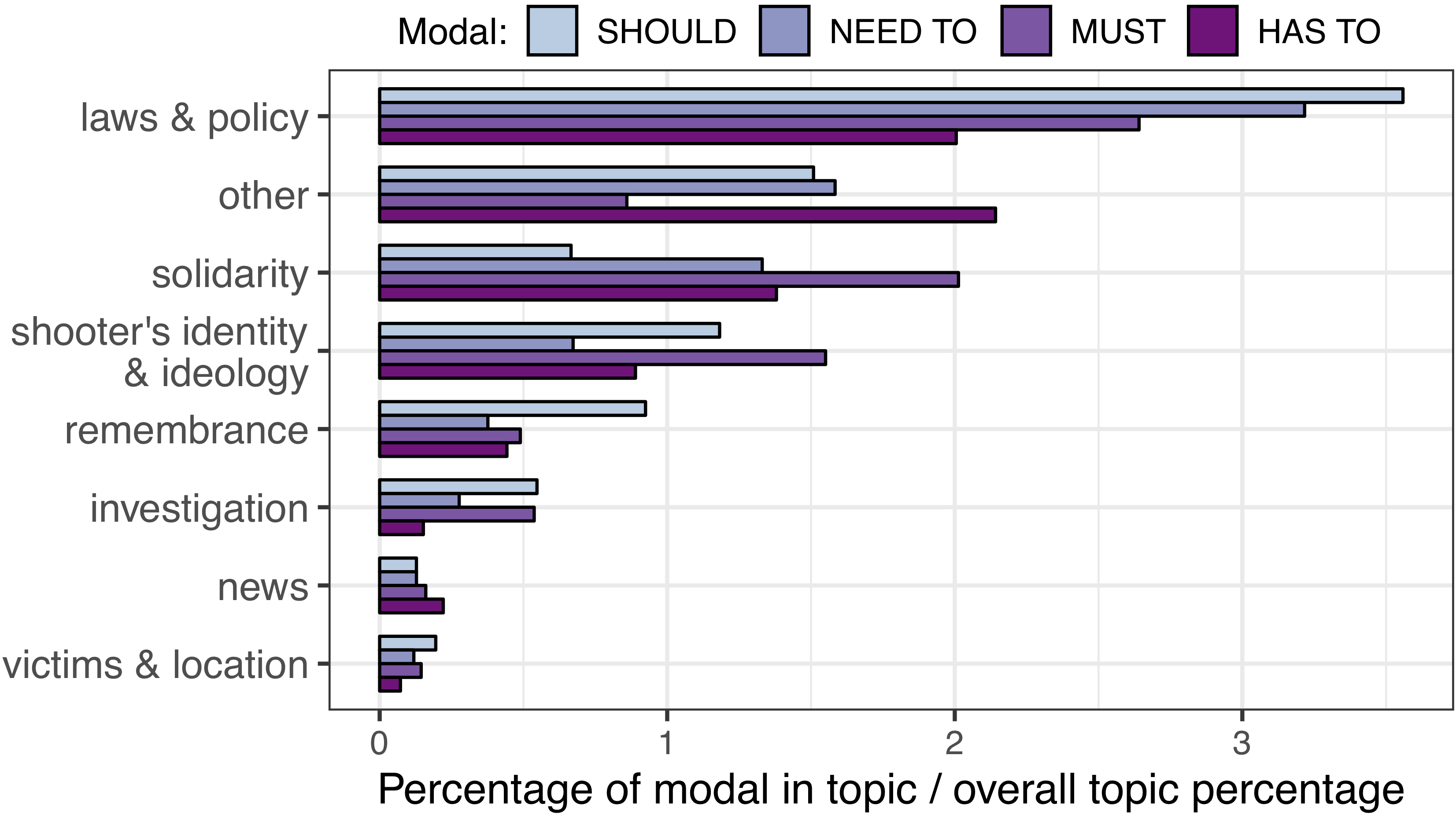}
   \caption{The representation of modals in each topic. Values represent averages across events.}
   \label{fig:modal_distr}
\end{figure}

Table~\ref{tab:modal_ex} shows a random sample of tweets containing some form of either \emph{should, must, have to}, or \emph{need to}. More collocations, as well as their partisanship, can be found in Appendix~\ref{sec:appendix_modals}. These examples, as well as our  annotation, support the hypothesis that these modals are primarily used to call for action. Of the 200 modal uses, $\sim$78\% express calls for change/action, $\sim$40\% express the user's mental state.\footnote{Other uses are primarily epistemic ones (e.g. `The suspect must be mentally ill').} We also compute the representation $p_x^m$ of each modal $m$ in each topic $x\in X$ via $(f_x^m /\sum_{x'\in X} f_{x'}^m )/(f_x /\sum_{{x'}\in X} f_{x'})$, where $f_x$ is the number of tweets from topic $x$, and $f_x^m$ the number of those also containing $m$. We find that that modals are over-represented in the \emph{laws \& policy} topic (see  Figure~\ref{fig:modal_distr}). This evidence suggests that calls for policy change --- especially gun control, based on annotated samples --- are a dominant subset of calls for action. 

The log odds ratio of modals shows that \emph{all} of them are more likely to be used by Democrats across events: \emph{have to} (mean:$-.39$, $p < 0.001$), \emph{must} (mean:$-.3$, $p < 0.001$), \emph{should} (mean:$-.18$, $p < 0.01$), \emph{need to} (mean:$-.18$, $p < 0.01$) --- where Democrat and Republican log odds are negative and positive, respectively.\footnote{$p$-values are from one sample t-test, comparing to $0$.} A two-tailed t-test shows that only \emph{should} exhibits statistically significant difference based on the shooter's race ($p<0.03$), as it is even more likely to be used by Democrats when the shooter is white.

To understand whether assigning blame in this domain is a partisan propensity, we also study uses of \emph{should have}.\footnote{On average across events, forms of \emph{should have} constitute $16\%$ of the uses of \emph{should} (SD=$5\%$).} The log odds of \emph{should have} (mean:$-.22$, $p<0.05$) show that it is similarly likely to be used by Democrats as \emph{should} ($p\approx0.8$ from two-tailed t-test). Interestingly the log odds ratio of \emph{should have}, unlike that of \emph{should}, does not differ significantly based on the shooter's race ($p\approx0.8$ from two-tailed t-test). Moreover, we did not find a significant difference in the partisanship of \emph{should have} nor any other modal based on the administration (Obama or Trump) a shooting took place under, suggesting that Democrats are more likely call for change and assign blame even if their preferred party is in power.



\section{Conclusion}

We show that inspecting polarization on social media from various angles can shed light on salient phenomena pertinent to group divisions. Applying the leave-out estimator of phrase partisanship to data on mass shootings, we find that reactions to these events are highly polarized politically.

To disentangle topic choice and topic-level framing --- two phenomena that contribute to polarization --- we introduce a tweet-clustering approach. By sampling, requiring words in the vocabulary to appear in multiple events and relying on the abstraction of a vector space model, we generate cohesive topic representations that are robust to disparities among event-level vocabularies and tweet counts. Human evaluation shows that our method outperforms LDA-based approaches.

Our induced topics suggest that Republicans preferentially discuss topics about the shooter's identity and ideology, investigation and news, while Democrats preferentially discuss solidarity and policy-related topics. We also find that the setting and the shooter's race interact with polarization. For example, Democrats are more likely to contextualize any mass shooting among school shootings and call white shooters ``terrorists'' than are Republicans, who in turn are more likely to liken any shooting to other violent events perpetrated by people of color --- whom they are more likely to call ``terrorist'' than are Democrats. Moreover, Democrats are more likely to frame the shooter as mentally ill when they are a person of color and Republicans when they are white.

We also demonstrate that looking at affect and illocutionary force can help us understand users' polarized responses to these tragic events: Republicans are more likely to express fear and disgust than are Democrats, while Democrats are more likely to express sadness and positive sentiment, to make calls for action and assign blame.

Polarization is a multi-faceted phenomenon: in this paper we present a set of measures to study these different facets through the lens of language. We show that these measures provide convergent evidence, creating a clearer picture of the complex ideological division permeating public life.









\paragraph*{Acknowledgements.} We thank Jure Leskovec and Adrijan Bradaschia for data, and Cleo Condoravdi, Chris Potts, Linda Ouyang, David Ritzwoller and Frank Yang for helpful feedback. We are grateful for the support of the Stanford Cyber Initiative, the Melvin and Joan Lane Stanford Graduate Fellowship (to D.D.), NSF GRF DGE-114747 (to N.G.), the Michelle and Kevin Douglas Stanford Interdisciplinary Graduate Fellowship (to R.V.), NSF grant CRII 1657155 (to J.Z.), the Stanford Institute for Economic Policy Research and the Knight Foundation (to M.G.) and the Brown University Population Studies and Training Center (to J.S.).

\bibliography{main}
\bibliographystyle{acl_natbib}

\clearpage

\appendix

\section{Data}
\label{sec:appendix_data}

\begin{figure*}[htbp]
 \centering
   \centering
   \includegraphics[width=.8\linewidth]{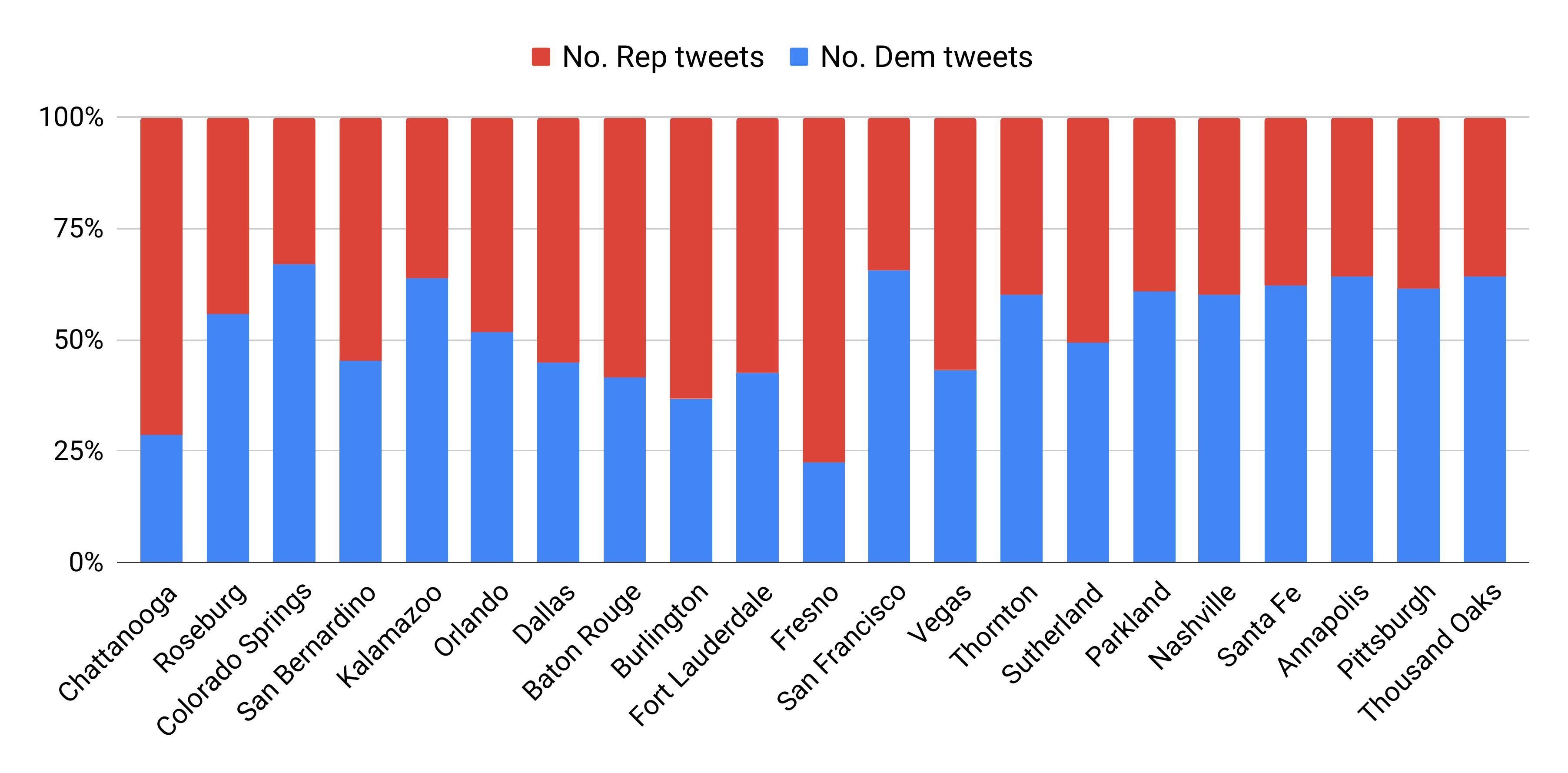}
   \caption{Distribution of partisan tweets for each event.}
   \label{fig:party_distr}
\end{figure*}

Table~\ref{tab:data} contains properties of the data. Figure~\ref{fig:party_distr} contains the distribution of partisan tweets for each event.

\paragraph{Event-specific keywords.} We use the following location-specific keywords (case insensitive) to find tweets on the events:
\begin{itemize}
    \item Chattanooga: chattanooga, military recruitment center
    \item Roseburg: umpqua, roseburg
    \item Colorado Springs: colorado springs, coloradosprings, planned parenthood, plannedparenthood
    \item San Bernardino: san bernardino, sanbernardino
    \item Kalamazoo: kalamazoo
    \item Orlando: orlando, pulse nightclub
    \item Dallas: dallas
    \item Baton Rouge: baton rouge, batonrouge
    \item Burlington: burlington, cascade mall
    \item Fort Lauderdale: lauderdale
    \item Fresno: fresno
    \item San Francisco: ups, san francisco
    \item Las Vegas: vegas, route91, harvest festival, harvestfestival, mandalay bay
    \item Thornton: thornton, walmart, denver
    \item Sutherland Springs: sutherland springs, sutherlandsprings
    \item Parkland: parkland, marjory stoneman
    \item Nashville: nashville, waffle house
    \item Santa Fe: santa fe, santafe
    \item Annapolis: annapolis, capital gazette
    \item Pittsburgh: pittsburgh, treeoflife, tree of life
    \item Thousand Oaks: thousand oaks, thousandoaks
\end{itemize}

\begin{table*}
\resizebox{\textwidth}{!}{%
\begin{tabular}{|l|l|l|l|l|l|l|l|l|l|}
\hline
\textbf{Event city / town} & \textbf{State} & \textbf{Specific location}                                                     & \textbf{Date} & \textbf{No. victims} & \textbf{\begin{tabular}[c]{@{}l@{}}Race  / ethnicity \\ of shooter\end{tabular}} & \textbf{No. tweets} & \textbf{No. partisan tweets} & \textbf{No. Dem tweets} & \textbf{No. Rep tweets} \\ \hline
Chattanooga                & TN             & Military Recruitment Center                                                    & 7/16/15       & 7                    & Middle Eastern                                                                   & 29573               & 20709                        & 5925                    & 14784                   \\ \hline
Roseburg                   & OR             & Umpqua Community College                                                       & 10/1/15       & 18                   & Mixed                                                                            & 18076               & 11505                        & 6419                    & 5086                    \\ \hline
Colorado Springs           & CO             & Planned Parenthood clinic                                                      & 11/27/15      & 12                   & White                                                                            & 55843               & 39719                        & 26614                   & 13105                   \\ \hline
San Bernardino             & CA             & Inland Regional Center                                                         & 12/2/15       & 35                   & Middle Eastern                                                                   & 70491               & 45819                        & 20798                   & 25021                   \\ \hline
Kalamazoo                  & MI             & multiple                                                                       & 2/20/16       & 8                    & White                                                                            & 10986               & 6807                         & 4350                    & 2457                    \\ \hline
Orlando                    & FL             & Pulse nightclub                                                                & 6/12/16       & 102                  & Middle Eastern                                                                   & 1831082             & 872022                       & 450784                  & 421238                  \\ \hline
Dallas                     & TX             & Black Lives Matter protest                                                     & 7/7/16        & 16                   & Black                                                                            & 260377              & 144205                       & 64628                   & 79577                   \\ \hline
Baton Rouge                & LA             & streets                                                                        & 7/17/16       & 6                    & Black                                                                            & 46126               & 29015                        & 12019                   & 16996                   \\ \hline
Burlington                 & WA             & Cascade Mall                                                                   & 9/23/16       & 5                    & Middle Eastern                                                                   & 8171                & 4993                         & 1838                    & 3155                    \\ \hline
Fort Lauderdale            & FL             & Fort Lauderdale airport                                                        & 1/6/17        & 11                   & Hispanic                                                                         & 12525               & 7194                         & 3073                    & 4121                    \\ \hline
Fresno                     & CA             & downtown                                                                       & 4/18/17       & 3                    & Black                                                                            & 8868                & 6128                         & 1377                    & 4751                    \\ \hline
San Francisco              & CA             & UPS store                                                                      & 6/14/17       & 5                    & Asian                                                                            & 10487               & 6627                         & 4346                    & 2281                    \\ \hline
Vegas                      & NV             & Route 91 Harvest Festival                                                      & 10/1/17       & 604                  & White                                                                            & 1286399             & 726739                       & 315343                  & 411396                  \\ \hline
Thornton                   & CO             & Walmart                                                                        & 11/1/17       & 3                    & White                                                                            & 14341               & 9170                         & 5527                    & 3643                    \\ \hline
Sutherland Springs         & TX             & Texas First Baptist Church                                                     & 11/5/17       & 46                   & White                                                                            & 154076              & 106220                       & 52513                   & 53707                   \\ \hline
Parkland                   & FL             & \begin{tabular}[c]{@{}l@{}}Marjory Stoneman Douglas\\ High School\end{tabular} & 2/14/18       & 31                   & White                                                                            & 272499              & 186570                       & 113856                  & 72714                   \\ \hline
Nashville                  & TN             & Waffle House                                                                   & 4/22/18       & 8                    & White                                                                            & 38680               & 24326                        & 14606                   & 9720                    \\ \hline
Santa Fe                   & CA             & Santa Fe High School                                                           & 5/18/18       & 23                   & White                                                                            & 73621               & 42968                        & 26784                   & 16184                   \\ \hline
Annapolis                  & MD             & Capital Gazette                                                                & 6/28/18       & 7                    & White                                                                            & 27715               & 18468                        & 11863                   & 6605                    \\ \hline
Pittsburgh                 & PA             & Tree of Life Synagogue                                                         & 10/27/18      & 18                   & White                                                                            & 59925               & 36920                        & 22735                   & 14185                   \\ \hline
Thousand Oaks              & CA             & Borderline Bar and Grill                                                       & 11/7/18       & 23                   & White                                                                            & 117815              & 62812                        & 40328                   & 22484                   \\ \hline
\end{tabular}%
}

\caption{Data properties.}
\label{tab:data}
\end{table*}

\clearpage\subsection{Stopwords}
\label{ssec:appendix_stopwords}
no, noone, nobody, nowhere, nothing, nor, not, none, non, a, able, about, above, according, accordingly, across, actually, after, afterwards, again, against, all, allow, allows, almost, alone, along, already, also, although, always, am, among, amongst, an, and, another, any, anybody, anyhow, anyone, anything, anyway, anyways, anywhere, apart, appear, appreciate, appropriate, are, aren, around, as, aside, ask, asking, associated, at, available, away, awfully, b, be, became, because, become, becomes, becoming, been, before, beforehand, behind, being, believe, below, beside, besides, best, better, between, beyond, both, brief, but, by, c, came, can, cannot, cant, cause, causes, certain, certainly, changes, clearly, co, com, come, comes, concerning, consequently, consider, considering, contain, containing, contains, corresponding, could, couldnt, couldn, couldve, course, currently, d, definitely, described, despite, did, didnt, didn, different, do, dont, don, does, doesn, doesnt, doing, done, down, downwards, during, e, each, edu, eg, eight, either, else, elsewhere, enough, entirely, especially, et, etc, even, ever, every, everybody, everyone, everything, everywhere, ex, exactly, example, except, f, far, few, fifth, first, five, followed, following, follows, for, former, formerly, forth, four, from, further, furthermore, g, get, gets, getting, given, gives, go, goes, going, gone, got, gotten, greetings, h, had, hadn, happens, hardly, has, hasnt, hasn, have, havent, haven, having, he, hes, hell, hello, help, hence, her, here, hereafter, hereby, herein, hereupon, hers, herself, hi, him, himself, his, hither, hopefully, how, howbeit, however, i, im, ive, ie, if, i, ignored, immediate, in, inasmuch, inc, indeed, indicate, indicated, indicates, inner, insofar, instead, into, inward, is, isn, it, its, itself, j, just, k, keep, keeps, kept, know, knows, known, l, last, lately, later, latter, latterly, least, less, lest, let, like, liked, likely, little, ll, look, looking, looks, ltd, m, mainly, many, may, maybe, me, mean, meanwhile, merely, might, mightve, more, moreover, most, mostly, much, must, mustn, mustnt, mustve, my, myself, n, name, namely, nd, near, nearly, necessary, need, neednt, needs, neither, never, nevertheless, new, next, nine, nor, normally, novel, now, o, obviously, of, off, often, oh, ok, okay, old, on, once, one, ones, only, onto, or, other, others, otherwise, ought, our, ours, ourselves, out, outside, over, overall, own, p, particular, particularly, per, perhaps, placed, please, plus, possible, presumably, probably, provides, q, que, quite, qv, r, rather, rd, re, really, reasonably, regarding, regardless, regards, relatively, respectively, right, s, said, same, saw, say, saying, says, second, secondly, see, seeing, seem, seemed, seeming, seems, seen, self, selves, sensible, sent, serious, seriously, seven, several, shall, she, shell, shes, should, shouldnt, shouldn, shouldve, since, six, so, some, somebody, somehow, someone, something, sometime, sometimes, somewhat, somewhere, soon, sorry, specified, specify, specifying, still, sub, such, sup, sure, t, take, taken, tell, tends, th, than, thank, thanks, thanx, that, thats, the, their, theirs, them, themselves, then, thence, there, thereafter, thereby, therefore, therein, theres, thereupon, these, they, theyre, theyve, think, third, this, thorough, thoroughly, those, though, three, through, throughout, thru, thus, to, together, too, took, toward, towards, tried, tries, truly, try, trying, twice, two, u, un, under, unfortunately, unless, unlikely, until, unto, up, upon, us, use, used, useful, uses, using, usually, uucp, v, value, various, ve, very, via, viz, vs, w, want, wants, was, way, we, welcome, well, went, were, what, whatever, when, whence, whenever, where, whereafter, whereas, whereby, wherein, whereupon, wherever, whether, which, while, whither, who, whoever, whole, whom, whose, why, will, willing, wish, with, within, without, wonder, would, wouldn, wouldnt, wouldve, x, y, yes, yet, you, youre, youve, your, yours, yourself, yourselves, z, zero

\clearpage

\section{Partisanship Assignment}
\label{sec:appendix_partisanship}
\subsection{Political Twitter Handles}
\label{ssec:appendix_politician_handles}
\paragraph{Democrat.}
AGBecerra, AlanGrayson, AngusKing2018, AnthonyBrownMD4, BarbaraBoxer, BenCardinforMD, BennetForCO, BennieGThompson, BernieSanders, BettyMcCollum04, BillPascrell, Bob\_Casey, BobbyScott, Booker4Senate, BradSherman, Call\_Me\_Dutch, ChrisCoons, ChrisMurphyCT, ChrisVanHollen, Clyburn, CongressmanRaja, CongressmanRuiz, CoryBooker, DWStweets, DianneFeinstein, DickBlumenthal, DickDurbin, DonaldNorcross, DorisMatsui, EPWDems, EdMarkey, EleanorNorton, EnergyDems, FrankPallone, GKButterfield, GerryConnolly, HELPCmteDems, HeidiHeitkamp, Heinrich4NM, HillaryClinton, HouseDemocrats, JECDems, JacksonLeeTX18, JeanneShaheen, JeffMerkley, JimLangevin, JoaquinCastrotx, JoeManchinWV, JohnCarneyDE, JuliaBrownley, JuliaBrownley26, KamalaHarris, LacyClayMO1, LloydDoggettTX, LorettaSanchez, MariaCantwell, MarkWarner, MartinHeinrich, McCaskillOffice, Menendez4NJ, MurrayCampaign, NancyPelosi, NelsonForSenate, NitaLowey, NormaJTorres, NydiaVelazquez, PattyMurray, PeterWelch, Peters4Michigan, RepAdamSchiff, RepAdamSmith, RepAlGreen, RepAlLawsonJr, RepAndreCarson, RepAnnaEshoo, RepAnnieKuster, RepBRochester, RepBarbaraLee, RepBarragan, RepBeatty, RepBera, RepBetoORourke, RepBillFoster, RepBobbyRush, RepBonamici, RepBonnie, RepBradAshford, RepBrady, RepBrendanBoyle, RepBrianHiggins, RepCarbajal, RepCardenas, RepCartwright, RepCharlieCrist, RepCheri, RepCicilline, RepCohen, RepCuellar, RepCummings, RepDanKildee, RepDannyDavis, RepDarrenSoto, RepDavidEPrice, RepDeSaulnier, RepDebDingell, RepDelBene, RepDennyHeck, RepDerekKilmer, RepDianaDeGette, RepDonBeyer, RepDonaldPayne, RepDwightEvans, RepEBJ, RepEliotEngel, RepEspaillat, RepEsty, RepFilemonVela, RepGaramendi, RepGeneGreen, RepGraceMeng, RepGregoryMeeks, RepGutierrez, RepGwenMoore, RepHanabusa, RepHankJohnson, RepHastingsFL, RepHuffman, RepJackyRosen, RepJaredPolis, RepJayapal, RepJeffries, RepJerryNadler, RepJimCosta, RepJimMcDermott, RepJimmyPanetta, RepJoeCourtney, RepJoeKennedy, RepJohnConyers, RepJohnDelaney, RepJohnLarson, RepJohnYarmuth, RepJoseSerrano, RepJoshG, RepJuanVargas, RepJudyChu, RepKClark, RepKarenBass, RepKathleenRice, RepKihuen, RepLawrence, RepLindaSanchez, RepLipinski, RepLoisCapps, RepLoisFrankel, RepLouCorrea, RepLowenthal, RepLujanGrisham, RepMaloney, RepMarciaFudge, RepMarcyKaptur, RepMarkTakai, RepMarkTakano, RepMaxineWaters, RepMcEachin, RepMcGovern, RepMcNerney, RepMikeHonda, RepMikeQuigley, RepOHalleran, RepPaulTonko, RepPerlmutter, RepPeteAguilar, RepPeterDeFazio, RepRaskin, RepRaulGrijalva, RepRichardNeal, RepRichmond, RepRickLarsen, RepRoKhanna, RepRobinKelly, RepRonKind, RepRoybalAllard, RepRubenGallego, RepSarbanes, RepSchakowsky, RepSchneider, RepSchrader, RepScottPeters, RepSeanMaloney, RepSheaPorter, RepSinema, RepSires, RepSpeier, RepStephMurphy, RepStephenLynch, RepSteveIsrael, RepSusanDavis, RepSwalwell, RepTedDeutch, RepTedLieu, RepTerriSewell, RepThompson, RepTimRyan, RepTimWalz, RepTomSuozzi, RepValDemings, RepVeasey, RepVisclosky, RepWilson, RepYvetteClarke, RepZoeLofgren, RonWyden, SanfordBishop, SchatzforHawaii, SenAngusKing, SenBennetCO, SenBillNelson, SenBlumenthal, SenBobCasey, SenBooker, SenBrianSchatz, SenCoonsOffice, SenCortezMasto, SenDonnelly, SenDuckworth, SenFeinstein, SenFranken, SenGaryPeters, SenGillibrand, SenJackReed, SenJeffMerkley, SenKaineOffice, SenKamalaHarris, SenMarkey, SenSanders, SenSchumer, SenSherrodBrown, SenStabenow, SenWarren, SenWhitehouse, Sen\_JoeManchin, SenateApprops, SenateDems, SenatorBaldwin, SenatorBarb, SenatorBoxer, SenatorCantwell, SenatorCardin, SenatorCarper, SenatorDurbin, SenatorHassan, SenatorHeitkamp, SenatorLeahy, SenatorMenendez, SenatorReid, SenatorShaheen, SenatorTester, SenatorTomUdall, SenatorWarner, SherrodBrown, StaceyPlaskett, SupJaniceHahn, TheDemocrats, TomCarperforDE, TomUdallPress, TulsiPress, USRepKCastor, USRepKeating, USRepMikeDoyle, USRepRHinojosa, USRepRickNolan, WhipHoyer, WydenForOregon, WydenPress, alfranken, amyklobuchar, brianschatz, cbrangel, chakafattah, chelliepingree, chuckschumer, clairecmc, collinpeterson, coons4delaware, daveloebsack, dscc, elizabethforma, gracenapolitano, jahimes, janschakowsky, jontester, keithellison, louiseslaughter, mazieforhawaii, maziehirono, mikecapuano, nikiinthehouse, pedropierluisi, repbenraylujan, repblumenauer, repcleaver, repdavidscott, repdinatitus, repdonnaedwards, repjimcooper, repjoecrowley, repjohnlewis, repmarkpocan, repsandylevin, rosadelauro, sethmoulton, stabenow, tammybaldwin, timkaine, tomudall

\paragraph{Republicans.}
AustinScottGA08, BankingGOP, CarlyFiorina, ChrisChristie, ChuckGrassley, ConawayTX11, CongCulberson, CongMikeSimpson, CongressmanDan, CongressmanGT, CongressmanHice, DanaRohrabacher, DarrellIssa, DrNealDunnFL2, DrPhilRoe, EPWGOP, EdWorkforce, EnergyGOP, FinancialCmte, GOP, GOPHELP, GOPLeader, GOPSenFinance, GOPoversight, GOPpolicy, GovMikeHuckabee, GrahamBlog, GrassleyPress, GreggHarper, HASCRepublicans, HerreraBeutler, HouseAdmnGOP, HouseAgNews, HouseAppropsGOP, HouseCommerce, HouseGOP, HouseHomeland, HouseJudiciary, HouseScience, HouseSmallBiz, HouseVetAffairs, HurdOnTheHill, InhofePress, JebBush, JeffFlake, JeffFortenberry, JerryMoran, JimInhofe, JimPressOffice, Jim\_Jordan, JohnBoozman, JohnCornyn, JohnKasich, JudgeCarter, JudgeTedPoe, KeithRothfus, KenCalvert, LamarSmithTX21, MacTXPress, MarioDB, MarkAmodeiNV2, MarshaBlackburn, McConnellPress, MikeCrapo, MikeKellyPA, NatResources, PatTiberi, PatrickMcHenry, PeteSessions, PeterRoskam, PortmanPress, RandPaul, Raul\_Labrador, RealBenCarson, RepAbraham, RepAdrianSmith, RepAndyBarr, RepAndyHarrisMD, RepAnnWagner, RepArrington, RepBillFlores, RepBillJohnson, RepBillShuster, RepBlainePress, RepBobGibbs, RepBradWenstrup, RepBrianBabin, RepBrianMast, RepBuddyCarter, RepByrne, RepCharlieDent, RepChrisCollins, RepChrisSmith, RepChrisStewart, RepChuck, RepComstock, RepCurbelo, RepDLamborn, RepDaveJoyce, RepDavid, RepDavidValadao, RepDavidYoung, RepDeSantis, RepDennisRoss, RepDianeBlack, RepDougCollins, RepDrewFerguson, RepEdRoyce, RepErikPaulsen, RepFrankLucas, RepFredUpton, RepFrenchHill, RepGallagher, RepGarretGraves, RepGoodlatte, RepGosar, RepGusBilirakis, RepGuthrie, RepHalRogers, RepHartzler, RepHensarling, RepHolding, RepHuizenga, RepHultgren, RepJBridenstine, RepJackBergman, RepJasonLewis, RepJasonSmith, RepJeffDenham, RepJeffDuncan, RepJimBanks, RepJimRenacci, RepJoeBarton, RepJoeWilson, RepJohnFaso, RepJohnKatko, RepKayGranger, RepKenBuck, RepKenMarchant, RepKevinBrady, RepKevinCramer, RepKevinYoder, RepKinzinger, RepKristiNoem, RepLaHood, RepLaMalfa, RepLanceNJ7, RepLarryBucshon, RepLeeZeldin, RepLoBiondo, RepLouBarletta, RepLoudermilk, RepLukeMesser, RepLynnJenkins, RepMGriffith, RepMarkMeadows, RepMarkWalker, RepMarthaRoby, RepMcCaul, RepMcClintock, RepMcKinley, RepMeehan, RepMiaLove, RepMikeBishop, RepMikeCoffman, RepMikeRogersAL, RepMikeTurner, RepMimiWalters, RepMoBrooks, RepMoolenaar, RepMullin, RepNewhouse, RepPaulCook, RepPaulMitchell, RepPeteKing, RepPeteOlson, RepPoliquin, RepRWilliams, RepRalphNorman, RepRichHudson, RepRickAllen, RepRobBishop, RepRodBlum, RepRussell, RepRyanCostello, RepRyanZinke, RepSanfordSC, RepScottPerry, RepSeanDuffy, RepShimkus, RepSmucker, RepStefanik, RepStevePearce, RepSteveStivers, RepTedYoho, RepThomasMassie, RepTipton, RepTomEmmer, RepTomGarrett, RepTomGraves, RepTomMacArthur, RepTomMarino, RepTomPrice, RepTomReed, RepTomRice, RepTrentFranks, RepTrey, RepWalberg, RepWalorski, RepWalterJones, RepWebster, RepWesterman, Rep\_Hunter, RepublicanStudy, RobWittman, Robert\_Aderholt, RodneyDavis, RosLehtinen, RoyBlunt, RoyBluntPress, RulesReps, SASCMajority, SamsPressShop, SenAlexander, SenBobCorker, SenCapito, SenCoryGardner, SenDanSullivan, SenDeanHeller, SenJohnBarrasso, SenJohnHoeven, SenJohnKennedy, SenJohnMcCain, SenJohnThune, SenJoniErnst, SenMikeLee, SenPatRoberts, SenRonJohnson, SenRubioPress, SenSasse, SenTedCruz, SenThadCochran, SenThomTillis, SenToddYoung, SenTomCotton, SenToomey, SenateCommerce, SenateGOP, SenateMajLdr, SenateRPC, SenatorBurr, SenatorCollins, SenatorEnzi, SenatorFischer, SenatorIsakson, SenatorLankford, SenatorRisch, SenatorRounds, SenatorTimScott, SenatorWicker, SpeakerRyan, SteveDaines, SteveKingIA, SteveScalise, SusanWBrooks, TGowdySC, TXRandy14, ToddRokita, TomColeOK04, TomRooney, Transport, USRepGaryPalmer, USRepLong, USRepRodney, VernBuchanan, WarrenDavidson, WaysandMeansGOP, amashoffice, boblatta, cathymcmorris, congbillposey, davereichert, farenthold, housebudgetGOP, lisamurkowski, marcorubio, michaelcburgess, mike\_pence, realDonaldTrump, rep\_stevewomack, repdavetrott, repdonyoung, repgregwalden, replouiegohmert, reppittenger, senategopfloor, sendavidperdue, senorrinhatch, senrobportman, tedcruz, virginiafoxx

\subsection{Partisan Assignment Sanity Check}
\begin{figure}[htbp]
 \centering
   \centering
   \includegraphics[width=\linewidth]{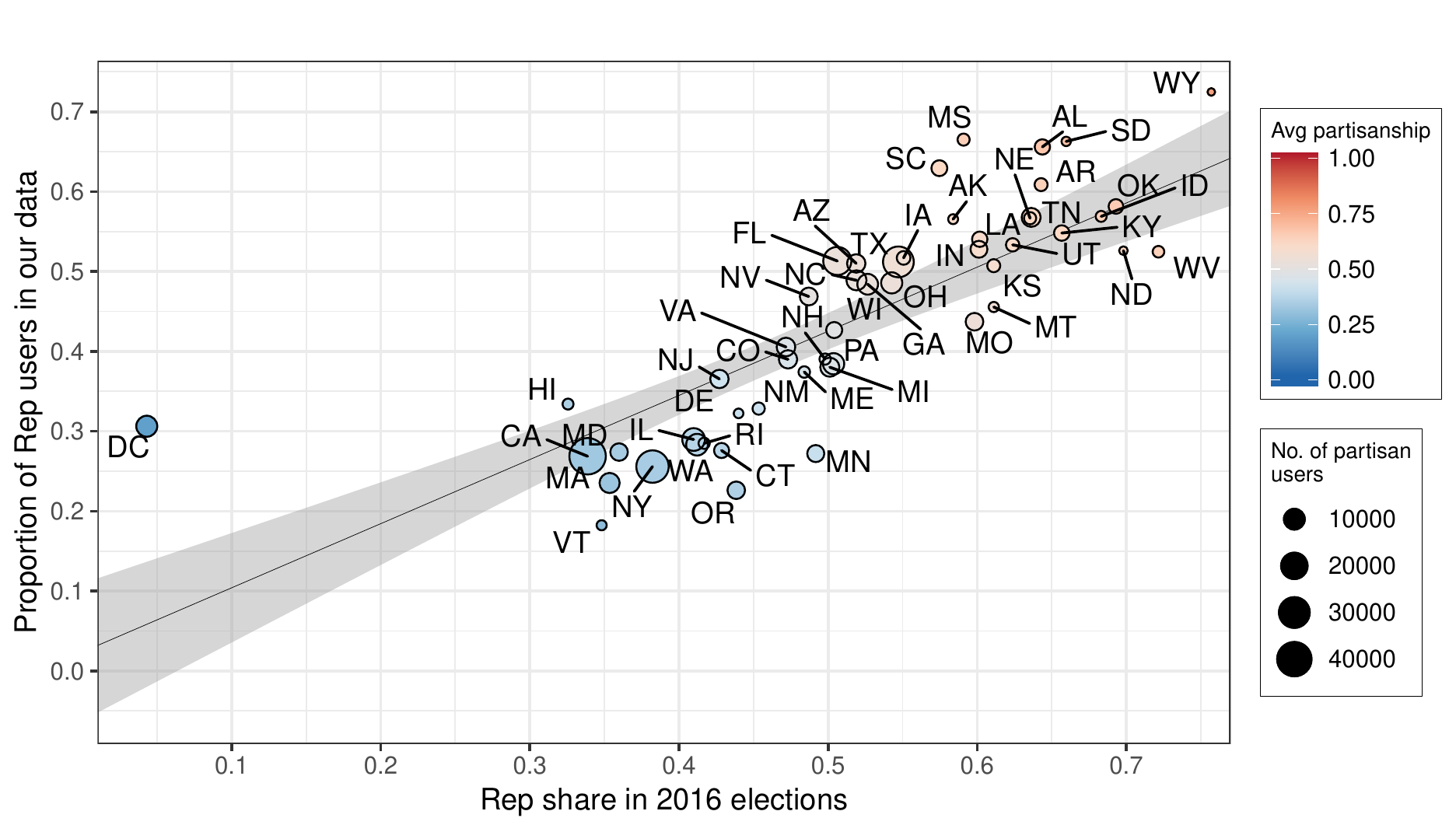}
   \caption{This plot was generated the same way as Figure~\ref{fig:sanity_check}, except that it also includes DC, which shows that it is an outlier in our data due to the fact that there are many Republican politicians and media outlets who are affiliated on Twitter with DC, while DC's voting population tends to be Democratic.} 
   \label{fig:sanity_check_dc}
\end{figure}

In our sanity check, we exclude DC, as 1) it is not an official US state and 2) there the population of users (including politicians and media) is expected to differ from the voting population. Figure~\ref{fig:sanity_check_dc} gives DC values.


\subsection{Russian account presence} We find no substantial presence of Russian accounts in the tweet set that we used for analysis, after all pre-processing. We use the list of Russian accounts identified by Twitter and banned in November 2017.\footnote{Available at \tiny\url{https://www.recode.net/2017/11/2/16598312/russia-twitter-trump-twitter-deactivated-handle-list}} This list is likely to be an underestimate of the true presence of foreign influence, but it nevertheless provides some indication of such activity. Table~\ref{tab:russianbots} contains a breakdown by event; we exclude the events that occurred after the accounts were banned. Orlando had one account that tweeted 115 times, and Vegas had 4 that tweeted a total of 70 times. 

\begin{table*}[!htbp] 
\begin{center}
\resizebox{.6\textwidth}{!}{%
\begin{tabular}{l|c c}
Event &          Number of Russian accounts &    Number of tweets\\\hline
Baton Rouge	&8	      &12 \\
Burlington	&3	       &9 \\
Chattanooga	&0	      &0 \\
Colorado Springs	&0	       &0 \\
Dallas	&4	      &39 \\
Fort Lauderdale	&0	      &0 \\
Fresno	&2	       &2 \\
Kalamazoo	&0	      &0 \\
Orlando	&1	       &115 \\
Roseburg	&0	       &0 \\
San Bernardino	&0	       &0 \\
San Francisco	&0	      &0 \\
Sutherland Springs	&0	       &0 \\
Thornton	&0	       &0 \\
Vegas	&4	       &70 \\
\end{tabular} %
}
\caption{Tweets by Russian accounts} 
  \label{tab:russianbots} 
\end{center}
\end{table*}

\clearpage\section{Topic Model Outputs}
\label{sec:appendix_topic_outputs}
\subsection{Topic Words}

Table~\ref{tab:topic_words_mallet} and Table~\ref{tab:topic_words_btm} show the highest probability words per topic (for $k=8$) for MALLET and BTM, respectively.
 \begin{table}[]
\centering
\resizebox{\linewidth}{!}{%
\begin{tabular}{|l|l|}
\hline
\rowcolor[HTML]{C0C0C0} 
\textbf{Topic} & \textbf{10 Highest Probability Stems}                                    \\ \hline
1              & gun, shoot, law, control, peopl, shooter, church, hous, stop, school     \\ \hline
2              & school, shoot, high, gun, student, parkland, kid, shooter, texa, kill    \\ \hline
3              & shoot, victim, famili, prayer, thought, pray, today, kill, gun, heart    \\ \hline
4              & shoot, polic, dead, shooter, report, suspect, peopl, shot, kill, airport \\ \hline
5              & shoot, shooter, attack, terrorist, gun, terror, san, trump, plan, call   \\ \hline
6              & shoot, vega, mass, las, thousand, victim, kill, california, bar, church  \\ \hline
7              & victim, shoot, trump, hous, flag, capit, honor, half, presid, staff      \\ \hline
8              & kill, peopl, white, hous, shoot, shooter, shot, black, polic, man        \\ \hline
\end{tabular}%
}
  \caption{The highest probability stems per topic (for $k=8$) for MALLET.}
\label{tab:topic_words_mallet}
\end{table}
 \begin{table}[]
\centering
\resizebox{\linewidth}{!}{%
\begin{tabular}{|l|l|}
\hline
\rowcolor[HTML]{C0C0C0} 
\textbf{Topic} & \textbf{10 Highest Probability Stems}                                         \\ \hline
1              & victim, famili, prayer, today, thought, life, communiti, violenc, pray, heart \\ \hline
2              & polic, report, suspect, shooter, dead, offic, shot, capit, break, multipl     \\ \hline
3              & shoot, mass, vega, news, gunman, thousand, las, dead, california, die         \\ \hline
4              & attack, san, trump, orlando, terrorist, plan, bernardino, call, terror, obama \\ \hline
5              & shoot, church, shot, live, airport, time, day, texa, fire, talk               \\ \hline
6              & kill, peopl, hous, white, stop, man, guy, black, murder, colorado             \\ \hline
7              & shooter, dalla, cop, killer, media, blame, make, show, cnn, post              \\ \hline
8              & gun, school, high, control, law, parkland, student, nra, kid, arm             \\ \hline
\end{tabular}%
}
  \caption{The highest probability stems per topic (for $k=8$) for BTM.}
\label{tab:topic_words_btm}
\end{table}

\subsection{Topic Tweets}
\label{ssec:appendix_sample_topic_tweets}
We present a sample of tweets belonging to one of the 8 topics assigned by our model.
\paragraph{News.}
\begin{itemize}
    \item HAPPENING NOW: Multiple people wounded in shooting at Colorado Walmart
    \item 3 people are reported dead in \#FtLauderdaleshooting
    \item UPDATE: Baton Rouge 2 Officers confirmed dead 7 officers have been shot in total. Details are slowly coming in.
    \item San Francisco police responding to reports of a shooting at a UPS facility, according to multiple reports.
    \item BREAKING: Police confirm several fatalities in \#Annapolis newsroom shooting and multiple people seriously injured. The suspect is in custody. @SkyNewsAust
\end{itemize}

\paragraph{Investigation.}
\begin{itemize}
    \item Alleged synagogue shooter charged in deaths of 11 people
    \item Michigan prosecutor: Suspect in Kalamazoo rampage that killed 6 admitted to shootings.
    \item \#COSshooting \#PlannedParenthood It's over...suspect walked out and they apprehended him-no altercation in arrest. Suspect turned himself in
    \item Capital Gazette shooter has been identified as Jarrod W. Ramos, 38, who had previously filed a defamation lawsuit against the paper and a columnist in 2012.
    \item Waffle House gunman's father facing charges for GIVING him gun used to kill four
\end{itemize}

\paragraph{Shooter's identity \& ideology.}
\begin{itemize}
    \item HATE CRIME : Fresno Islamic Killer Referred to White People as ``Devils"
    \item To say that extremist Muslims represent all Muslims is like saying the gunmen in Colorado springs represents all Christians
    \item So who has \#BatonRouge blamed for this shooting Imperialist Obama BLM Movement The Black Man Or the isolated mentally ill white lone wolf
    \item Again, the Lafayette white killer is a ``lone wolf" but the Chattanooga Arab killer is an entire religion. Aggressively insane troll logic.
    \item Who is surprised that the San Bernardino shooters were radicalized Muslims?
\end{itemize}

\paragraph{Victims \& location.}
\begin{itemize}
    \item 12 Killed in California Shooting; Gunman Targeted Bar in Thousand Oaks
    \item Synagogue shooting in Pittsburgh: what we know
    \item Navy Veteran Who Survived Afghanistan Dies in Las Vegas Shooting
    \item Aaron Feis, who died in the mass shooting at Marjory Stoneman Douglas High, was praised by college recruiters and former players for helping dozens of high school athletes land a chance to play in college.
    \item Texas church, site of deadly massacre, to be demolished (Via WLTX 19)
\end{itemize}

\paragraph{Laws \& policy.}
\begin{itemize}
    \item This Parkland Tragedy was completely a security meltdown in or security officials. 100\% preventable. Gun laws had nothing to do with this massacre. But gun control could have diminish the carnage. Two different things...
    \item NRA allowed acts like \#Chattanooga to become commonplace. Their lobbying permits people on the terror watch list to buy guns. Remember that.
    \item I will not just \#prayforsutherlandsprings, today I vote for @PhilMurphyNJ and stronger gun control laws in NJ \#ElectionDay \#GunControlNow
    \item Again the mental health flags were waving about shooter in Santa Fe no one acted. By now, if ones in charge cared and wanted to protect students, Every School would have had a security assessment and have hardened access points. Can never stop these but can make it harder.
    \item This is a mental health issue, security issue AND a gun issue. Our government has taken action on NONE of these to protect our students. So clearly we are not being heard, and our kids are being executed. What do we do now? \#EnoughIsEnough \#DoSomething \#SantaFe \#WeMustBeHeard
\end{itemize}

\paragraph{Solidarity.}
\begin{itemize}
    \item Our thoughts and prayers go out the victims and their families involved in the \#Chattanooga tragedy 2nite.
    \item Praying for the loved ones of the victims of the shooting yesterday at the synagogue in Pittsburgh.
    \item Our prayers go out to the victims, families, friends \& everyone affected in \#SanBernardino \#BeSafe \#BeAware
    \item My heart goes out to the friends and family of the victims in that parkland shooting :-(
    \item Our hearts goes to \#UmpquaCommunityCollege in the \#oregonshooting \#ChristCenteredEnt is praying 4 u \#AllLivesMatter
\end{itemize}

\paragraph{Remembrance.}
\begin{itemize}
    \item @HarfordCC will honor victims @UmpquaCC TODAY by observing a National Moment of Silence 1:45PM in Quad. Students should attend \#IamUCC
    \item Photo: US flag flying at half-staff at White House for victims of Roseburg, Ore., school shooting
    \item More than 100 Romans gather in solidarity against hate, honor victims of Orlando shooting
    \item Trump Denies Request To Lower Flags To Honor Capital Gazette Shooting Victims \#DouchebagDon \#CrookedTrump @FLOTUS @realDonaldTrump
    \item Live from \#ChattanoogaShootings memorial next on \#11Alive
\end{itemize}

\paragraph{Other.}
\begin{itemize}
    \item I'm sure coming from you, this has nothing to do with the Fresno shooting?
    \item I realize the dude from the Waffle House shooting did a miraculous thing disarming the shooter, but if I see the gaping wound on his forearm one more time I'm gonna lose my mind.
    \item The only thing that stops a bad guy with a gun, is a good guy with a gun!!! \#Kalamazoo
    \item The little clip I saw of what DT just said about \#PittsburghShooting \#PittsburghSynagogueShooting was ok (I guess)
    \item But when is someone gonna reassure me a planned parenthood attack won't happen
\end{itemize}

\clearpage\section{Topic Model Evaluation}
\label{sec:appendix_topic_eval}
In the next two subsections we describe the tasks which we crowdsourced to compare the three topic models: MALLET, BTM and our embedding-based model.
\subsection{Word intrusion}
Our word intrusion task is the same as is described in \newcite{chang2009reading}. Our topic-word distance metric for MALLET and BTM is probability (we use the exact topic-word matrix that these models output) and for our model it is cosine distance. We created 2850 experimental items (i.e. sets of words) with the following procedure: 
\begin{enumerate}
    \item Sample a model $M$, a $k$ (between $6-10$) and a topic $x$ ranging between $1-k$.
    \item For a given choice of $M$, $k$ and $x$,
    \begin{enumerate}
        \item sample $5$ words among the closest $10$ words to $x$.
        \item sample one word that is among the $5\%$ of the furthest words from $x$ but also among the $5\%$ of the closest words to another topic.
    \end{enumerate}
    \item Shuffle the words.
\end{enumerate}

Turkers were asked to pick the odd word out among a set of words. Each MTurk HIT consisted of $6$ sets of words created via this procedure. 

\subsection{Tweet intrusion}
Since we only assign one topic to each tweet, we would like to evaluate how coherent the tweets are that are assigned to the same topic. Therefore, we define the tweet intrusion task analogously to word intrusion. Our distance metric in this case is the ratio of proximities (probability for MALLET and BTM and cosine for our model) between the closest topic and second closest topic --- this value quantifies the proximity to the closest topic as well as how uniquely close that topic is (in contrast to the second topic). We created 1800 experimental items via the following procedure:
\begin{enumerate}
    \item Sample a model $M$, a $k$ (between $6-10$) and a topic $x$ ranging between $1-k$.
    \item For a given choice of $M$, $k$ and $x$,
    \begin{enumerate}
        \item sample $3$ tweets among the closest $1\%$ of tweets to $x$.
        \item sample one tweet that is among the $1\%$ of the furthest tweets from $x$ but also among the $1\%$ of the closest tweets to another topic.
    \end{enumerate}
    \item Shuffle the tweets.
\end{enumerate}
Turkers were asked to pick the odd tweet out from these tweets. Each MTurk HIT consisted of three sets of tweets.

\clearpage\section{Emotion Lexicon}
\label{sec:appendix_emotion}

The following words were the final stems in our emotional lexicon.
\begin{description}
\item[positive] love, friend, pray, thought, affect, bless, god, pleas, communiti, hope, stand, thank, help, condol, will, comfort, time, strong, work, support, effect, strength, feel, peac, word, rest, give, great, action, good

\item[negative] hate, violenc, hatr, of, evil, tragedi, will, word, attack, sad, feel, anger, murder, shoot, massacr, want, need, pain, kill, griev, crime, ignor, victim, lost, grief, senseless, tragic, fear, loss, sick

\item[sadness] senseless, loss, tragedi, lost, devast, sad, love, griev, horrif, terribl, pain, violenc, condol, broken, hurt, feel, victim, mourn, horrifi, will, grief, ach, suffer, sick, kill, aw, sicken, evil, massacr, mad

\item[disgust] disgust, sick, shame, ignor, wrong, blame, hell, ridicul, idiot, murder, evil, coward, sicken, feel, disgrac, slaughter, action, bad, insan, attack, pathet, outrag, polit, terrorist, mad, damn, lose, shit, lie, asshol

\item[anger] gun, will, murder, kill, violenc, wrong, shoot, bad, death, attack, feel, shot, action, arm, idiot, crazi, crimin, terrorist, mad, hell, crime, blame, fight, ridicul, insan, shit, die, threat, terror, hate

\item[fear] danger, threat, fear, arm, gun, still, shooter, attack, feel, fight, hide, murder, shot, shoot, bad, kill, chang, serious, violenc, forc, risk, defend, warn, govern, concern, fail, polic, wrong, case, terrorist

\item[trust] school, like, good, real, secur, show, nation, don, protect, call, teacher, help, law, great, save, true, wonder, respons, sad, answer, person, feel, safe, thought, continu, love, guard, church, fact, support
\end{description}

The following words were used as seeds to generate this lexicon, as described in the main text.

\begin{description}
\item[positive]  love, donat, heart, thought, strength, bless, solidar

\item[negative]  hatr, hate, griev, grief, wrong

\item[sadness]  mourn, sadden, griev, grief, sad, suffer, affect, broken, senseless, loss, heartbroken

\item[disgust]  disgust, disgrac, shame, gut, slaughter, sicken, sick, ill, lunat, coward

\item[anger]  deserv, lynch, gang, threat, mad, sicken, harm, enforc, firearm, ridicul, assault

\item[fear]  risk, hide, danger, warn, fear

\item[trust]  secur, coach, safe, hero, nation

\end{description}

\clearpage\section{Most Partisan Phrases}
\label{sec:appendix_partisan_words}

\subsection{Most Partisan Phrases Overall}
\label{ssec:appendix_overall_partisan_words}

We list the 20 most Democrat and Republican unigrams and bigrams that occur at least 100 times in tweets about a particular event. The number in brackets indicates the $z$-score of the log odds of these words \cite{monroe2008fightin} --- values with an absolute value greater than 2 are significantly partisan.

\paragraph{Chattanooga.}
\emph{Most Republican phrases:}  obama (7.41), gun free (5.89), zone (5.70), free (5.69), free zone (5.53), flag (5.33), \#tcot (5.33), \#chattanoogaattack (5.13), \#wakeupamerica (4.94), islam (4.69), \#obama (4.01), \#islam (3.83), attack (3.67), \#gunfreezon (3.66), lower (3.66), liber (3.54), arm (3.44), workplac (3.32), white hous (3.23), workplac violenc (3.21) \\
\emph{Most Democrat phrases:}  blame georg (-8.62), bush invad (-8.62), invad iraq (-8.60), invad (-8.60), base lie (-8.55), georg bush (-8.53), war base (-8.51), lie happen (-8.51), georg (-8.35), iraq war (-8.31), iraq (-8.24), bush (-7.77), lie (-7.45), \#charleston (-6.99), mass (-6.82), \#lafayett (-6.48), happen (-6.19), \#charlestonshoot (-5.96), blame (-5.40), \#gunsens (-5.09)
\paragraph{Roseburg.}
\emph{Most Republican phrases:}  obama (8.02), \#2a (6.28), \#obama (6.01), gun free (5.37), \#tcot (5.32), free (5.22), christian (5.09), zone (5.08), chicago (5.06), shooter (4.78), free zone (4.75), \#gunfreezon (4.04), agenda (3.95), religion (3.85), train (3.79), liber (3.73), \#christian (3.71), secur (3.45), guard (3.40), skarlato (3.39) \\
\emph{Most Democrat phrases:}  \#gunsens (-4.69), heart (-4.59), gun nut (-3.92), fuck (-3.83), mass (-3.82), gun violenc (-3.81), violenc (-3.80), nra (-3.54), thought (-3.47), nut (-3.28), school (-3.22), \#gunviol (-3.10), countri (-3.08), chang (-3.04), congress (-3.02), love (-2.99), vigil (-2.86), mass shoot (-2.75), protest (-2.74), america (-2.63)
\paragraph{Colorado Springs.}
\emph{Most Republican phrases:}  babi (13.62), liber (11.39), kill babi (8.77), polic (8.46), kill (8.30), left (7.42), offic (7.01), bank (6.97), babi kill (6.84), lib (6.42), obama (6.34), \#activeshoot (6.23), \#tcot (6.13), activ (6.12), parenthood kill (6.01), report (5.99), injur (5.90), \#break (5.81), activ shooter (5.75), plan (5.72) \\
\emph{Most Democrat phrases:}  terrorist (-13.88), terror (-9.71), attack (-9.52), terrorist attack (-9.09), white (-8.33), \#plannedparenthoodshoot (-7.63), \#plannedparenthood (-7.29), gop (-7.04), domest (-6.67), candid (-6.40), \#gopdeb (-6.32), \#standwithpp (-6.27), attack \#plannedparenthood (-6.23), women (-5.84), radic (-5.74), \#gop (-5.39), defund (-5.05), christian (-5.02), call (-4.89), aliv (-4.73)
\paragraph{San Bernardino.}
\emph{Most Republican phrases:}  obama (12.59), attack (12.08), \#tcot (9.33), terrorist (9.26), islam (9.20), terrorist attack (8.65), muslim (7.48), liber (6.94), \#2a (6.82), climat (6.62), climat chang (6.34), blame (5.91), \#obama (5.79), islam terrorist (5.65), \#pjnet (5.61), \#wakeupamerica (5.35), workplac violenc (5.19), foxnew (5.05), call (5.02), vet (4.89) \\
\emph{Most Democrat phrases:}  mass (-13.70), mass shoot (-10.50), \#gunsens (-10.35), shoot (-7.97), disabl (-7.17), gop (-5.89), development (-5.82), fuck (-5.73), development disabl (-5.67), \#gopdeb (-5.64), heart (-5.37), center (-5.34), thought (-5.29), day (-4.96), action (-4.93), domest (-4.78), \#gunviol (-4.73), congress (-4.50), sick (-4.40), normal (-4.32)
\paragraph{Kalamazoo.}
\emph{Most Republican phrases:}  michigan (4.02), \#break (3.39), ap (3.37), barrel (3.03), counti (2.98), cracker (2.98), cracker barrel (2.93), suspect (2.80), polic (2.71), area (2.56), dead (2.53), random shoot (2.49), mich (2.43), year girl (2.33), 14 year (2.24), victim (2.22), shoot (2.22), charg (2.20), 7 (2.20), counti michigan (2.18) \\
\emph{Most Democrat phrases:}  gun (-4.37), white (-4.11), mass (-4.00), terrorist (-3.63), america (-3.39), \#gunviol (-3.08), white man (-2.72), mental (-2.72), gun violenc (-2.60), coverag (-2.58), \#kalamazooshoot (-2.52), ill (-2.42), mental ill (-2.38), unarm (-2.37), guy (-2.27), countri (-2.27), white male (-2.24), black (-2.20), violenc (-2.02), talk (-1.88)
\paragraph{Orlando.}
\emph{Most Republican phrases:}  islam (59.38), attack (53.07), terrorist (48.13), obama (47.01), \#tcot (39.27), fbi (38.32), terror (37.21), blame (36.48), terror attack (35.32), terrorist attack (33.95), mateen (32.23), hillari (32.04), jihad (31.77), shooter (31.72), isi (31.63), radic (31.07), democrat (30.47), killer (29.49), liber (29.01), islam terror (28.69) \\
\emph{Most Democrat phrases:}  victim (-54.39), love (-40.34), heart (-34.47), hate (-34.40), violenc (-31.13), communiti (-29.11), \#loveislov (-28.11), gun violenc (-27.06), lgbt (-26.84), lgbtq (-26.44), donat (-24.12), \#prayfororlando (-24.04), mass (-23.94), \#weareorlando (-23.04), \#lgbt (-22.97), \#endgunviol (-22.94), \#gunviol (-22.68), thought (-22.49), peopl (-21.87), fuck (-21.81)
\paragraph{Dallas.}
\emph{Most Republican phrases:}  obama (23.09), \#bluelivesmatt (17.78), offic (12.76), \#obama (11.52), white (10.33), polic offic (10.13), hillari (10.10), kill white (9.77), racist (9.09), foxnew (9.00), shot (8.98), offic shot (8.95), \#tcot (8.81), hate crime (8.67), democrat (8.51), blame gun (8.49), crime (8.47), white cop (8.46), cop (8.30), 5 (8.03) \\
\emph{Most Democrat phrases:}  \#altonsterl (-21.29), \#philandocastil (-21.13), \#altonsterl \#philandocastil (-16.32), guy gun (-15.96), good guy (-15.26), violenc (-14.83), gun (-14.54), open carri (-13.00), guy (-12.63), carri (-12.47), open (-10.82), \#gunviol (-10.54), good (-10.53), stop (-10.23), \#philandocastil \#altonsterl (-10.07), nra (-10.02), gun violenc (-9.80), \#nra (-9.23), \#disarmh (-8.64), answer (-8.54)
\paragraph{Baton Rouge.}
\emph{Most Republican phrases:}  \#bluelivesmatt (8.69), obama (8.13), islam (5.87), nation islam (5.62), \#obama (5.17), shot (5.06), cop killer (5.06), killer (5.05), terrorist (4.99), nation (4.74), offic shot (4.63), hillari (4.34), \#backtheblu (4.25), \#tcot (4.05), offic (3.99), thug (3.96), islam member (3.79), 3 (3.53), \#trumppence16 (3.45), democrat (3.41) \\
\emph{Most Democrat phrases:}  gun (-8.88), violenc (-8.34), gun violenc (-6.65), \#nra (-6.19), guy gun (-5.77), \#altonsterl (-5.57), open carri (-5.48), assault (-5.48), weapon (-5.30), good guy (-5.14), blame (-4.88), citizen (-4.87), assault weapon (-4.84), carri (-4.74), race relat (-4.74), relat (-4.71), open (-4.61), guy (-4.58), civilian (-4.47), \#enough (-4.46)
\paragraph{Burlington.}
\emph{Most Republican phrases:}  cetin (5.12), arcan cetin (5.06), arcan (5.01), muslim (4.72), turkish (4.57), vote (4.37), turkey (4.32), hispan (4.14), immigr (4.04), citizen (3.79), hillari (3.64), elect (3.45), immigr turkey (3.19), turkish muslim (3.14), id (3.11), ed arcan (3.05), id ed (3.03), shooter id (3.00), citizen vote (2.97), cetin immigr (2.96) \\
\emph{Most Democrat phrases:}  victim (-4.05), gun (-3.98), famili (-2.97), thought (-2.65), peopl (-2.48), dead (-2.36), heart (-2.32), seattl (-2.30), violenc (-2.23), larg (-2.11), mile (-2.10), shooter larg (-2.08), latest (-2.05), day (-2.04), tonight (-2.03), safe (-2.01), watch (-2.00), communiti (-1.98), shoot (-1.96), \#break (-1.94)
\paragraph{Fort Lauderdale.}
\emph{Most Republican phrases:}  garag (4.54), shot fire (3.45), terrorist (3.14), fox (2.93), free zone (2.76), attack (2.71), gun free (2.70), fire (2.62), zone (2.61), free (2.43), muslim (2.36), shot (2.31), obama (2.09), park (2.08), terrorist attack (2.08), park garag (2.06), shooter (2.01), fox news (2.01), terror (1.93), report shot (1.81) \\
\emph{Most Democrat phrases:}  stop (-2.65), violenc (-2.52), custodi 9 (-2.25), gun (-2.14), mass (-2.12), fll (-2.10), 9 injur (-2.05), week (-2.00), multipl peopl (-1.91), 2017 (-1.86), airport suspect (-1.85), stay safe (-1.83), love (-1.78), heart (-1.70), safe (-1.70), report fire (-1.66), smh (-1.65), local (-1.64), msnbc (-1.64), thought (-1.61)
\paragraph{Fresno.}
\emph{Most Republican phrases:}  akbar (4.67), allahu (3.84), allahu akbar (3.76), yell (3.32), yell allahu (2.98), ap (2.61), suspect yell (2.38), allah (2.33), shout (2.25), terror attack (2.06), shout allahu (2.06), allah akbar (2.03), terrorist (1.98), terrorist attack (1.97), muslim (1.96), kill suspect (1.94), msm (1.83), islam (1.81), god (1.80), akbar hate (1.79) \\
\emph{Most Democrat phrases:}  chief (-3.97), victim (-3.89), polic (-3.74), famili (-3.65), fatal (-3.62), offic (-2.98), downtown (-2.92), dyer (-2.75), men (-2.71), polic chief (-2.70), gunman (-2.69), tuesday (-2.47), gun (-2.29), shoot (-2.22), suspect custodi (-2.17), kill california (-2.10), mass (-2.09), sad (-2.08), white men (-1.95), kill hate (-1.95)
\paragraph{San Francisco.}
\emph{Most Republican phrases:}  polic (4.85), multipl (4.41), report (4.32), pistol (4.18), assault pistol (4.12), 4 injur (3.41), shooter (3.33), assault (3.24), free (3.19), warehous (3.05), injur (3.01), facil 4 (2.98), multipl victim (2.90), report facil (2.85), stolen (2.82), shot facil (2.76), hospit (2.68), law (2.60), gun law (2.59), compani (2.51) \\
\emph{Most Democrat phrases:}  today (-4.82), mass (-4.18), sf (-4.07), die (-3.47), mention (-3.29), coverag (-2.97), yesterday (-2.95), america (-2.75), \#upsshoot (-2.49), \#sf (-2.40), gun violenc (-2.40), mass shoot (-2.40), barclay (-2.31), virginia (-2.30), \#up (-2.20), violenc (-2.17), morn (-2.17), shoot (-2.17), gop (-2.07), peopl kill (-2.04)
\paragraph{Vegas.}
\emph{Most Republican phrases:}  shooter (41.28), fbi (36.07), video (31.96), isi (31.08), democrat (26.78), paddock (26.34), liber (26.07), multipl (25.37), antifa (23.76), multipl shooter (23.61), \#maga (22.44), cbs (21.55), truth (21.49), mandalay (21.06), left (20.68), islam (20.57), guard (19.73), dem (19.51), hillari (19.02), proof (18.78) \\
\emph{Most Democrat phrases:}  \#guncontrolnow (-48.73), gun (-42.42), nra (-34.84), terrorist (-33.42), gun violenc (-31.92), \#guncontrol (-30.54), violenc (-29.87), domest (-29.17), mass (-28.41), white (-27.19), terror (-24.88), domest terror (-24.68), congress (-24.64), mass shoot (-23.77), gop (-22.84), thought (-22.82), \#nra (-22.19), talk (-22.06), fuck (-21.43), talk gun (-21.10)
\paragraph{Thornton.}
\emph{Most Republican phrases:}  multipl (6.04), suspect (5.24), parti (5.06), multipl parti (5.03), break (4.90), news (4.80), injur (4.36), dead (4.23), report (3.98), updat (3.87), polic (3.84), suspect custodi (3.75), activ (3.71), chicago (3.57), detail (3.55), 2 (3.52), report multipl (3.28), \#break (3.27), video (3.09), activ shooter (3.08) \\
\emph{Most Democrat phrases:}  white (-7.61), guy (-5.37), gun (-5.24), white guy (-3.92), bibl (-3.75), terrorist (-3.75), week (-3.61), white man (-3.48), good guy (-3.46), stack bibl (-3.33), live stack (-3.29), stack (-3.28), furnitur (-3.27), terror (-3.16), guy gun (-3.06), penalti (-3.04), bibl furnitur (-3.02), death penalti (-3.01), talk (-3.01), vega (-2.97)
\paragraph{Sutherland Springs.}
\emph{Most Republican phrases:}  shooter (19.89), church shooter (17.26), liber (16.58), antifa (16.38), democrat (15.16), atheist (14.70), attack (14.58), christian (13.31), texa church (13.17), zone (13.15), gun free (13.03), texa (12.66), free zone (12.58), leftist (12.45), illeg (12.40), hero (11.78), free (11.09), citizen (10.60), carri (10.43), \#antifa (10.27) \\
\emph{Most Democrat phrases:}  \#guncontrolnow (-17.29), prayer (-16.62), school (-15.11), thought prayer (-13.98), thought (-12.83), girl (-12.76), gun violenc (-12.56), talk (-11.53), mental (-11.29), white (-11.17), church pray (-10.81), concert (-10.79), prosecut (-10.76), \#guncontrol (-10.63), mass shoot (-10.52), gop (-10.36), violenc (-10.21), children (-10.03), congress (-9.95), jone (-9.88)
\paragraph{Parkland.}
\emph{Most Republican phrases:}  fbi (30.42), sheriff (25.67), liber (21.67), school (21.47), cruz (19.47), shooter (18.68), \#2a (18.39), broward (17.87), fail (17.17), israel (17.12), polic (16.73), deputi (16.03), failur (15.41), democrat (15.41), counti (15.05), \#qanon (14.65), enforc (14.40), gun free (14.35), truck (14.30), free zone (14.23) \\
\emph{Most Democrat phrases:}  \#gunreformnow (-26.02), \#guncontrolnow (-25.00), \#neveragain (-22.39), nra (-18.15), gop (-16.64), gun violenc (-16.45), \#parklandstrong (-15.80), \#nrabloodmoney (-15.00), vote (-14.48), trump (-13.27), violenc (-13.23), congress (-12.82), ar 15 (-12.74), ar (-12.46), \#banassaultweapon (-12.28), \#marchforourl (-11.85), survivor (-11.57), support (-11.51), assault (-11.33), fuck (-11.22)
\paragraph{Nashville.}
\emph{Most Republican phrases:}  gun free (11.69), zone (10.94), free zone (10.69), free (10.32), photo shoot (8.02), \#wbb \#wilsonbrothersband (7.96), \#wilsonbrothersband (7.96), \#wbb (7.96), band photo (7.96), brother band (7.96), wilson brother (7.96), wilson (7.96), fbi (7.94), band (7.72), gun (7.22), photo (6.95), law (6.65), brother (6.32), liber (6.21), hous gun (6.20) \\
\emph{Most Democrat phrases:}  black (-11.21), white (-9.41), trump (-8.06), terrorist (-7.67), tweet (-6.16), hero (-6.13), american (-6.05), shaw (-5.77), black man (-5.73), jame shaw (-5.58), jr (-5.44), jame (-5.42), shaw jr (-5.41), mention (-5.30), domest (-5.05), bond (-5.02), unarm (-4.74), domest terrorist (-4.69), black peopl (-4.63), man (-4.63)
\paragraph{Santa Fe.}
\emph{Most Republican phrases:}  shotgun (10.99), revolv (7.57), illeg (7.10), shooter (6.20), liber (6.20), ban (6.14), metal detector (6.12), detector (6.00), secur (5.93), metal (5.87), truck (5.79), rack (5.53), high school (5.45), bomb (5.43), high (5.33), leftist (5.30), gun rack (5.27), law stop (5.15), updat (5.15), rifl (5.12) \\
\emph{Most Democrat phrases:}  \#guncontrolnow (-12.17), \#gunreformnow (-10.36), \#neveragain (-9.79), \#enoughisenough (-8.90), nra (-8.57), children (-7.85), vote (-7.69), gun violenc (-7.59), america (-6.75), thought prayer (-6.72), violenc (-6.66), congress (-6.63), thought (-6.50), \#enough (-6.21), fuck (-6.11), white (-6.01), \#parkland (-5.74), gun (-5.64), republican (-5.44), gop (-5.40)
\paragraph{Annapolis.}
\emph{Most Republican phrases:}  blame (8.23), blame trump (7.82), liber (5.86), maryland (5.76), reuter (5.70), reuter editor (5.62), apolog (5.34), \#break (5.23), editor apolog (5.22), disciplin (4.90), apolog disciplin (4.84), hat (4.84), disciplin blame (4.74), maga hat (4.60), claim shooter (4.48), fals (4.39), polic (4.38), \#fakenew (4.36), democrat (4.32), wore maga (4.32) \\
\emph{Most Democrat phrases:}  journalist (-7.03), enemi (-6.10), press (-5.62), lower flag (-5.52), request lower (-5.36), request (-5.02), gazett victim (-4.55), declin request (-4.46), memori capit (-4.42), lower (-4.40), enemi peopl (-4.39), press enemi (-4.37), white (-4.35), flag memori (-4.34), declin (-4.30), flag (-4.13), memori (-4.12), trump declin (-4.09), mass (-4.08), \#guncontrolnow (-4.05)
\paragraph{Pittsburgh.}
\emph{Most Republican phrases:}  moment silenc (16.24), silenc (15.99), interrupt (14.74), interrupt moment (14.68), scream (13.09), moment (12.37), march (12.13), blackburn (11.85), leftist (11.73), silenc life (11.59), blame trump (11.28), silenc synagogu (11.20), protest (10.78), protest interrupt (10.56), rabbi blame (9.64), leftist interrupt (9.55), scream leftist (9.55), rage scream (9.52), horribl rage (9.43), scream insult (9.40) \\
\emph{Most Democrat phrases:}  violenc (-6.73), heart (-6.29), supremacist (-6.05), white supremacist (-6.05), muslim (-5.96), white (-5.94), mr (-5.88), trump vile (-5.86), result (-5.84), of (-5.80), result of (-5.80), inevit (-5.80), inevit result (-5.78), group (-5.77), synagogu inevit (-5.77), vile nation (-5.73), muslim group (-5.73), of trump (-5.71), roger (-5.69), massacr heart (-5.63)
\paragraph{Thousand Oaks.}
\emph{Most Republican phrases:}  california (16.51), zone (12.80), free (12.40), gun free (12.22), free zone (11.80), bar (9.98), california bar (9.61), strictest (9.01), strictest gun (8.52), men (7.63), \#foxnew (7.33), killer ian (7.12), california strictest (7.11), fear resid (7.11), report killer (7.08), prayer massacr (7.04), communist (7.04), long mock (7.00), mock hope (6.80), ian long (6.77) \\
\emph{Most Democrat phrases:}  inact (-17.50), pattern (-17.46), pattern inact (-17.45), shoot pattern (-17.43), shoot (-16.92), januari (-13.09), inact januari (-12.92), \#guncontrolnow (-8.09), mass shoot (-7.89), day (-7.34), fuck (-6.94), nra (-6.94), violenc (-6.71), mother (-6.64), mass (-6.59), thought (-6.44), high (-6.37), januar (-6.22), inact januar (-6.15), \#potus (-6.05)

\subsection{Most Partisan Phrases Per Topic}
\label{ssec:appendix_partisan_phrases_topic}

\begin{table}[]
\centering
\begin{tabular}{|l|}
\hline
\textbf{Topic colors} \\ \hline
\rowcolor[HTML]{F2D7D5}
shooter's identity
\& ideology \\ \hline
\rowcolor[HTML]{FCF3CF}
news \\ \hline
\rowcolor[HTML]{EAEDED}
victims \& location \\ \hline
\rowcolor[HTML]{D0ECE7}
laws \& policy \\ \hline
\rowcolor[HTML]{D6EAF8}
investigation \\ \hline
\rowcolor[HTML]{EBDEF0}
solidarity \\ \hline
\rowcolor[HTML]{D7FFCE}
remembrance \\ \hline
\rowcolor[HTML]{FFE8DD}
other \\ \hline
\end{tabular}
\end{table}
\begin{table*}[!h]
\centering
\resizebox{\linewidth}{!}{%
\begin{tabular}{|l|l|}
\hline
Republican & Democrat\\ \hline
\rowcolor[HTML]{F2D7D5}
\#obama (0.45), toler (0.45), presid (0.45), \#wakeupamerica (0.45), celebr (0.45) & blame georg (-0.94), invad iraq (-0.94), bush invad (-0.94), iraq war (-0.94), base lie (-0.94) \\
\hline
\rowcolor[HTML]{FCF3CF}
zone (0.35), fox news (0.34), 5th victim (0.33), \#tcot (0.33), class (0.32) & mass (-0.68), \#prayforchattanooga (-0.37), horrif (-0.37), \#breakingnew (-0.35), 4 dead (-0.32) \\
\hline
\rowcolor[HTML]{EAEDED}
victim (0.25), dead (0.18), 5 (0.15), 4 marin (0.09), gunman (0.09) & skip (-0.29), skip well (-0.25), well (-0.23), carson (-0.23), holmquist (-0.23) \\
\hline
\rowcolor[HTML]{D0ECE7}
\#chattanoogaattack (0.33), \#wakeupamerica (0.33), clinton (0.33), safe (0.33), muhammad (0.33) & nra (-0.99), nut (-0.97), \#charleston (-0.86), \#lafayett (-0.82), gun violenc (-0.71) \\
\hline
\rowcolor[HTML]{D6EAF8}
flag (0.41), terrorist (0.32), fire (0.31), search (0.30), navi (0.30) & \#fbi (-0.47), special (-0.43), treat (-0.38), natur (-0.35), agent (-0.31) \\
\hline
\rowcolor[HTML]{EBDEF0}
\#rednationris (0.44), \#islam \#rednationris (0.44), \#chattanoogashoot \#chattanoogastrong (0.44), marin \#wakeupamerica (0.44), \#chattanoogastrong \#isi (0.44) & \#guncontrol (-0.95), \#charleston (-0.95), \#sandrabland (-0.82), gun violenc (-0.81), \#charlestonshoot (-0.72) \\
\hline
\rowcolor[HTML]{D7FFCE}
decid (0.28), refus (0.28), command (0.28), troop (0.28), kill tennesse (0.28) & local (-0.63), mourn (-0.56), victim shoot (-0.52), tune (-0.46), communiti (-0.41) \\
\hline
\rowcolor[HTML]{FFE8DD}
\#teardownthismosqu (0.42), \#islamocar (0.42), \#stoprush (0.42), \#obama (0.42), \#wakeupamerica (0.42) & middl (-0.75), \#nra (-0.75), mass (-0.73), \#aurora (-0.73), \#charleston (-0.71) \\
\hline
\end{tabular} %
}
\caption{Most partisan phrases per topic for \textbf{Chattanooga}. Brackets show the $z$-scores of the log odds of each phrase.}
\end{table*}
\begin{table*}[!h]
\centering
\resizebox{\linewidth}{!}{%
\begin{tabular}{|l|l|}
\hline
Republican & Democrat\\ \hline
\rowcolor[HTML]{F2D7D5}
skarlato (0.46), \#2a (0.46), gun control (0.39), ask (0.37), agenda (0.34) & \#gunsens (-0.65), news (-0.55), pro gun (-0.42), pro (-0.41), mass (-0.39) \\
\hline
\rowcolor[HTML]{FCF3CF}
ore ap (0.27), secur (0.27), author (0.26), \#umpquacommunitycolleg (0.23), thought (0.23) & \#liveonk2 (-0.53), \#gunsens (-0.52), merci (-0.50), center (-0.46), 9 (-0.45) \\
\hline
\rowcolor[HTML]{EAEDED}
colleg (0.12), communiti colleg (0.05), communiti (0.05), chris (0.00) & chris (0.00), communiti (0.05), communiti colleg (0.05), colleg (0.12) \\
\hline
\rowcolor[HTML]{D0ECE7}
liber (0.67), guard (0.65), chicago (0.63), ban (0.58), secur (0.58) & gun nut (-0.65), fuck (-0.58), legisl (-0.57), \#sandyhook (-0.55), congress (-0.54) \\
\hline
\rowcolor[HTML]{D6EAF8}
father (0.51), christian (0.50), releas (0.42), identifi (0.30), 2 (0.30) & sheriff (-0.41), gun (-0.25), year (-0.21), 20 year (-0.17), victim (-0.13) \\
\hline
\rowcolor[HTML]{EBDEF0}
famili involv (0.42), 9 (0.41), pray famili (0.40), school communiti (0.38), 7 (0.38) & \#gunviol (-0.52), fuck (-0.50), \#gunsens (-0.40), chang (-0.38), gun violenc (-0.37) \\
\hline
\rowcolor[HTML]{D7FFCE}
\#2a (0.80), \#tcot (0.76), special (0.70), rt (0.66), media (0.54) & 3 (-0.45), tonight (-0.39), 11 (-0.37), dougla (-0.36), dougla counti (-0.35) \\
\hline
\rowcolor[HTML]{FFE8DD}
\#obama (0.69), chicago (0.68), \#2a (0.57), blame (0.56), stay (0.52) & countri (-0.61), nut (-0.53), \#roseburgoregon (-0.49), world (-0.43), thought (-0.42) \\
\hline
\end{tabular} %
}
\caption{Most partisan phrases per topic for \textbf{Roseburg}. Brackets show the $z$-scores of the log odds of each phrase.}
\end{table*}
\begin{table*}[!h]
\centering
\resizebox{\linewidth}{!}{%
\begin{tabular}{|l|l|}
\hline
Republican & Democrat\\ \hline
\rowcolor[HTML]{F2D7D5}
regist vote (1.01), young (1.01), shop (1.01), mcdonald (1.01), \#liber (1.01) & parenthood \#gopdeb (-0.37), \#carlyfiorina (-0.37), health clinic (-0.37), \#tedcruz (-0.37), morn joe (-0.37) \\
\hline
\rowcolor[HTML]{FCF3CF}
connect (0.56), \#bluelivesmatt (0.56), \#tcot (0.54), \#new (0.52), shooter surrend (0.47) & \#standwithpp (-0.63), \#gunsens (-0.62), \#terror (-0.60), \#plannedparenthoodshoot (-0.58), month (-0.57) \\
\hline
\rowcolor[HTML]{EAEDED}
kill (0.17), plan parenthood (0.13), parenthood (0.13), plan (0.11), \#plannedparenthood (-0.03) & war (-0.18), veteran (-0.17), thousand (-0.10), iraq (-0.08), \#plannedparenthood (-0.03) \\
\hline
\rowcolor[HTML]{D0ECE7}
kill babi (0.92), unborn (0.91), parenthood kill (0.88), sell bodi (0.87), gun free (0.83) & \#terror (-0.39), focus (-0.39), terrorist attack (-0.39), forc (-0.39), attack women (-0.39) \\
\hline
\rowcolor[HTML]{D6EAF8}
parenthood employe (0.81), employe (0.72), femal (0.49), gunman attack (0.42), record (0.38) & \#plannedparenthood gunman (-0.58), \#laquanmcdonald (-0.58), terrorist (-0.47), domest (-0.46), custodi aliv (-0.45) \\
\hline
\rowcolor[HTML]{EBDEF0}
babi (0.94), \#tcot (0.71), pro life (0.63), pro (0.55), \#prolif (0.54) & \#istandwithpp (-0.39), \#standwithpp (-0.39), terror (-0.37), terrorist attack (-0.37), parenthood attack (-0.35) \\
\hline
\rowcolor[HTML]{D7FFCE}
clinic (0.57), support (0.47), swasey (0.44), hous (0.43), friday (0.40) & carson (-0.42), terrorist (-0.42), cover (-0.41), terror (-0.41), terrorist attack (-0.41) \\
\hline
\rowcolor[HTML]{FFE8DD}
ppact (0.87), profit (0.84), lib (0.81), chicago (0.76), bodi part (0.74) & 4 cop (-0.42), parenthood \#gopdeb (-0.42), discuss (-0.42), \#terroristattack (-0.42), captur aliv (-0.42) \\
\hline
\end{tabular} %
}
\caption{Most partisan phrases per topic for \textbf{Colorado Springs}. Brackets show the $z$-scores of the log odds of each phrase.}
\end{table*}
\begin{table*}[!h]
\centering
\resizebox{\linewidth}{!}{%
\begin{tabular}{|l|l|}
\hline
Republican & Democrat\\ \hline
\rowcolor[HTML]{F2D7D5}
moron call (0.44), shooter bonni (0.44), warm (0.44), global warm (0.44), \#2a (0.44) & shooter post (-0.85), gun nut (-0.81), ted cruz (-0.76), domest terrorist (-0.74), ted (-0.67) \\
\hline
\rowcolor[HTML]{FCF3CF}
muslim (0.66), \#pjnet (0.66), \#sanbernardinoattack (0.66), \#tcot \#pjnet (0.66), \#tcot (0.61) & shoot today (-0.63), \#gunsens (-0.62), 355th (-0.56), houston (-0.55), savannah (-0.52) \\
\hline
\rowcolor[HTML]{EAEDED}
husband wife (0.40), massacr (0.35), husband (0.30), wife (0.22), calif (0.20) & live (-0.42), mass shoot (-0.27), shoot (-0.24), mass (-0.15), attack (-0.15) \\
\hline
\rowcolor[HTML]{D0ECE7}
law work (0.67), disarm (0.67), lib (0.67), administr (0.60), pari (0.60) & normal (-0.72), \#ppshoot (-0.72), gun nut (-0.64), gop (-0.63), topic (-0.62) \\
\hline
\rowcolor[HTML]{D6EAF8}
chief burguan (0.59), foxnew (0.51), \#sanbernardinoattack (0.50), \#tcot (0.48), devout (0.48) & presser (-0.51), expect (-0.46), purchas legal (-0.44), suspect home (-0.44), detain (-0.41) \\
\hline
\rowcolor[HTML]{EBDEF0}
\#tcot (0.75), sanbernardinopd (0.67), terrorist attack (0.55), job (0.55), polic offic (0.53) & \#gunsens (-0.63), savannah (-0.63), \#gunviol (-0.54), development disabl (-0.54), send love (-0.52) \\
\hline
\rowcolor[HTML]{D7FFCE}
\#tcot (0.70), \#pjnet (0.70), \#tcot \#pjnet (0.70), \#ccot (0.70), wh (0.47) & donat (-0.68), rais (-0.53), 10 (-0.44), public (-0.42), governor (-0.40) \\
\hline
\rowcolor[HTML]{FFE8DD}
blame gun (0.70), \#wakeupamerica (0.70), \#liber (0.70), pc (0.70), fast (0.63) & \#endgunviol (-0.69), movi (-0.69), 352 (-0.69), \#sandyhook (-0.59), shoot day (-0.59) \\
\hline
\end{tabular} %
}
\caption{Most partisan phrases per topic for \textbf{San Bernardino}. Brackets show the $z$-scores of the log odds of each phrase.}
\end{table*}
\begin{table*}[!h]
\centering
\resizebox{\linewidth}{!}{%
\begin{tabular}{|l|l|}
\hline
Republican & Democrat\\ \hline
\rowcolor[HTML]{F2D7D5}
polit (0.66), motiv (0.65), violenc (0.46), obama (0.26), peopl (0.23) & coverag (-0.44), news (-0.44), open (-0.44), cover (-0.44), white male (-0.37) \\
\hline
\rowcolor[HTML]{FCF3CF}
counti michigan (0.41), cracker (0.27), random shoot (0.27), cracker barrel (0.27), barrel (0.27) & coverag (-0.59), public (-0.52), \#michigan (-0.50), connect (-0.50), pray (-0.43) \\
\hline
\rowcolor[HTML]{EAEDED}
 &  \\
\hline
\rowcolor[HTML]{D0ECE7}
safeti (0.45), pass (0.34), suspect (0.26), prevent (0.24), \#uber (0.22) & talk (-0.36), \#guncontrol (-0.34), prayer (-0.32), \#kalamazooshoot (-0.29), gun violenc (-0.27) \\
\hline
\rowcolor[HTML]{D6EAF8}
saturday (0.45), ap (0.43), mich (0.38), 5 (0.36), 14 year (0.32) & black (-0.54), white man (-0.53), white (-0.41), white male (-0.36), pick (-0.35) \\
\hline
\rowcolor[HTML]{EBDEF0}
crazi (0.53), peopl kill (0.32), prayer victim (0.29), \#prayforkalamazoo (0.27), 6 (0.27) & \#gunviol (-0.53), place (-0.53), america (-0.45), work (-0.43), gun violenc (-0.36) \\
\hline
\rowcolor[HTML]{D7FFCE}
\#kalamazoostrong (0.29), victim (0.21), \#kalamazooshoot (0.18), michigan (0.12), kill (0.11) & rememb (-0.35), mass (-0.35), gun (-0.12), saturday (-0.05), 6 (0.00) \\
\hline
\rowcolor[HTML]{FFE8DD}
chicago (0.74), cracker barrel (0.62), barrel (0.62), cracker (0.62), 1 (0.44) & male (-0.39), white male (-0.39), gun violenc (-0.39), mass shooter (-0.39), matter (-0.31) \\
\hline
\end{tabular} %
}
\caption{Most partisan phrases per topic for \textbf{Kalamazoo}. Brackets show the $z$-scores of the log odds of each phrase.}
\end{table*}
\begin{table*}[!h]
\centering
\resizebox{\linewidth}{!}{%
\begin{tabular}{|l|l|}
\hline
Republican & Democrat\\ \hline
\rowcolor[HTML]{F2D7D5}
gun boston (0.47), state dept (0.47), boston pressur (0.47), attack solut (0.47), gohmert gun (0.47) & \#stoprush \#uniteblu (-0.92), \#connecttheleft (-0.92), \#p2 \#connecttheleft (-0.92), muslim allianc (-0.92), lgbtq hate (-0.92) \\
\hline
\rowcolor[HTML]{FCF3CF}
\#homelandsecur (0.61), \#pjnet (0.61), investig islam (0.61), \#trump2016 (0.61), park spot (0.61) & \#floridashoot \#puls (-0.78), derail (-0.78), maryland (-0.78), \#enough (-0.78), bbcworld (-0.78) \\
\hline
\rowcolor[HTML]{EAEDED}
regist democrat (0.69), shooter regist (0.69), zone state (0.69), state law (0.69), blog gay (0.69) & \#gay \#new (-0.70), connecticut nativ (-0.70), \#new \#glbt (-0.70), \#stoprush \#uniteblu (-0.70), \#lgbtnew (-0.70) \\
\hline
\rowcolor[HTML]{D0ECE7}
help shooter (0.80), sourc suggest (0.80), dc sourc (0.80), terrorist work (0.80), leftist (0.80) & \#lalovesorlando (-0.59), implor (-0.59), implor fight (-0.59), tragedi implor (-0.59), 91 (-0.59) \\
\hline
\rowcolor[HTML]{D6EAF8}
break insid (0.49), \#8230 (0.49), hous \#8230 (0.49), \#8230 \#8217 (0.49), shooter make (0.49) & bump poll (-0.89), give bump (-0.89), insid disappoint (-0.89), poll report (-0.89), disappoint give (-0.89) \\
\hline
\rowcolor[HTML]{EBDEF0}
reaction tragic (0.92), wrong reaction (0.92), search massacr (0.92), kohn (0.92), sick muslim (0.92) & lgbtqia (-0.47), lgbtqia communiti (-0.47), inform horrif (-0.47), readingatthedisco (-0.47), happen omar (-0.47) \\
\hline
\rowcolor[HTML]{D7FFCE}
netanyahu releas (0.80), lgbt hop (0.80), obama contact (0.80), hop make (0.80), trump \#nra★bangunfreezon (0.80) & \#stoprush \#uniteblu (-0.58), demand end (-0.58), \#connecttheleft (-0.58), \#p2 \#connecttheleft (-0.58), \#gay \#new (-0.58) \\
\hline
\rowcolor[HTML]{FFE8DD}
\#americafirst (0.80), killari (0.80), told word (0.80), carri terror (0.80), \#buildthewal (0.80) & learn hate (-0.59), hint wasn (-0.59), hate hint (-0.59), wasn osama (-0.59), shot girl (-0.59) \\
\hline
\end{tabular} %
}
\caption{Most partisan phrases per topic for \textbf{Orlando}. Brackets show the $z$-scores of the log odds of each phrase.}
\end{table*}
\begin{table*}[!h]
\centering
\resizebox{\linewidth}{!}{%
\begin{tabular}{|l|l|}
\hline
Republican & Democrat\\ \hline
\rowcolor[HTML]{F2D7D5}
ag lynch (0.45), doj (0.45), lectur (0.45), \#maga (0.45), \#dallaspoliceshoot \#bluelivesmatt (0.45) & \#disarmh (-0.90), communiti color (-0.90), \#disarmh address (-0.90), address \#gunviol (-0.90), 2 \#disarmh (-0.90) \\
\hline
\rowcolor[HTML]{FCF3CF}
\#pjnet (0.57), mt prayer (0.57), \#bluelivesmatt \#pjnet (0.57), panther (0.57), \#tcot (0.57) & lawrenc (-0.73), npr (-0.69), bbc (-0.63), \#enough (-0.60), read (-0.58) \\
\hline
\rowcolor[HTML]{EAEDED}
3 (0.36), attack (0.33), \#dallasshoot (0.31), iraq (0.28), live (0.23) & 2 (-0.38), soldier (-0.33), shot (-0.29), show (-0.23), surviv (-0.18) \\
\hline
\rowcolor[HTML]{D0ECE7}
shot arm (0.81), 12 shot (0.81), \#fbi (0.81), \#dallaspoliceshoot \#bluelivesmatt (0.81), black caucus (0.81) & scrutini (-0.58), \#disarmh (-0.58), denomin (-0.58), beget (-0.58), gun answer (-0.58) \\
\hline
\rowcolor[HTML]{D6EAF8}
comey (0.67), recommend (0.67), \#tcot (0.67), \#obama (0.67), sex (0.67) & plaster (-0.62), \#altonsterl \#philandocastil (-0.58), dallaspd man (-0.58), innoc man (-0.54), \#philandocastil (-0.53) \\
\hline
\rowcolor[HTML]{EBDEF0}
\#pjnet (0.64), mt prayer (0.64), \#bluelivesmatt \#pjnet (0.64), sam (0.64), \#maga (0.64) & poc (-0.74), \#blacklivesmatt \#altonsterl (-0.74), \#endgunviol (-0.69), \#gunsens (-0.66), \#gunviol (-0.64) \\
\hline
\rowcolor[HTML]{D7FFCE}
lectur (0.55), \#maga (0.55), agenda (0.55), incit (0.55), \#bluelivesmatt \#dallaspoliceshoot (0.55) & \#gunviol (-0.73), program (-0.72), \#philandocastil \#dallaspoliceshoot (-0.67), brother (-0.57), red (-0.57) \\
\hline
\rowcolor[HTML]{FFE8DD}
\#maga (0.65), alabama (0.65), \#wakeupamerica (0.65), mr presid (0.65), miss alabama (0.65) & \#disarmh (-0.74), carri law (-0.74), access (-0.74), famili friend (-0.74), \#stopgunviol (-0.74) \\
\hline
\end{tabular} %
}
\caption{Most partisan phrases per topic for \textbf{Dallas}. Brackets show the $z$-scores of the log odds of each phrase.}
\end{table*}
\begin{table*}[!h]
\centering
\resizebox{\linewidth}{!}{%
\begin{tabular}{|l|l|}
\hline
Republican & Democrat\\ \hline
\rowcolor[HTML]{F2D7D5}
\#orlando (0.46), \#maga (0.46), strike (0.46), 3 dead (0.46), assassin (0.46) & peopl blame (-0.92), crazi (-0.82), polic race (-0.81), blame \#blacklivesmatt (-0.80), report polic (-0.79) \\
\hline
\rowcolor[HTML]{FCF3CF}
\#pjnet (0.52), \#tcot (0.52), realdonaldtrump (0.52), servic (0.52), \#backtheblu (0.52) & polic arriv (-0.79), \#altonsterl (-0.49), msnbc (-0.46), arriv (-0.46), sterl (-0.42) \\
\hline
\rowcolor[HTML]{EAEDED}
killer (0.30), video (0.26), identifi (0.24), gunman (0.16), marin serv (0.12) & veteran (-0.25), jackson (-0.25), mo (-0.25), heartbreak (-0.25), citi mo (-0.25) \\
\hline
\rowcolor[HTML]{D0ECE7}
\#bluelivesmatt (0.72), policemen (0.70), earth (0.64), liber (0.61), order (0.60) & guy gun (-0.56), \#enough (-0.56), thought prayer (-0.56), \#gun (-0.48), good guy (-0.48) \\
\hline
\rowcolor[HTML]{D6EAF8}
shooter sovereign (0.57), \#breakingnew (0.57), fbi (0.38), citi missouri (0.33), la (0.33) & msnbc (-0.59), identifi shooter (-0.46), matter (-0.43), pd (-0.43), live (-0.42) \\
\hline
\rowcolor[HTML]{EBDEF0}
\#backtheblu (0.61), leadership (0.61), \#thinbluelin (0.53), killer (0.51), 2 offic (0.50) & gun violenc (-0.78), \#gunviol (-0.78), \#disarmh (-0.78), \#altonsterl (-0.66), news polic (-0.66) \\
\hline
\rowcolor[HTML]{D7FFCE}
offic shot (0.40), realdonaldtrump (0.37), \#bluelivesmatt (0.35), 7 (0.33), fallen (0.31) & \#altonsterl (-0.62), funer (-0.42), montrel (-0.35), black (-0.30), moment (-0.29) \\
\hline
\rowcolor[HTML]{FFE8DD}
\#trumppence16 (0.61), hillari (0.61), lectur (0.61), hillaryclinton (0.50), assassin (0.50) & \#enough (-0.71), \#nra (-0.67), \#philandocastil (-0.66), unjustifi (-0.64), open carri (-0.58) \\
\hline
\end{tabular} %
}
\caption{Most partisan phrases per topic for \textbf{Baton Rouge}. Brackets show the $z$-scores of the log odds of each phrase.}
\end{table*}
\begin{table*}[!h]
\centering
\resizebox{\linewidth}{!}{%
\begin{tabular}{|l|l|}
\hline
Republican & Democrat\\ \hline
\rowcolor[HTML]{F2D7D5}
turkey (0.19), hillari support (0.19), turkish muslim (0.19), shooter arcan (0.19), immigr turkey (0.19) & gun (-0.47), shoot (-0.45), trump (-0.39), peopl (-0.38), news (-0.29) \\
\hline
\rowcolor[HTML]{FCF3CF}
zone (0.60), free zone (0.60), gun free (0.60), free (0.60), cetin (0.52) & famili (-0.55), watch (-0.53), latest (-0.46), 4 women (-0.32), male (-0.29) \\
\hline
\rowcolor[HTML]{EAEDED}
victim (-0.28) & victim (-0.28) \\
\hline
\rowcolor[HTML]{D0ECE7}
shooter citizen (0.26), illeg (0.26), id (0.26), citizen vote (0.24), 3 elect (0.23) & support (-0.39), peopl (-0.36), gun (-0.35), control (-0.28), law (-0.26) \\
\hline
\rowcolor[HTML]{D6EAF8}
turkish muslim (0.41), hillari (0.41), voter (0.41), clinton (0.41), ed arcan (0.38) & motiv (-0.47), komo (-0.44), gun (-0.44), activ shooter (-0.40), 3 (-0.40) \\
\hline
\rowcolor[HTML]{EBDEF0}
pray (0.34), peopl (0.27), sad (0.24), prayer victim (0.21), washington (0.16) & time (-0.45), gun (-0.44), love (-0.30), affect (-0.30), mass (-0.29) \\
\hline
\rowcolor[HTML]{D7FFCE}
victim (-0.25) & victim (-0.25) \\
\hline
\rowcolor[HTML]{FFE8DD}
attack (0.49), polic (0.40), muslim (0.33), hispan (0.27), live (0.24) & seattl (-0.67), fuck (-0.35), news (-0.31), hope (-0.27), gun (-0.26) \\
\hline
\end{tabular} %
}
\caption{Most partisan phrases per topic for \textbf{Burlington}. Brackets show the $z$-scores of the log odds of each phrase.}
\end{table*}
\begin{table*}[!h]
\centering
\resizebox{\linewidth}{!}{%
\begin{tabular}{|l|l|}
\hline
Republican & Democrat\\ \hline
\rowcolor[HTML]{F2D7D5}
attack (0.24), terrorist (0.20), terrorist attack (0.16), ft airport (0.15), ft (0.10) & shooter (-0.22), white (-0.08), muslim (-0.01), terror (-0.01), obama (0.05) \\
\hline
\rowcolor[HTML]{FCF3CF}
airport victim (0.59), obama (0.47), garag (0.47), realdonaldtrump (0.45), phone (0.42) & custodi 9 (-0.40), stori (-0.37), msnbc (-0.36), report fire (-0.34), sad (-0.31) \\
\hline
\rowcolor[HTML]{EAEDED}
 &  \\
\hline
\rowcolor[HTML]{D0ECE7}
airport gun (0.47), free zone (0.42), gun free (0.42), free (0.38), zone (0.33) & mass (-0.50), \#ftlauderdal (-0.19), peopl (-0.07), gun airport (-0.07), control (-0.06) \\
\hline
\rowcolor[HTML]{D6EAF8}
airport (0.23), shooter (-0.04) & shooter (-0.04), airport (0.23) \\
\hline
\rowcolor[HTML]{EBDEF0}
affect airport (0.41), shooter (0.34), make (0.33), involv ft (0.28), prayer affect (0.27) & gun (-0.42), violenc (-0.41), week (-0.39), stay (-0.37), travel (-0.37) \\
\hline
\rowcolor[HTML]{D7FFCE}
airport (-0.01) & airport (-0.01) \\
\hline
\rowcolor[HTML]{FFE8DD}
shot fire (0.42), fire (0.38), attack (0.33), terror (0.30), man (0.30) & \#airport (-0.37), gunman (-0.32), week (-0.32), stop (-0.27), 2017 (-0.25) \\
\hline
\end{tabular} %
}
\caption{Most partisan phrases per topic for \textbf{Fort Lauderdale}. Brackets show the $z$-scores of the log odds of each phrase.}
\end{table*}
\begin{table*}[!h]
\centering
\resizebox{\linewidth}{!}{%
\begin{tabular}{|l|l|}
\hline
Republican & Democrat\\ \hline
\rowcolor[HTML]{F2D7D5}
islamist (0.18), fake (0.18), fake news (0.18), white male (0.18), facebook (0.18) & mass (-0.64), nation (-0.44), matter (-0.43), tweet (-0.38), support (-0.31) \\
\hline
\rowcolor[HTML]{FCF3CF}
2 (0.37), open (0.34), open fire (0.33), cnn (0.33), muslim (0.33) & 4 (-0.52), offic (-0.44), dyer (-0.33), fatal (-0.26), spree (-0.24) \\
\hline
\rowcolor[HTML]{EAEDED}
2 (0.46), 1 (0.46), california (0.37), kill (0.16) & kill (0.16), california (0.37), 1 (0.46), 2 (0.46) \\
\hline
\rowcolor[HTML]{D0ECE7}
california (0.32), stop (0.23), control (0.18), gun control (0.17), law (0.09) & mental (-0.47), shot (-0.38), shooter (-0.26), kill (-0.07), peopl (-0.02) \\
\hline
\rowcolor[HTML]{D6EAF8}
shooter shout (0.27), translat (0.27), \#maga (0.27), fake (0.27), fake news (0.27) & offic (-0.68), fatal (-0.59), gun (-0.58), connect (-0.54), kori muhammad (-0.40) \\
\hline
\rowcolor[HTML]{EBDEF0}
attack (0.25), love (0.23), 3 (0.21), kill (0.21), prayer (0.10) & sad (-0.40), hate (-0.34), today (-0.23), peopl (-0.16), shoot (-0.11) \\
\hline
\rowcolor[HTML]{D7FFCE}
victim (0.17), kill (-0.03) & kill (-0.03), victim (0.17) \\
\hline
\rowcolor[HTML]{FFE8DD}
yesterday (0.31), terror (0.24), black man (0.23), kill 4 (0.21), target (0.21) & trump (-0.39), spree (-0.35), happen (-0.32), tweet (-0.30), make (-0.29) \\
\hline
\end{tabular} %
}
\caption{Most partisan phrases per topic for \textbf{Fresno}. Brackets show the $z$-scores of the log odds of each phrase.}
\end{table*}
\begin{table*}[!h]
\centering
\resizebox{\linewidth}{!}{%
\begin{tabular}{|l|l|}
\hline
Republican & Democrat\\ \hline
\rowcolor[HTML]{F2D7D5}
cover (0.53), workplac (0.39), trump (0.28), shooter (0.19), sf (0.16) & shoot (-0.40), blame (-0.25), terror (-0.25), today (-0.21), gun (-0.21) \\
\hline
\rowcolor[HTML]{FCF3CF}
4 injur (0.59), issu (0.58), compani (0.55), facil 4 (0.46), fox news (0.44) & barclay (-0.59), cover (-0.59), america (-0.59), 3 kill (-0.52), talk (-0.48) \\
\hline
\rowcolor[HTML]{EAEDED}
4 (0.21), gunfir (0.19), kill 4 (0.19), dead (0.19), worker (0.17) & victim (-0.19), facil includ (-0.15), kill facil (-0.13), time (-0.10), facil (-0.08) \\
\hline
\rowcolor[HTML]{D0ECE7}
assault pistol (0.87), pistol (0.81), stolen (0.73), free (0.72), california (0.71) & congress (-0.47), secur (-0.43), die (-0.39), american (-0.39), gop (-0.38) \\
\hline
\rowcolor[HTML]{D6EAF8}
motiv (0.53), identifi (0.30), assault pistol (0.30), polic (0.28), warehous (0.28) & sfpd (-0.36), suspect (-0.32), \#up (-0.32), lam (-0.31), jimmi (-0.31) \\
\hline
\rowcolor[HTML]{EBDEF0}
polic (0.54), find (0.41), colleagu (0.27), world (0.26), 2 (0.22) & countri (-0.33), gun violenc (-0.23), sf (-0.22), mass (-0.22), peopl (-0.21) \\
\hline
\rowcolor[HTML]{D7FFCE}
sf (-0.13), today (-0.14), victim (-0.15) & victim (-0.15), today (-0.14), sf (-0.13) \\
\hline
\rowcolor[HTML]{FFE8DD}
person (0.49), law (0.49), killer (0.39), grown (0.39), down (0.39) & america (-0.29), shot kill (-0.28), morn (-0.27), place (-0.27), murder (-0.26) \\
\hline
\end{tabular} %
}
\caption{Most partisan phrases per topic for \textbf{San Francisco}. Brackets show the $z$-scores of the log odds of each phrase.}
\end{table*}
\begin{table*}[!h]
\centering
\resizebox{\linewidth}{!}{%
\begin{tabular}{|l|l|}
\hline
Republican & Democrat\\ \hline
\rowcolor[HTML]{F2D7D5}
clinton blame (0.52), cover call (0.52), today cover (0.52), elitist (0.52), \#ccot (0.52) & \#whiteterror (-0.86), time talk (-0.86), patch (-0.86), \#whiteterrorist (-0.86), patch pm (-0.86) \\
\hline
\rowcolor[HTML]{FCF3CF}
shooter long (0.52), long detail (0.52), video point (0.52), wsbtv report (0.52), seanhann (0.52) & global stay (-0.86), jule (-0.86), dhs jule (-0.86), jule rt (-0.86), \#gunsens (-0.86) \\
\hline
\rowcolor[HTML]{EAEDED}
\#dtmag (0.52), \#uscg \#usarmi (0.52), \#uscg (0.52), \#usarmi \#usairforc (0.52), \#usairforc (0.52) & \#lalat (-0.87), ▶ (-0.87), rt time (-0.87), \#lasvegasmassacr stephen (-0.87), \#lalat \#breakingnew (-0.87) \\
\hline
\rowcolor[HTML]{D0ECE7}
gowdi (0.78), massacr arm (0.78), bash gun (0.78), gowdi difficult (0.78), paddock patsi (0.78) & \#tytliv (-0.61), call rep (-0.61), money \#guncontrolnow (-0.61), tweet thought (-0.61), obama era (-0.61) \\
\hline
\rowcolor[HTML]{D6EAF8}
fbi sourc (0.44), make statement (0.44), made lot (0.44), gun make (0.44), life made (0.44) & npr (-0.95), 95 (-0.95), click full (-0.95), white terrorist (-0.87), full stori (-0.86) \\
\hline
\rowcolor[HTML]{EBDEF0}
rt realdonaldtrump (0.71), \#mandalaybay \#nra (0.71), \#bluelivesmatt (0.71), shooter reno (0.71), leftist (0.71) & trophi (-0.68), violenc text (-0.68), \#notonemor (-0.68), mourn tomorrow (-0.68), tomorrow fight (-0.68) \\
\hline
\rowcolor[HTML]{D7FFCE}
injur stand (0.60), impeach postpon (0.60), green trump (0.60), \#dtmag (0.60), stand countri (0.60) & receiv year (-0.79), click full (-0.79), raw stori (-0.79), \#theresist (-0.79), \#bakersfield (-0.79) \\
\hline
\rowcolor[HTML]{FFE8DD}
kain (0.67), tim kain (0.67), kain attack (0.67), room wish (0.67), kaya (0.67) & \#tytliv (-0.72), \#notmypresid (-0.72), senjohnthun (-0.72), prayer stop (-0.72), prayer nice (-0.72) \\
\hline
\end{tabular} %
}
\caption{Most partisan phrases per topic for \textbf{Vegas}. Brackets show the $z$-scores of the log odds of each phrase.}
\end{table*}
\begin{table*}[!h]
\centering
\resizebox{\linewidth}{!}{%
\begin{tabular}{|l|l|}
\hline
Republican & Democrat\\ \hline
\rowcolor[HTML]{F2D7D5}
news (0.60), motiv (0.42), blame (0.34), talk (0.32), tweet (0.31) & live stack (-0.29), furnitur (-0.29), bibl furnitur (-0.29), stack (-0.27), stack bibl (-0.27) \\
\hline
\rowcolor[HTML]{FCF3CF}
detail (0.34), involv (0.34), \#coloradoshoot (0.34), park lot (0.34), \#colorado \#walmart (0.33) & talk (-0.67), \#guncontrolnow (-0.58), control (-0.55), ago (-0.48), white (-0.46) \\
\hline
\rowcolor[HTML]{EAEDED}
vega (-0.05) & vega (-0.05) \\
\hline
\rowcolor[HTML]{D0ECE7}
free (0.49), crime (0.48), \#walmart (0.44), black (0.44), skill (0.33) & talk gun (-0.48), violenc (-0.48), place (-0.38), las (-0.37), las vega (-0.36) \\
\hline
\rowcolor[HTML]{D6EAF8}
updat (0.52), danger (0.44), kill peopl (0.43), \#break (0.41), drive (0.40) & week (-0.50), delay (-0.50), white guy (-0.46), time (-0.45), guy (-0.44) \\
\hline
\rowcolor[HTML]{EBDEF0}
pray (0.39), time (0.36), night (0.30), peopl (0.21), love (0.18) & vega (-0.41), tonight (-0.32), dead (-0.22), hear (-0.18), heart (-0.17) \\
\hline
\rowcolor[HTML]{D7FFCE}
shot (0.21), kill (-0.25) & kill (-0.25), shot (0.21) \\
\hline
\rowcolor[HTML]{FFE8DD}
skill (0.78), chicago (0.75), lol (0.64), cover (0.61), friend (0.50) & open fire (-0.35), reason (-0.28), \#guncontrol (-0.27), hispan (-0.25), man kill (-0.24) \\
\hline
\end{tabular} %
}
\caption{Most partisan phrases per topic for \textbf{Thornton}. Brackets show the $z$-scores of the log odds of each phrase.}
\end{table*}
\begin{table*}[!h]
\centering
\resizebox{\linewidth}{!}{%
\begin{tabular}{|l|l|}
\hline
Republican & Democrat\\ \hline
\rowcolor[HTML]{F2D7D5}
lunat (0.48), handler blame (0.48), countri music (0.48), sudanes (0.48), chelseahandl (0.48) & \#thoughtsandpray (-0.91), 4 black (-0.91), black girl (-0.83), studi (-0.82), bibl studi (-0.82) \\
\hline
\rowcolor[HTML]{FCF3CF}
\#antifa (0.62), \#theresist (0.62), \#nov4thitbegin (0.62), \#uniteblu (0.62), \#atheist (0.62) & \#enough (-0.64), \#gunsens (-0.63), mall (-0.63), \#gun (-0.61), \#nra (-0.60) \\
\hline
\rowcolor[HTML]{EAEDED}
\#usmarin \#tricar (0.60), \#tricar (0.60), \#usnavi \#usmarin (0.60), \#usmarin (0.60), \#usairforc \#usnavi (0.60) & \#guncontrolnow (-0.79), york time (-0.55), nightclub (-0.52), npr (-0.46), massacr texa (-0.41) \\
\hline
\rowcolor[HTML]{D0ECE7}
stop legal (0.74), \#antifa (0.74), joe biden (0.74), shirt (0.74), democrat kill (0.74) & msnbc (-0.64), spineless (-0.64), earli talk (-0.64), prayer good (-0.64), prayer thought (-0.64) \\
\hline
\rowcolor[HTML]{D6EAF8}
fan (0.58), \#antifa (0.58), \#usairforc (0.58), gun stop (0.58), texa hero (0.58) & black girl (-0.81), klan (-0.81), bomber (-0.81), success (-0.81), birmingham church (-0.81) \\
\hline
\rowcolor[HTML]{EBDEF0}
frank (0.69), fbc (0.66), agenda (0.61), polit agenda (0.59), involv church (0.59) & relax (-0.69), gop (-0.69), governor church (-0.69), liter pray (-0.69), epidem (-0.69) \\
\hline
\rowcolor[HTML]{D7FFCE}
ted (0.54), lieu (0.54), ted lieu (0.54), song (0.54), app (0.54) & gun violenc (-0.68), thought (-0.62), \#guncontrolnow (-0.61), great (-0.58), medic (-0.58) \\
\hline
\rowcolor[HTML]{FFE8DD}
tn church (0.75), biden man (0.75), church tn (0.75), hero stephen (0.75), septemb (0.75) & kill girl (-0.64), 4 black (-0.64), \#impeachtrump (-0.64), \#nraterror (-0.64), elementari school (-0.64) \\
\hline
\end{tabular} %
}
\caption{Most partisan phrases per topic for \textbf{Sutherland Springs}. Brackets show the $z$-scores of the log odds of each phrase.}
\end{table*}
\begin{table*}[!h]
\centering
\resizebox{\linewidth}{!}{%
\begin{tabular}{|l|l|}
\hline
Republican & Democrat\\ \hline
\rowcolor[HTML]{F2D7D5}
bush (0.76), student push (0.76), dirti (0.76), page (0.76), boston bomber (0.76) & attack survivor (-0.62), \#trumprussia (-0.62), \#russianbot (-0.62), \#impeachtrump (-0.62), \#votethemout (-0.62) \\
\hline
\rowcolor[HTML]{FCF3CF}
\#emet news (0.65), \#emet (0.65), news press (0.65), broadcast \#emet (0.65), \#new broadcast (0.65) & \#enough (-0.74), result death (-0.74), fuck (-0.66), veloc (-0.66), high veloc (-0.63) \\
\hline
\rowcolor[HTML]{EAEDED}
redondo (0.73), california high (0.73), expos (0.73), absolut (0.73), app (0.67) & make (-0.66), facebook post (-0.66), anonym (-0.66), washington post (-0.57), midterm (-0.56) \\
\hline
\rowcolor[HTML]{D0ECE7}
media politic (0.87), \#greatawaken (0.87), \#antifa (0.87), back window (0.87), \#noguncontrol (0.87) & \#throwthemout (-0.51), vote candid (-0.51), dollar nra (-0.51), \#nrabloodmoney \#nraisaterroristorgan (-0.51), concert goer (-0.51) \\
\hline
\rowcolor[HTML]{D6EAF8}
\#emet news (0.54), \#emet (0.54), news press (0.54), broadcast \#emet (0.54), \#new broadcast (0.54) & \#maga hat (-0.84), wear maga (-0.76), grade (-0.73), cap (-0.73), hat (-0.72) \\
\hline
\rowcolor[HTML]{EBDEF0}
\#qanon (0.89), join pray (0.79), \#pray (0.70), pastor (0.68), heaven (0.68) & blah (-0.50), \#gunreformnow \#neveragain (-0.50), admiss (-0.50), blah blah (-0.50), \#nraboycott (-0.50) \\
\hline
\rowcolor[HTML]{D7FFCE}
\#emet news (0.82), \#emet (0.82), news press (0.82), \#new broadcast (0.82), broadcast \#emet (0.82) & support \#marchforourl (-0.56), \#marchforourl \#boycottnra (-0.56), refus accept (-0.56), \#banassaultweapon (-0.56), \#gunsensenow (-0.56) \\
\hline
\rowcolor[HTML]{FFE8DD}
\#liberalismisamentaldisord (0.89), \#firesheriffisrael (0.89), school 2015 (0.89), \#gunfreezoneskil (0.89), \#thegreatawaken (0.89) & \#stopthenra (-0.49), prayer work (-0.49), \#nomorethoughtsandpray (-0.49), \#standwithparkland (-0.49), black brown (-0.49) \\
\hline
\end{tabular} %
}
\caption{Most partisan phrases per topic for \textbf{Parkland}. Brackets show the $z$-scores of the log odds of each phrase.}
\end{table*}
\begin{table*}[!h]
\centering
\resizebox{\linewidth}{!}{%
\begin{tabular}{|l|l|}
\hline
Republican & Democrat\\ \hline
\rowcolor[HTML]{F2D7D5}
left (0.76), liber (0.58), democrat (0.55), anti (0.44), blame (0.42) & school (-0.34), wait (-0.34), church (-0.34), hous attack (-0.34), 4 peopl (-0.34) \\
\hline
\rowcolor[HTML]{FCF3CF}
polic suspect (0.49), hold (0.48), east (0.45), fox news (0.43), pd (0.42) & abc news (-0.46), rt (-0.43), death (-0.43), tweet (-0.35), citi (-0.35) \\
\hline
\rowcolor[HTML]{EAEDED}
gun (0.32), waffl hous (0.06), hous (0.06), waffl (0.05), worker (-0.07) & diploma (-0.43), mother (-0.33), cbs (-0.30), cbs news (-0.30), kill (-0.28) \\
\hline
\rowcolor[HTML]{D0ECE7}
democrat (0.56), agenda (0.56), gun seiz (0.54), gun confisc (0.52), free zone (0.50) & assault style (-0.73), 4 dead (-0.63), terrorist (-0.61), thought prayer (-0.60), gop (-0.60) \\
\hline
\rowcolor[HTML]{D6EAF8}
come (0.75), awar (0.63), antioch waffl (0.62), 6 (0.54), suspect 2 (0.49) & plastic (-0.64), mass murder (-0.57), domest (-0.56), \#guncontrolnow (-0.54), murder 4 (-0.54) \\
\hline
\rowcolor[HTML]{EBDEF0}
student (0.41), colleg (0.37), opri mill (0.37), opri (0.37), mill (0.37) & condol victim (-0.51), massacr (-0.41), hous massacr (-0.41), rest (-0.38), hous tn (-0.37) \\
\hline
\rowcolor[HTML]{D7FFCE}
antioch waffl (0.69), visit (0.62), antioch (0.59), held (0.58), song (0.43) & hous white (-0.42), white shooter (-0.42), volum (-0.42), cnn (-0.41), hero white (-0.40) \\
\hline
\rowcolor[HTML]{FFE8DD}
robberi (0.77), goal (0.73), liber (0.72), pred (0.69), \#2a (0.69) & sight (-0.40), theater (-0.40), justic (-0.40), movi theater (-0.40), effort (-0.40) \\
\hline
\end{tabular} %
}
\caption{Most partisan phrases per topic for \textbf{Nashville}. Brackets show the $z$-scores of the log odds of each phrase.}
\end{table*}
\begin{table*}[!h]
\centering
\resizebox{\linewidth}{!}{%
\begin{tabular}{|l|l|}
\hline
Republican & Democrat\\ \hline
\rowcolor[HTML]{F2D7D5}
leftist (0.91), narrat (0.81), \#2a (0.78), left (0.70), control (0.67) & creat (-0.48), nut (-0.48), male (-0.42), immigr (-0.39), white male (-0.39) \\
\hline
\rowcolor[HTML]{FCF3CF}
school victim (0.54), femal (0.51), road (0.49), patient (0.49), fox news (0.42) & \#gunreformnow (-0.65), \#gunsens (-0.65), 22 school (-0.65), \#resist (-0.65), \#gunreform (-0.65) \\
\hline
\rowcolor[HTML]{EAEDED}
video (0.40), famili (0.33), star (0.21), mass (0.21), student (0.18) & parkland survivor (-0.55), dead (-0.21), david (-0.21), hogg (-0.21), news (-0.20) \\
\hline
\rowcolor[HTML]{D0ECE7}
leftist (0.83), rack (0.83), gun rack (0.82), pickup (0.80), truck (0.79) & terrorist organ (-0.51), address gun (-0.51), prayer stop (-0.51), \#ifidieinaschoolshoot (-0.51), 16th (-0.51) \\
\hline
\rowcolor[HTML]{D6EAF8}
\#khou11 (0.74), nikola cruz (0.53), nikola (0.53), coach (0.52), watch (0.51) & minut (-0.65), shot kill (-0.55), nra (-0.54), loesch (-0.53), dana (-0.52) \\
\hline
\rowcolor[HTML]{EBDEF0}
secur (0.77), polic offic (0.67), brave (0.63), pray high (0.61), personnel (0.59) & \#votethemout (-0.51), \#nraisaterroristorgan (-0.51), action gun (-0.51), custodi (-0.51), fun (-0.51) \\
\hline
\rowcolor[HTML]{D7FFCE}
\#txlege (0.62), kill 10 (0.57), pres (0.50), church (0.46), roundtabl (0.43) & \#guncontrolnow (-0.70), show high (-0.70), wave (-0.70), school scene (-0.70), \#enoughisenough (-0.70) \\
\hline
\rowcolor[HTML]{FFE8DD}
gun rack (0.89), dixon (0.86), rack (0.83), 38 revolv (0.83), truck (0.82) & entitl (-0.42), incel (-0.42), ritalin (-0.42), supremacist (-0.42), white supremacist (-0.42) \\
\hline
\end{tabular} %
}
\caption{Most partisan phrases per topic for \textbf{Santa Fe}. Brackets show the $z$-scores of the log odds of each phrase.}
\end{table*}
\begin{table*}[!h]
\centering
\resizebox{\linewidth}{!}{%
\begin{tabular}{|l|l|}
\hline
Republican & Democrat\\ \hline
\rowcolor[HTML]{F2D7D5}
reuter editor (0.85), apolog disciplin (0.78), disciplin blame (0.78), disciplin (0.78), gun control (0.76) & \#guncontrolnow (-0.54), enemi american (-0.49), multipl (-0.45), call enemi (-0.45), domest (-0.44) \\
\hline
\rowcolor[HTML]{FCF3CF}
pray (0.57), alert (0.54), prayer victim (0.53), depart wisconsin (0.51), wisconsin (0.51) & milo (-0.60), gun journalist (-0.60), yiannopoulo (-0.60), milo yiannopoulo (-0.60), start gun (-0.60) \\
\hline
\rowcolor[HTML]{EAEDED}
victim (0.51), editor (0.39), hiaasen (0.33), rob hiaasen (0.32), 5 (0.28) & kill (-0.15), newspap (-0.07), mass (-0.02), newsroom (-0.01), capit (0.11) \\
\hline
\rowcolor[HTML]{D0ECE7}
person (0.51), shotgun (0.49), crime (0.48), arm (0.47), mental health (0.47) & journal (-0.52), continu (-0.43), free press (-0.42), freedom (-0.40), \#gunreformnow (-0.40) \\
\hline
\rowcolor[HTML]{D6EAF8}
hat (0.59), break (0.46), year jarrod (0.45), gazett kill (0.43), counti (0.39) & aliv (-0.55), accus shooter (-0.50), exclus (-0.47), dark link (-0.46), exclus accus (-0.46) \\
\hline
\rowcolor[HTML]{EBDEF0}
scene (0.54), blame (0.51), conscienc (0.47), address (0.47), rt (0.47) & \#guncontrolnow (-0.44), congress (-0.44), \#neveragain (-0.44), remind (-0.44), report capit (-0.44) \\
\hline
\rowcolor[HTML]{D7FFCE}
sander (0.59), game (0.59), speechless (0.56), opinion (0.53), build (0.49) & press enemi (-0.39), mind (-0.39), milo (-0.39), gazett mass (-0.39), yiannopoulo (-0.39) \\
\hline
\rowcolor[HTML]{FFE8DD}
blame trump (0.74), motiv shooter (0.63), bring (0.58), \#fakenew (0.57), narrat (0.56) & start gun (-0.45), victim capit (-0.45), shot dead (-0.45), journalist gun (-0.38), nra (-0.38) \\
\hline
\end{tabular} %
}
\caption{Most partisan phrases per topic for \textbf{Annapolis}. Brackets show the $z$-scores of the log odds of each phrase.}
\end{table*}
\begin{table*}[!h]
\centering
\resizebox{\linewidth}{!}{%
\begin{tabular}{|l|l|}
\hline
Republican & Democrat\\ \hline
\rowcolor[HTML]{F2D7D5}
trump cnn (0.88), sarsour (0.88), cnn press (0.88), rabbi blame (0.84), camerota (0.81) & \#magabomb (-0.51), black church (-0.51), spread (-0.51), long (-0.51), uncl (-0.47) \\
\hline
\rowcolor[HTML]{FCF3CF}
kill 6 (0.61), dozen (0.60), treat (0.59), synagogu pray (0.57), read (0.57) & arm (-0.51), white (-0.47), ⁦ (-0.47), morn report (-0.45), respond saturday (-0.45) \\
\hline
\rowcolor[HTML]{EAEDED}
polit insid (0.91), polit (0.91), defend (0.86), insid (0.79), survivor life (0.69) & site (-0.48), come (-0.48), stay (-0.47), tell (-0.47), offici (-0.47) \\
\hline
\rowcolor[HTML]{D0ECE7}
democrat (0.94), final (0.90), danger (0.77), free (0.76), understand (0.73) & rip (-0.35), blame victim (-0.35), avoid (-0.35), vote (-0.31), weak (-0.31) \\
\hline
\rowcolor[HTML]{D6EAF8}
pull (0.87), synagogu age (0.51), thursday (0.49), warn (0.49), expect (0.43) & includ 11 (-0.52), trail anti (-0.43), white man (-0.43), long (-0.42), review (-0.42) \\
\hline
\rowcolor[HTML]{EBDEF0}
target kill (0.88), horrif target (0.74), netanyahu (0.65), kill jewish (0.61), murder attack (0.57) & holocaust (-0.51), angri (-0.51), gun violenc (-0.48), rabbi synagogu (-0.48), synagogu call (-0.48) \\
\hline
\rowcolor[HTML]{D7FFCE}
michigan host (0.69), univers michigan (0.69), jihad day (0.69), palestinian jihad (0.69), advoc palestinian (0.69) & pictur (-0.70), add (-0.70), victim hate (-0.70), hebrew (-0.70), meltdown (-0.67) \\
\hline
\rowcolor[HTML]{FFE8DD}
wrong (0.45), wait (0.45), mind (0.41), kill 11 (0.41), lot (0.40) & insid (-0.40), templ (-0.40), liter (-0.33), black (-0.32), synagogu victim (-0.31) \\
\hline
\end{tabular} %
}
\caption{Most partisan phrases per topic for \textbf{Pittsburgh}. Brackets show the $z$-scores of the log odds of each phrase.}
\end{table*}
\begin{table*}[!h]
\centering
\resizebox{\linewidth}{!}{%
\begin{tabular}{|l|l|}
\hline
Republican & Democrat\\ \hline
\rowcolor[HTML]{F2D7D5}
leftist (0.84), anti gun (0.84), anti (0.77), left (0.74), liber (0.66) & violenc women (-0.55), republican (-0.49), women (-0.48), histori (-0.47), link (-0.47) \\
\hline
\rowcolor[HTML]{FCF3CF}
forc (0.65), insid california (0.57), want (0.57), california wildfir (0.53), camp (0.51) & mass 12 (-0.61), 12 victim (-0.61), gun violenc (-0.47), hour (-0.42), violenc (-0.42) \\
\hline
\rowcolor[HTML]{EAEDED}
knelt block (0.84), wit men (0.84), knelt (0.84), real men (0.84), men knelt (0.84) & lane (-0.55), stay lane (-0.55), doctor stay (-0.55), nra told (-0.55), offer thought (-0.55) \\
\hline
\rowcolor[HTML]{D0ECE7}
divers (0.86), steal (0.86), 98 mass (0.86), communist divers (0.86), die communist (0.86) & \#gunviol (-0.52), 307th (-0.52), \#enoughisenough (-0.52), 307th mass (-0.52), marshablackburn (-0.52) \\
\hline
\rowcolor[HTML]{D6EAF8}
ball (0.77), earlier (0.77), commit gunman (0.77), gunman ian (0.77), clear (0.70) & women (-0.50), week (-0.49), men (-0.49), white men (-0.49), terrorist (-0.48) \\
\hline
\rowcolor[HTML]{EBDEF0}
men (0.88), involv california (0.86), strictest gun (0.75), gt gt (0.72), strictest (0.72) & \#gunviol (-0.41), gun amp (-0.41), access gun (-0.41), mass place (-0.41), gunman kill (-0.41) \\
\hline
\rowcolor[HTML]{D7FFCE}
proclam (0.80), proclam honor (0.80), presidenti proclam (0.80), presidenti (0.80), tragedi california (0.63) & warm (-0.56), share (-0.56), \#gunviol (-0.56), buck wear (-0.56), plan (-0.49) \\
\hline
\rowcolor[HTML]{FFE8DD}
california strictest (0.84), liber (0.84), strictest (0.81), left (0.81), california gun (0.81) & blackburn (-0.45), marsha (-0.45), mass die (-0.45), live gun (-0.45), gop (-0.45) \\
\hline
\end{tabular} %
}
\caption{Most partisan phrases per topic for \textbf{Thousand Oaks}. Brackets show the $z$-scores of the log odds of each phrase.}
\end{table*}

\clearpage\section{Personal Pronouns} 
\label{sec:appendix_pronouns}

Pronoun usage has been found to express different degrees of self-reflection, awareness, inclusivity, perspective-taking, among several other psychological phenomena, both on Twitter \cite{qiu2012you} and in various other domains \cite{hirsh2009personality, yarkoni2010personality, pennebaker2011secret}. We rely on these findings to treat pronouns as proxies to measure polarization in terms of users' personalization of the experience (1st person sg.), inclusion  (1st person pl.), deflection (3rd person).

\subsection{Methods}
To quantify the partisanship of personal pronouns, we take the five personal pronoun categories in LIWC \cite{pennebaker2001linguistic} (I, You, We, SheHe, They) and then calculate their partisan log-odds ratio.

\subsection{Results}

The log odds of pronouns suggest that \emph{first and second person pronouns} are significantly more likely to be used by Democrats across the events, while \emph{They} is somewhat more likely to be used by Democrats and \emph{SheHe} is used similarly by the two parties: \emph{I} (mean:$-.26$, $p < .001$), \emph{We} (mean:$-.26$, $p < .001$), \emph{You} (mean:$-.13$, $p < 0.05$), \emph{They} (mean:$-.06$, $p < .1$), \emph{SheHe} (mean:$.05$, $p \approx0.36$) (see Figure~\ref{fig:pronoun_distr}). 

The use of two pronouns is significantly different based on the shooter's race: \emph{SheHe} and \emph{You} are both more likely to be used by Democrats when the shooter is white and by Republicans if the shooter is a person of color ($p<0.01$ for \emph{SheHe} and $p<0.05$ for \emph{You} from two-tailed t test). The pattern pertaining to \emph{SheHe} might be partly explained by differential mentions of the president (see Appendix~\ref{sec:appendix_partisan_words} with the most partisan words), since it so happens that most of the events where the shooter was a person of color occurred under Obama's presidency while the ones where the shooter was white predominantly occurred under Trump's presidency. 
\begin{figure}[]
 \centering
   \centering
   \includegraphics[width=.9\linewidth]{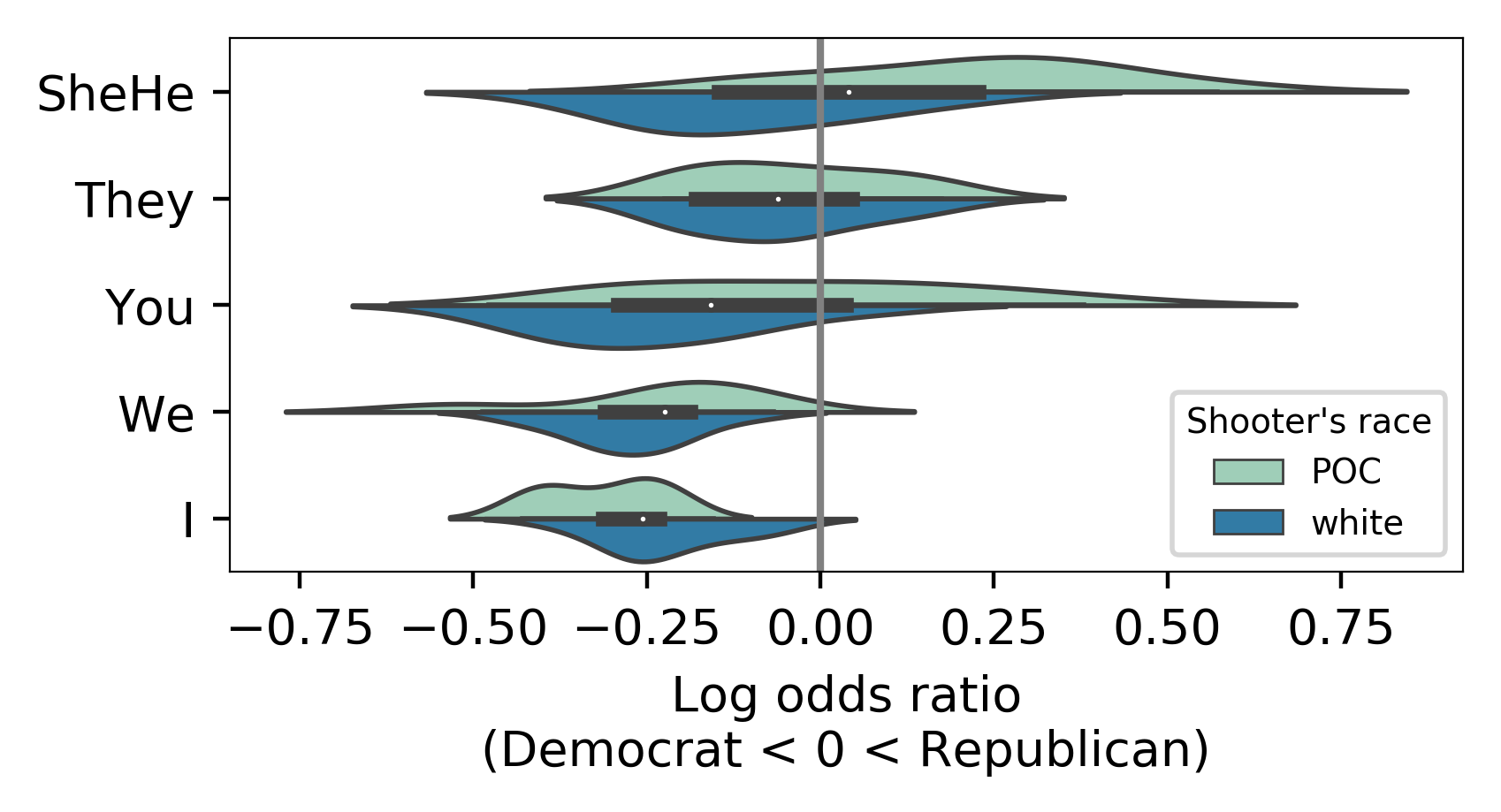}
   \caption{The partisan log odds ratio of pronouns.}
   \label{fig:pronoun_log_odds}
\end{figure}

To better understand this result, we are again interested to see if there is a link between pronoun usage and topic preferences --- we use the same procedure to measure the representation of pronouns in topics as in the case of modals. Our findings (see Figure~\ref{fig:pronoun_distr}) show that first \emph{I} predominantly occurs in \emph{solidarity} and \emph{other}, which, coupled with previous findings about these topics being preferred by Democrats and about affect, suggest that Democrats in our dataset are more likely to personalize their tweets and write about their own mental state and feelings towards the event and the victims. Similarly \emph{We} is overrepresented in \emph{laws \& policy}, also a topic that is preferred by Democrats, which, building on our results about modals (Section~\ref{ssec:exp5_results}), provide evidence that Democrats are more likely to call for collective action.

\emph{SheHe}, on the other hand, are most frequent in \emph{investigation}, \emph{shooter's identity \& ideology} and \emph{victims \& location} --- topics that are more likely to be discussed by Republicans. This result, supported by our finding about affect, suggests that Republicans in this domain are more likely to author tweets that focus on a third person (e.g. the shooter). 

\begin{figure}[]
 \centering
   \centering
   \includegraphics[width=\linewidth]{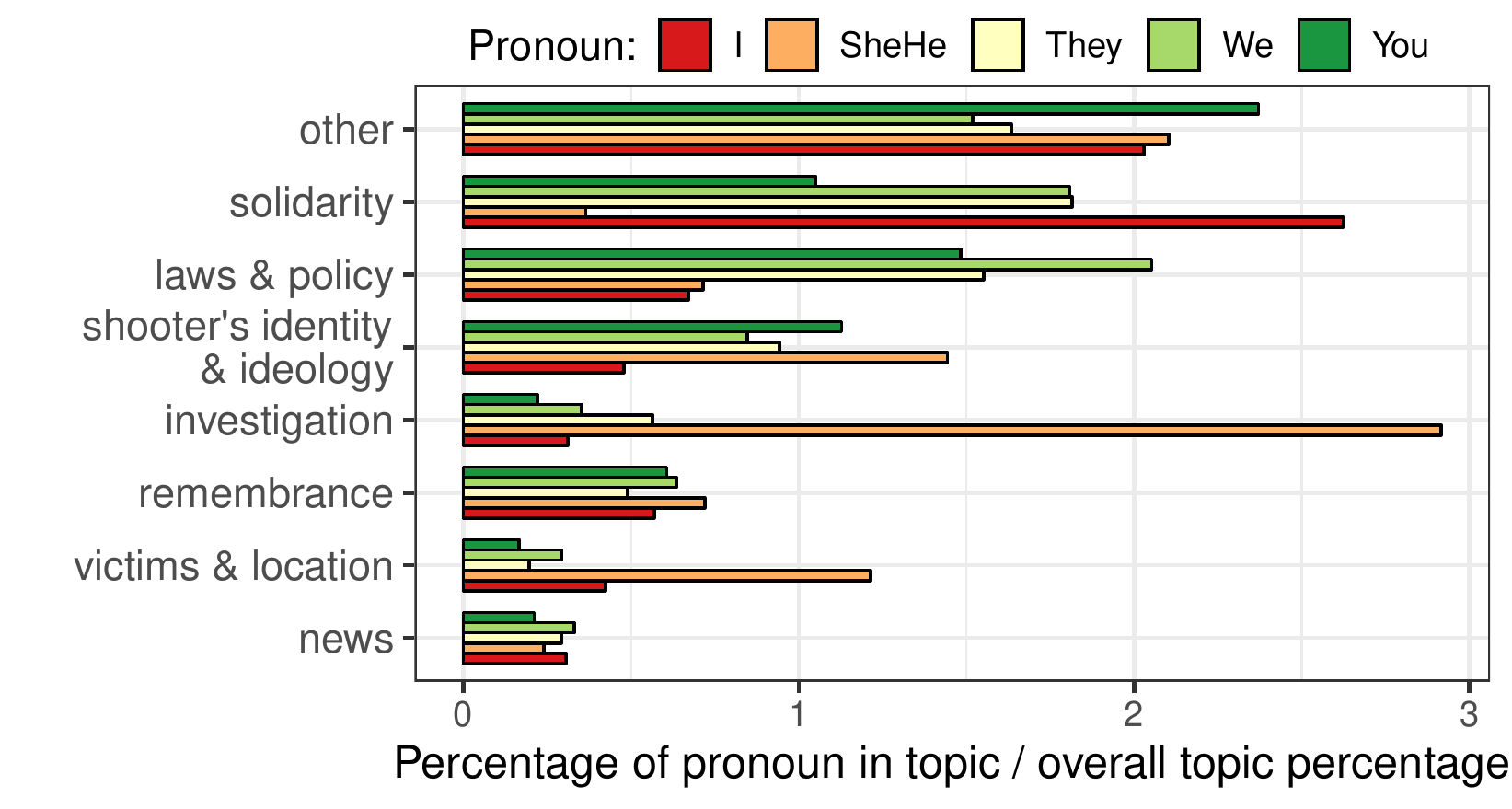}
   \caption{Distribution of pronouns across topics (mean across 21 events).}
   \label{fig:pronoun_distr}
\end{figure}

\clearpage\section{Modal Collocations}
\label{sec:appendix_modals}

We study the subjects and complements of modals and their partisanship to get a better view of \emph{how} these modals are used. We calculate the $z$-scored partisan log odds ratio of the collocations (subject + modal + complement constructions) at the event- and modal-level. We keep all collocations that are more likely to be used by Democrats or Republicans by at least $.5$ SD of the event- and modal-level log odds ratios (in other words, whose $z$-score has an absolute value that is at least $.5$). In the list below, we show those collocations that are partisan in the same direction (Democrat or Republican) in at least three events.

Note that before calculating the log odds ratio, we replace contracted forms with their non-contracted form (e.g. ``shouldn't'' with ``should not'', ``should've'' with ``should have''). The number following the collocations is the number of events for which a particular term is partisan towards a given party. Often when a collocation seems ungrammatical, it is because it is comes from a question (e.g. ``long must we'' $\rightarrow$ ``how long must we''). 

Note that the patterns in the collocations accord with the findings discussed in the paper. Democrats are more likely to use modals to call for collective and proactive action (e.g. ``we must do'',  ``we have to do'', ``something needs to be done'', ``we need to act'') and also to express emotion (e.g. ``[why do] people have to die'', ``[why do] people need to die'') than Republicans. Republicans are more likely to use modals epistemically (e.g. ``it must have been'') and in other, idiomatic, senses that do not imply necessity (e.g. ``it has to do [with]'', ``I have to say'', ``I must say''). Republicans are less likely to use modals in contexts with first person plural subjects (``we'') than Democrats, but are more likely to use ``we'' in contexts where the modal's prejacent implies ``stopping'' something rather than ``doing'' something (e.g. ``we must ban'', ``we must protect'', ``we should ban'', ``we need to protect'').

\begin{center}\textbf{MUST}\end{center}
\noindent\emph{Democrat}\\we must do (14), this must stop (8), we must end (7), something must be done (7), we must act (7), long must we (7), congress must act (7), we must stand (6), we must stop (6), he must be white (6), we must address (6), violence must end (6), violence must stop (6), lives must be lost (6), killing must stop (6), we must make (6), times must this (5), it must stop (5), we must all (5), people must die (5), many must die (5), we must keep (5), and must do (5), we must pass (5), we must continue (5), times must we (5), this must end (5), we must take (5), we must honor (5), shooter must be white (5), shooter must have been (4), insanity must stop (4), we must treat (4), we must remember (4), shootings must stop (4), is a must  (4), shootings must end (4), change must happen (4), congress must take (4), this must change (4), we must get (4), action must be taken (4), control must happen (4), we must push (4), we must find (4), we must have gun (4), we must change (4), we must work (4), we must combat (4), we must deal (4), lives must be taken (4), we must try (4), congress must pass (4), madness must end (4), we must hold (4), \#gunviolence must end (4), we must enact (4), rifles must be banned (4), they must know (3), i must say (3), we must ask (3), we must always (3), \#guncontrol must happen (3), we must reclaim (3), madness must stop (3), it must take (3), there must have been (3), action must follow (3), we must condemn (3), i must ask (3), we must \#stopgunviolence (3), innocents must die (3), nra must be so (3), why must we (3), we must come (3), we must fight (3), things must change (3), tragedies must end (3), violence must be stopped (3), we must confront (3), violence must be addressed (3), we must stay (3), you must be so (3), laws must change (3), we must  (3), shootings must we (3), we must join (3), you must really (3), weapons must be banned (3), we must have change (3), love must prevail (3), shooting must have been (3), more must we (3), we must not become (3), we must create (3), we must allow (3), children must die (3), you must be proud (3), that must mean (3), we must fix (3), shooting must stop (3), we must look (3),  must feel (3), we must as (3), more must die (3), there must be something (3), you must do (3), and must be stopped (3), we must talk (3), we must not forget (3), we must vote (3)\\
\noindent\emph{Republican}\\it must have been (6), they must be stopped (5),  must watch (5), i must say (5), we must ban (5), they must know (4), you must know (4), we must protect (4), we must remain (4), i must have missed (4), he must resign (4), why must you (4), you must protect (4), we must be vigilant (4), attacks must stop (4), you must be happy (4), we must reform (4), it must be stopped (4), this must be one (4), rhetoric must stop (4), we must understand (4), we must stand (4), we must return (4)\\
\begin{center}\textbf{SHOULD}\end{center}
\noindent\emph{Democrat}\\this should not be normal (8), we should all (7), who should not have guns (6), who should not have them (6), this should not be happening (6), this should not happen (6), who should not have had (6), this should not have happened (5), we should just (5), you should be ashamed (5), congress should be forced (5), that should not be happening (5), people should be able (5), we should never (5), who should not have a gun (5), civilians should not have access (5), we should know (5), you should see (5), never should have had (5), everyone should be able (4), we should ban (4), we should start (4), you should  (4), we should probably (4),  should we (4), you should not be able (4), we should bring (4), one should die (4), there should be some (4), we should not do (4), we should wait (4), this should never (4), we should not need (4), it should not be this (4), one should be able (4), we should not be afraid (4), everyone should have a gun (4), civilians should have access (4), they should have had (4), we should now (4), people should not have access (4), one should be afraid (4), he should not have had (4), this should not be our (4), we should try (4), one should live (4), we should not allow (4), hook should have been (4), people should have access (4), everyone should have guns (4), everyone should own (4), it should not take (4), this should have never (4), guns should be allowed (4), there should be a ban (3), media should be ashamed (3), we should change (3), people should be more (3), you should ask (3), we should politicize (3), you should never (3), i should have known (3), shootings should not be the norm (3), we should do (3), we should be upset (3), we should be afraid (3), there should have been (3), there should never (3), it should not be that (3), we should be focused (3), one should have access (3), we should not care (3), you should also (3), you should tweet (3), something should be done (3),  should not they (3), it should have been (3), we should not be surprised (3), they should know (3), there should be more (3), why should it (3), you should call (3), congress should do (3), he should go (3), why should anyone (3), people should not be shot (3), guns should be outlawed (3), one should feel (3), weapons should be outlawed (3), rifles should be banned (3), guns should be illegal (3), people should not be killed (3), never should have happened (3), one should have to fear (3), people should feel (3), we should leave (3), what should be a safe (3), guns should be legal (3), people should be allowed (3), anyone should be able (3), media should stop (3), we should remember (3), but should not we (3), that should be illegal (3), they should feel (3), you should all (3), one should lose (3), you should too (3), civilian should have access (3), we should focus (3), we should  (3), person should not be able (3), you should donate (3),  should i (3), one should be getting (3), weapons should not be sold (3), everyone should have the right (3), guns should never (3), civilians should not have assault (3), people should not have to worry (3), you should not be allowed (3), shooting should not happen (3), we should stand (3), you should be embarrassed (3), we should build (3), you should leave (3), you should be more (3), we should not be able (3), what should you (3), you should not own (3), civilians should be able (3), guns should not be allowed (3), one should have to worry (3), we should have gun (3), he should never (3), everyone should read (3), we should not rush (3), it should not be allowed (3), we should say (3), all should  (3), this should not be a political (3), he should be able (3), we should be having (3), people should not be able (3), i should not feel (3), they should put (3), man should not have been (3), civilian should ever (3), it should not be lost (3), civilians should not own (3), weapons should be legal (3), i should never (3), people should have died (3), fbi should investigate (3), we should attack (3), when should we (3), parent should ever (3), we should name (3), shooting should of (3), u should get (3), civilians should own (3), we should arm (3), it should be for (3), teachers should have guns (3), nra should have to pay (3), you should not have the right (3), shooting should have never (3), they should have been (3), we should really (3), it should not matter (3), who should get (3), that should not have guns (3), can should do (3), we should discuss (3), it should  (3), one should be shot (3),  should people (3), we should treat (3), we should not talk (3), they should  (3), rifles should not be available (3), someone should tell (3), you should be shot (3), that should have been (3), we should not have laws (3), people should not be allowed (3), it should not even (3), or should we (3), civilian should own (3), they should call (3), you should be tweeting (3), one should have to go (3), that should be safe (3), that should never (3), we should definitely (3), child should ever (3), somebody should tell (3), nothing should be done (3), we should not let (3)\\
\noindent\emph{Republican}\\we should ban (8), we should make (7), and should be prosecuted (6), they should just (6), police should have guns (5), i should have been (5), you should know (5), they should of (5), you should go (5), he should have been (5), we should keep (5), he should be arrested (5), they should not be allowed (5), government should have guns (5), we should see (5),  should we (5), they should give (4), they should be allowed (4), everyone should know (4), we should at (4), they should make (4), shooting should not we (4), obama should resign (4), someone should inform (4), we should outlaw (4), you should do (4), you should ask (4), he should be fired (4), us should take (4), you should be fired (4), what should be done (4), fbi should have been (4), they should  (4), why should he (4), and should have been (4), we should be asking (4), we should have guns (4), parents should be held (4), he should have stopped (4), we should add (4), he should take (4), and should be fired (4), they should use (4), democrats should not be allowed (4), they should hang (4), you should be mad (4), we should work (4), what should they (4), they should take (4), we should be able (4), guns should be taken (4), you should stop (4), you should see (4), there should never (4), shooter should get (4), who should we (4), shooter should not have had (4)\\
\begin{center}\textbf{NEED TO}\end{center}
\noindent\emph{Democrat}\\we need to take (8), more need to die (8), we need to act (8), we need to talk (8), we need to do (8), we need to stand (7), we need to stop (7), people need to die (7), shootings need to happen (6), laws need to change (6), we need to change (6), we need to fix (6), we need to vote (6), we need to make (6), we need to call (5), they need to carry (5), we need to remember (5), don't need to see (5), all need to take (5), we need to hear (4), you need to rethink (4), we need to be better (4), we need to start (4), guns need to go (4), we need to figure (4), people need to stop (4), you need to change (4), we need to keep (4), we need to solve (4), laws need to be changed (4), we need to work (4), we need to end (4), we need to be able (4), we need to address (4), we need to help (4), we need to demand (4), weapons need to be banned (4), we need to have more (4), i need to know (4), lives need to be lost (3), acts need to stop (3), you need to know (3), we need to focus (3), don't need to keep (3), many need to die (3), we need to have before (3), we need to look (3), we need to say (3), we need to see (3), we need to pass (3), massacres need to happen (3), you need to recognize (3), also need to make (3), this need to happen (3), lives need to be taken (3), we need to \#endgunviolence (3), all need to stand (3), we need to acknowledge (3), we need to mourn (3), shootings need to stop (3), don't need to know (3), violence need to end (3), we need to have better (3), we need to live (3), laws need to happen (3), violence need to stop (3), they need to know (3), people need to be shot (3), we need to wait (3), you need to talk (3), we need to listen (3), we need to honor (3), people need to stand (3), i need to stop (3), we need to be talking (3), we need to also (3), we need to speak (3), who need to get (3), we need to go (3), people need to hear (3), we need to wake (3), guns need to be restricted (3), really need to get (3), we need to have a real (3), felt the need to take (3), we need to understand (3), all need to see (3), we need to be outraged (3), people need to start (3), i need to add (3), they need to pay (3), don't need to do (3), we need to ban (3), leaders need to do (3), you need to tell (3), we need to limit (3), we need to prevent (3), we need to really (3)\\
\noindent\emph{Republican}\\you need to know (8), killings need to stop (6), we need to protect (6), we need to get (6), we need to focus (6), we need to know (6), we need to be more (5), really need to look (5), people need to realize (5), people need to stop (5), people need to know (5), we need to pray (4), we need to wait (4), we need to allow (4), they need to kill (4), americans need to be armed (4), really need to stop (4), we need to return (4), we need to kill (4), you need to watch (4), seriously need to change (4), heads need to roll (4), you need to investigate (4), democrats need to stop (4), people need to get (4), you need to address (4), you need to go (4), we need to enforce (4), we need to find (4), guns need to be banned (4), we need to look (4), you need to look (4), you need to ask (4), we need to punish (4)\\
\begin{center}\textbf{NEEDS TO}\end{center}
\noindent\emph{Democrat}\\something needs to be done (8), this needs to stop (8), violence needs to stop (7), it needs to end (6), something needs to change (5), there needs to be stricter (5), this needs to end (5), violence needs to end (5), gop needs to stop (5), america needs to do (4), change needs to happen (4), that needs to be addressed (4), shit needs to stop (4), congress needs to stop (4), it needs to stop (4), america needs to end (3), else needs to happen (3), seriously needs to be done (3), control needs to happen (3), one needs to own (3), what needs to happen (3), reform needs to happen (3), control needs to be a thing (3), there needs to be more (3), there needs to be gun (3), hatred needs to stop (3), it needs to be said (3), world needs to change (3), work needs to be done (3), law needs to change (3), seriously needs to stop (3), control needs to be addressed (3), cnn needs to stop (3), congress needs to do (3), someone needs to tell (3), story needs to be told (3), he needs to take (3), it needs to happen (3), he needs to get (3), congress needs to act (3), this needs to be stopped (3), trump needs to stop (3)\\
\noindent\emph{Republican}\\she needs to go (4), someone needs to ask (4)\\
\begin{center}\textbf{HAVE TO}\end{center}
\noindent\emph{Democrat}\\people have to die (10), we have to do (7), more have to die (6), many have to die (6), don't have to live (5), should not have to fear (4), we have to say (4), don't have to do (4), this have to happen (4), you have to say (4), things have to change (4), should have to fear (4), we have to change (4), we have to talk (4), we have to act (4), we have to keep (4), i have to worry (4), we have to read (3), we have to go (3), we have to have before (3), we have to endure (3), we have to lose (3), lives have to be lost (3), we have to accept (3), we have to hear (3), and not have to worry (3), you have to offer (3), we have to end (3), i have to see (3), shootings have to stop (3), we have to look (3), should have to go (3), we have to pay (3), children have to die (3)\\
\noindent\emph{Republican}\\i have to wonder (7), that have to do (6), shooting have to do (6), you have to say (6), we have to ask (5), i have to say (5), will have to live (5), shootings have to stop (5), i have to go (5), you have to kill (5), i have to agree (5), you have to change (5), you have to be to shoot (5), does have to do (5), nra have to do (5), trump have to do (5), these have to happen (4), we have to listen (4), you have to go (4), you have to put (4), will have to wait (4), going to have to work (4), you have to talk (4), we have to pray (4), we have to see (4), you have to know (4), you have to stop (4), this have to do (4), people have to stop (4), you have to give (4), going to have to get (4), who have to deal (4), you have to ask (4), we have to remember (4), you have to be really (4)\\
\begin{center}\textbf{HAS TO}\end{center}
\noindent\emph{Democrat}\\this has to stop (11), something has to change (6), it has to stop (5), violence has to end (5), shooting has to do (4), this has to end (4), something has to be done (4), more has to happen (4), killing has to stop (4), it has to end (4), what the has to say (4), shit has to stop (3), what has to happen (3), this has to change (3), madness has to end (3), simply has to stop (3), humanity has to offer (3), what has to be done (3), stuff has to stop (3), else has to happen (3), there has to be footage (3), killings has to stop (3)\\
\noindent\emph{Republican}\\it has to do (8), that has to do (6), obama has to say (5), he has to say (5), this has to end (4), there has to be some (4), what has to say (4), one has to ask (4), shit has to end (4)\\

\clearpage\section{Results: Additional Plots}
\label{sec:appendix_plots}

\begin{figure}[H]
 \centering
   \centering
   \includegraphics[width=\linewidth]{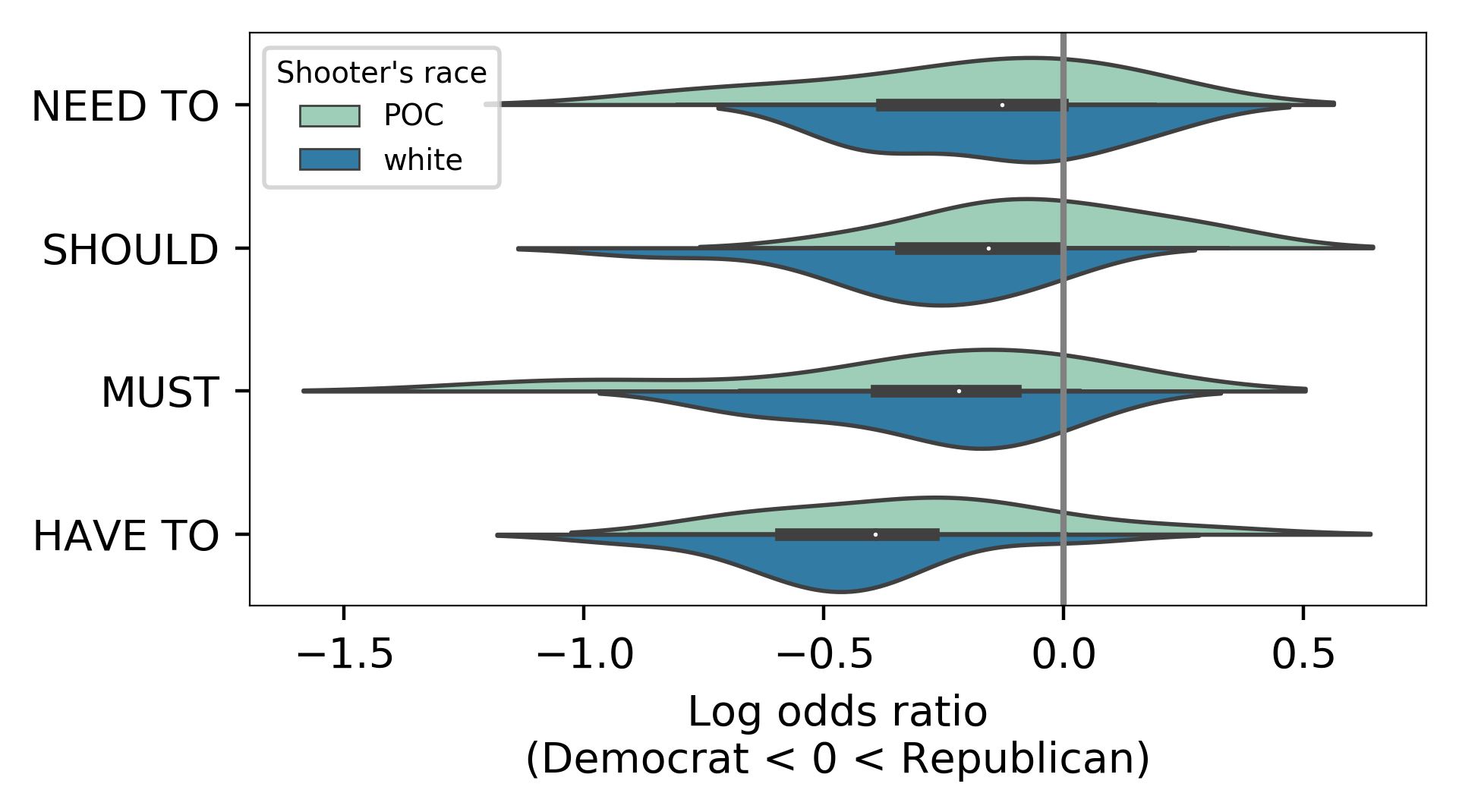}
   \caption{The log odds ratio of necessity modals.}
   \label{fig:modals_log_odds}
\end{figure}

\begin{figure}[H]
   \includegraphics[width=1.01\linewidth]{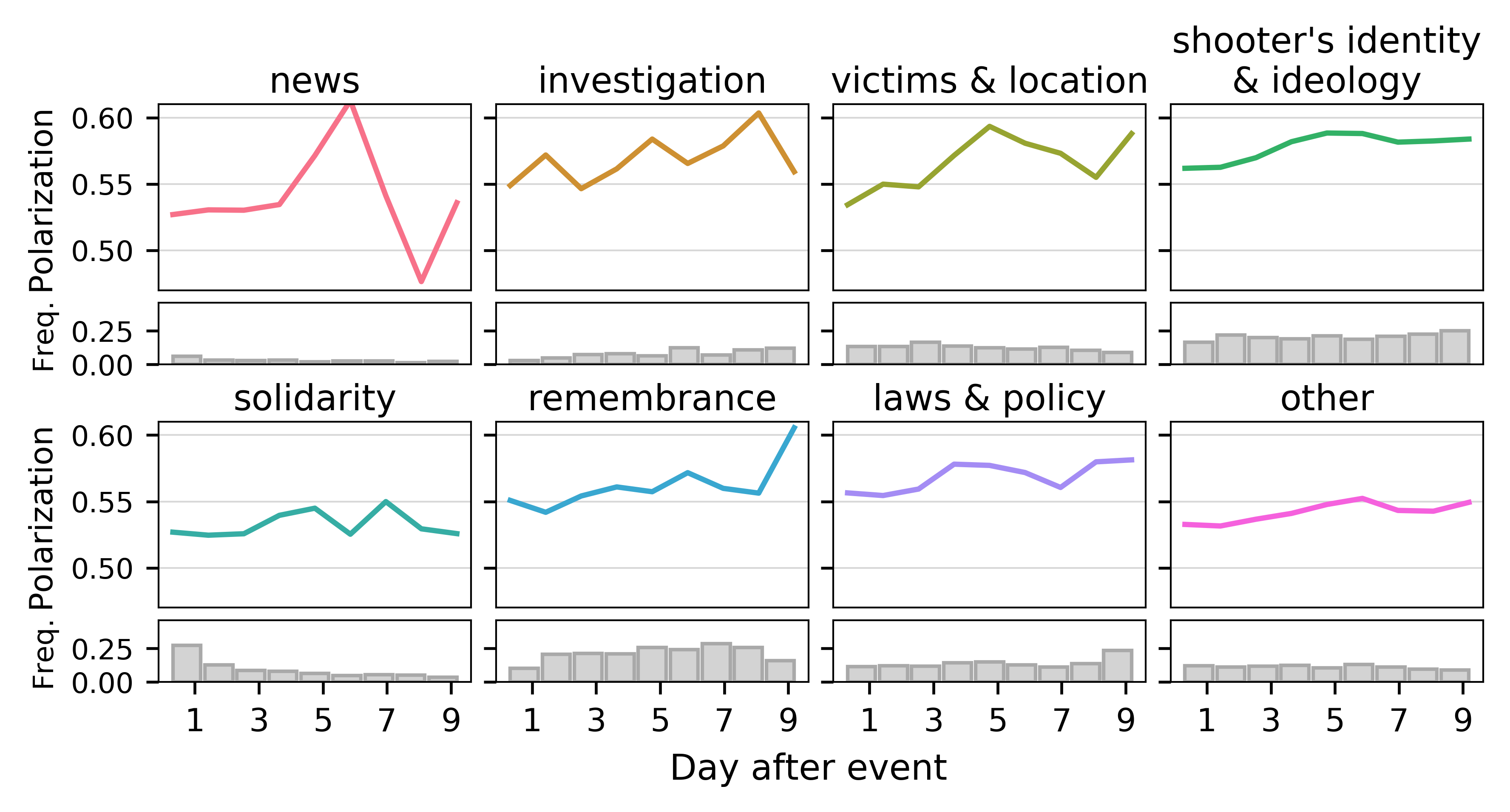}
   \caption{Topic polarization of Orlando over time, as measured by the leave-out estimate of phrase partisanship. The bar charts show the proportion of each topic in the data at a given time.
}
   \label{fig:orlando_topic_overtime}
\end{figure}

\end{document}